\documentclass[twoside,11pt]{article}

\usepackage{blindtext}

\usepackage{dmlr2e}

\usepackage[utf8]{inputenc} \usepackage[T1]{fontenc}    \usepackage{hyperref}       \usepackage{url}            \usepackage{booktabs}       \usepackage{amsfonts}       \usepackage{nicefrac}       \usepackage{microtype}      \usepackage[dvipsnames]{xcolor}         

\usepackage{amsmath}
\usepackage{bm}
\usepackage{bbm}
\usepackage{subfigure}
\usepackage{threeparttable}
\usepackage{nicematrix}
\usepackage{colortbl}
\usepackage{fontawesome6}
\usepackage{simpleicons}
\usepackage{twemojis}

\usepackage{enumitem}

\usepackage{contour}
\usepackage[normalem]{ulem}
\definecolor{color_blue}{HTML}{1f77b4}
\definecolor{color_orange}{HTML}{ff7f0e}
\definecolor{color_gold}{HTML}{c9a227}
\definecolor{color_green}{HTML}{2ca02c}
\definecolor{color_purple}{HTML}{9467bd}

\definecolor{webblue}{HTML}{4285F4}
\definecolor{hfyellow}{HTML}{FFD21E}

\colorlet{light_blue}{color_blue!15}
\colorlet{light_orange}{color_orange!15}
\colorlet{light_gold}{color_gold!15}
\colorlet{light_green}{color_green!15}
\colorlet{light_purple}{color_purple!15}

\contourlength{0.2ex}
\newcommand{\ulc}[2]{{\color{#2}\uline{\phantom{#1}}}\llap{\contour{white}{#1}}}

\newcommand{\nn}{\operatorname{NN}}

\usepackage{lastpage}
\dmlrheading{26}{2026}{1-\pageref{LastPage}}{9/25; Revised 8/26}{8/26}{21-0000}{Elias J\"a\"asaari, Ville Hyv\"onen, Matteo Ceccarello, Teemu Roos, and Martin Aum\"uller} \def\openreview{\url{https://openreview.net/forum?id=6Sx6ra1QLz}} 

\ShortHeadings{VIBE: Vector Index Benchmark for Embeddings}{J\"a\"asaari, Hyv\"onen, Ceccarello, Roos, and Aum\"uller}
\firstpageno{1}

\begin{document}

\title{VIBE: Vector Index Benchmark for Embeddings}

\author{\name Elias J\"a\"asaari \email elias.jaasaari@helsinki.fi \\
       \addr Department of Computer Science\\
       University of Helsinki, Finland
       \AND
       \name Ville Hyv\"onen \email ville.o.hyvonen@helsinki.fi \\
       \addr Department of Computer Science\\
       University of Helsinki, Finland
       \AND
       \name Matteo Ceccarello \email matteo.ceccarello@unipd.it \\
       \addr Department of Information Engineering\\
       University of Padova, Italy
       \AND
       \name Teemu Roos \email teemu.roos@helsinki.fi \\
       \addr Department of Computer Science\\
       University of Helsinki, Finland
       \AND
       \name Martin Aumüller \email maau@itu.dk \\
       \addr IT University of Copenhagen, Denmark
       }

\editor{Matthias Feurer}

\maketitle

\begin{abstract}Approximate nearest neighbor (ANN) search is a performance-critical component of many machine learning pipelines, and rigorous benchmarking is essential for assessing the performance of vector indexes for ANN search. However, the datasets of existing benchmarks no longer represent modern ANN applications, creating a need for an up-to-date benchmark. To address this gap, we introduce Vector Index Benchmark for Embeddings (\textsc{VIBE}), an open-source framework for benchmarking ANN algorithms. \textsc{VIBE} provides a pipeline for generating benchmark datasets with dense embedding models representative of modern applications, including retrieval-augmented generation (RAG). To represent real-world workloads, we also include out-of-distribution (OOD) datasets where the queries and the corpus are drawn from different distributions. These include multimodal retrieval datasets and maximum inner product search (MIPS) datasets covering two recent use cases: approximate attention computation and reductions of multi-vector retrieval to single-vector MIPS. We use \textsc{VIBE} to conduct a comprehensive evaluation of 22 open-source vector-index implementations across 11 in-distribution and 8 out-of-distribution datasets. The benchmark is available at \url{https://github.com/vector-index-bench/vibe}
\end{abstract}


\section{Introduction}

In modern machine learning applications, data such as text and images are often represented by \emph{embeddings} produced by neural encoders. Embeddings are typically dense, high-dimensional vector representations that are trained to have the property that relevant or semantically similar items have similar representations in the embedding space. This makes efficient similarity search in high-dimensional vector spaces an essential operation in applications where corpus items relevant to a query or user context are retrieved, such as information retrieval~\citep{xiong2021approximate,izacard2022unsupervised}, recommendation systems~\citep{covington2016deep,yang2020mixed}, and question answering~\citep{seo2019real,lewis2021paq}.

\begin{figure}[!t]
    \centering
    \includegraphics[width=\linewidth]{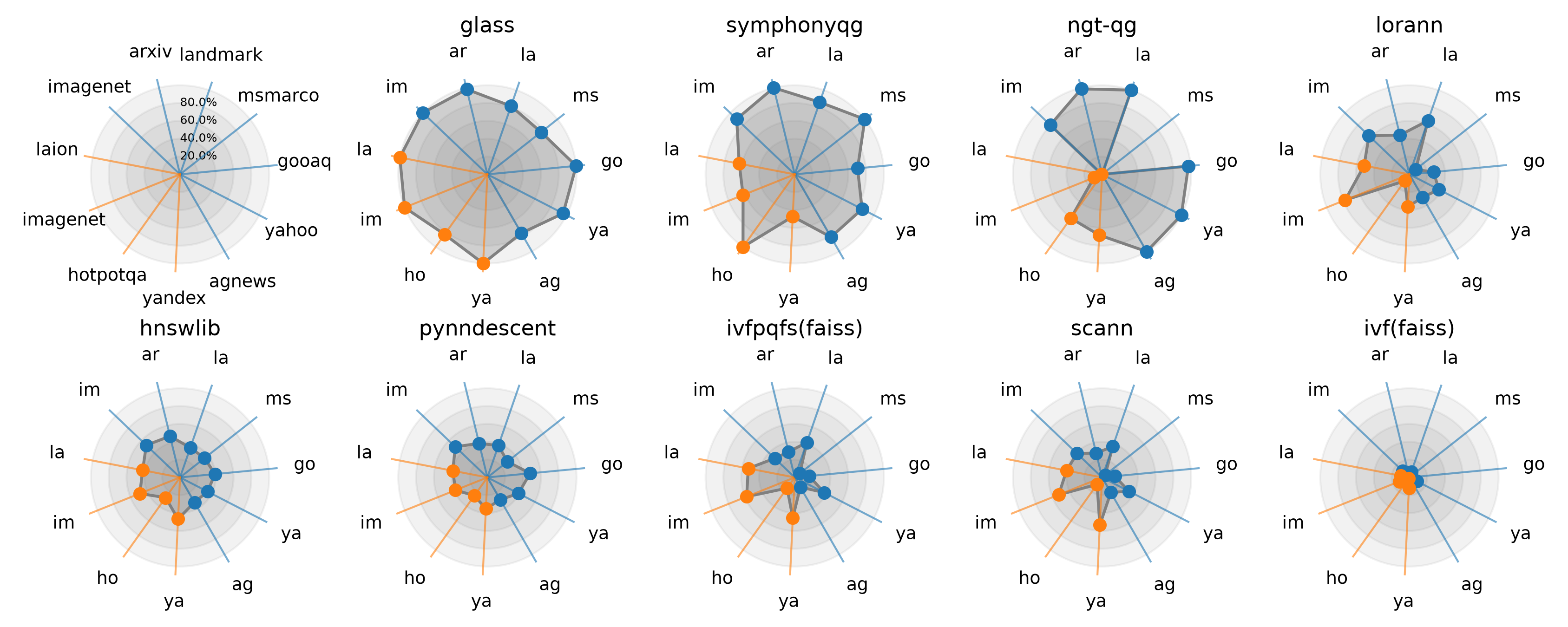}
    \caption[]{Relative query throughput of each algorithm's fastest configuration achieving an average recall of at least $95\%$ on selected datasets with \ulc{in-distribution}{color_blue} and \ulc{out-of-distribution}{color_orange} queries, normalized by the throughput of the best algorithm on each dataset.
    Each circle corresponds to an algorithm, arranged by average normalized query throughput.
    Datasets are arranged in clockwise order by decreasing difficulty as measured by the median query relative contrast $\mathit{RC}_{100}$~\citep{he2012difficulty}, with \texttt{imagenet-clip} being the easiest.
    }
    \label{fig:results-overview}
\end{figure}

Similarity search over vectors is commonly formulated as $k$-nearest neighbor ($k$-nn) search, which retrieves the $k$ corpus vectors most similar to a query vector. Because embedding datasets are typically large and high-dimensional, vector indexes commonly use \emph{approximate nearest neighbor} (ANN) \emph{search}~\citep{indyk1998approximate,bruch2024foundations} to answer these queries with low latency. An important recent use case is retrieval-augmented generation (RAG)~\citep{guu2020retrieval,lewis2020retrieval,borgeaud2022improving,wang2024instructretro}.

Many important modern ANN workloads are \emph{out-of-distribution} (OOD): the corpus and the queries have significantly different distributions. For instance, in multimodal search, the corpus may consist of images while the queries are expressed as text. Even in text retrieval, OOD workloads can arise from inherent differences between queries and documents, such as their length, as well as when instruction-tuned embedding models~\citep{su2023one} use different instructions to encode documents and queries.

Important OOD workloads also arise in maximum inner product search (MIPS). Approximate attention computation in LLMs can be formulated as a MIPS application~\citep{bertsch2023unlimiformer,liu2024retrievalattention,mazare2025inference}, in which the task is to find the keys that have the largest inner products with the query. Another MIPS workload arises in reducing multi-vector retrieval to single-vector retrieval. Multi-vector embedding models such as ColBERT~\citep{khattab2020colbert} represent queries and documents as sets of embeddings and score them using the MaxSim similarity. Specialized retrieval engines~\citep[e.g.,][]{santhanam2022plaid} can search these multi-vector representations directly, but single-vector MIPS reductions~\citep{jayaram2024muvera, jaasaari2026lemur} of multi-vector retrieval are attractive because they allow the use of existing single-vector ANN infrastructure.

Because of the significance of ANN search in modern AI applications, accurate performance evaluation of vector indexes is essential. Currently, the ANN-Benchmarks project~\citep{aumuller2020ann} includes the most comprehensive collection of ANN algorithms. However, most of its datasets, such as MNIST, Fashion-MNIST, SIFT, and GIST, are not representative of modern applications. These datasets have representations, such as raw pixels and image descriptors, that have been superseded by dense vector embeddings. In addition, existing systematic benchmarks do not cover important recent OOD use cases of ANN search. Hence, there is a need for a comprehensive, up-to-date benchmark suite containing both in-distribution embedding workloads and OOD workloads.

To address these issues, we introduce Vector Index Benchmark for Embeddings (\textsc{VIBE}), a benchmarking suite for ANN algorithms on modern embedding datasets. \textsc{VIBE} contains a straightforward pipeline for creating benchmark datasets: we use modern embedding models to generate representative embedding datasets from sources such as arXiv and ImageNet for both in-distribution (Section~\ref{sec:data_sets}) and out-of-distribution (Section~\ref{sec:ood_datasets}) settings. We also introduce four challenging OOD MIPS benchmark datasets covering approximate attention computation and multi-vector-to-single-vector reduction (Section~\ref{sec:ood_datasets}). The benchmark is open, ongoing, and extensible, enabling authors of new ANN algorithms to easily integrate their implementations (Appendix~\ref{app:datasets}). Thus, beyond presenting an up-to-date performance evaluation, our work enables rigorous evaluation in the future. \textsc{VIBE} also features an interactive website that supports deeper analysis of the results (Appendix~\ref{app:website}). 

We use \textsc{VIBE} to evaluate the performance of state-of-the-art ANN algorithms (Section~\ref{sec:results}). An overview of our evaluation for selected algorithms is presented in Figure~\ref{fig:results-overview}. 
Our key findings, discussed further in Section~\ref{sec:key_findings}, are as follows: (i) The top-performing graph-based (SymphonyQG, Glass, NGT-QG) and clustering-based (LoRANN, ScaNN, IVF-PQ) algorithms use various forms of quantization. 
(ii) On the text-retrieval and text-to-image OOD datasets, the best general-purpose methods still outperform specialized OOD methods (see Figure~\ref{fig:OOD-text-to-image}), somewhat surprisingly, even though the performance of these general-purpose methods deteriorates significantly on OOD queries (see Figure~\ref{fig:performance-gap}).
(iii) On approximate attention computation datasets, general-purpose ANN methods fail to reach the highest recall levels, and specialized OOD methods outperform them (see Figure~\ref{fig:OOD-approximate-attention-computation}).

We summarize our main contributions below: 
\begin{itemize}
    \item \textbf{Benchmark for modern use cases.} \textsc{VIBE} is an open benchmarking framework for evaluating the performance of vector indexes on embedding datasets used in modern applications, such as retrieval-augmented generation. The framework also supports high-performance computing (HPC) environments and GPU-based implementations.
    \item \textbf{Extensible architecture.} \textsc{VIBE} contains a transparent pipeline for generating new benchmark datasets using embedding models, making the datasets easy to update. Its modular architecture makes it straightforward to add new algorithms (Appendix~\ref{app:datasets}).
    \item \textbf{Evaluation in the OOD setting.} We introduce novel OOD datasets spanning several recent use cases (Section~\ref{sec:ood_datasets}). These datasets enable systematic comparisons between general-purpose ANN methods and OOD-specific methods (Section~\ref{sec:OOD_setting}).
    \item \textbf{Comprehensive evaluation.}
    We use \textsc{VIBE} to conduct a performance evaluation of 22 vector index implementations across 11 in-distribution and 8 OOD datasets (Section~\ref{sec:results}). Our evaluation and visualization tools (Appendix~\ref{app:website}) support in-depth comparisons and reveal opportunities for future work (Section~\ref{sec:key_findings}).
    \item \textbf{Fine-grained performance analysis.} Beyond recall--throughput curves, \textsc{VIBE} reports finer-grained performance breakdowns, including comparisons between easy and hard queries (Figure~\ref{fig:in-distribution-split-performance}) and between in-distribution and OOD queries (Figure~\ref{fig:performance-gap}). 

\end{itemize}

\section{Background}

\subsection{ANN search}
\label{sec:ann:search}

\paragraph{Definition} Denote the $m$ \emph{corpus} points by $\{\mathbf{c}_1, \dots, \mathbf{c}_m\} \subset \mathbb{R}^d$ and let $\rho\colon\mathbb{R}^d \times \mathbb{R}^d \rightarrow \mathbb{R}$ be a dissimilarity measure. The task of \(k\)-nearest neighbor (\(k\)-nn) search is to retrieve the indices of the \(k\) corpus points that are most similar to the \emph{query point} \(\mathbf{x} \in \mathbb{R}^d\), i.e., to retrieve
\[
\nn_k(\mathbf{x}) = \{\pi(1), \dots, \pi(k)\}, \qquad \rho(\mathbf{x}, \mathbf{c}_{\pi(1)}) \leq \cdots \leq \rho(\mathbf{x}, \mathbf{c}_{\pi(m)}),
\]
where \(\pi\) is a permutation of \([m]=\{1,\ldots,m\}\), with ties broken according to a fixed rule. The task of \emph{approximate nearest neighbor} (ANN) \emph{search} is to design a \emph{vector index} that enables approximating the exact $k$-nn solution in sublinear time. 
In this paper, we use \emph{approximate} to refer to \emph{inexact solutions} without guarantees~\citep{DBLP:journals/debu/AumullerC23}.

\paragraph{Dissimilarity measures} The most common dissimilarity measures for ANN search are the Euclidean distance $\rho(\mathbf{a}, \mathbf{b}) = \|\mathbf{a} - \mathbf{b}\|_2$, the cosine distance $\rho(\mathbf{a}, \mathbf{b}) = 1 - \langle \mathbf{a} / \|\mathbf{a}\|_2, \mathbf{b} / \|\mathbf{b}\|_2 \rangle$, and the negative inner product $\rho(\mathbf{a}, \mathbf{b}) = -\langle \mathbf{a}, \mathbf{b} \rangle$. Minimizing the negative inner product is equivalent to \emph{maximum inner product search} (MIPS). Embeddings are commonly normalized to unit norm, in which case all three dissimilarity measures yield the same nearest neighbors.

\paragraph{Performance measures} For a query point, \emph{recall} is the fraction of its $k$ returned neighbors that are among its true $k$ nearest neighbors. We measure the effectiveness of ANN algorithms by the \emph{average recall}, obtained by averaging this per-query recall over all test queries.
The efficiency of ANN algorithms is typically measured by the average query latency or, equivalently, by throughput, which is measured by \emph{queries per second} (QPS). 

\paragraph{Difficulty measures}
To assess the difficulty of a query $\mathbf{x}$, we consider its \emph{relative contrast} at $k$,
$\mathit{RC}_k(\mathbf{x}) = \frac{1}{\rho(\mathbf{x}, \mathbf{c}_{\pi(k)})} \sum_{j\in[m]} \rho(\mathbf{x}, \mathbf{c}_j) / m$,
that is, the ratio of the average distance between the query and all the corpus points in the dataset to the distance between the query and its $k$-th nearest neighbor~\citep{he2012difficulty}.
The relative contrast is a good proxy for the effort required to distinguish nearby and distant points~\citep{DBLP:journals/is/AumullerC21,ceccarello2025evaluating}, with small values characterizing difficult queries.

\paragraph{Out-of-distribution (OOD) setting} There is recent interest in ANN search in the \emph{out-of-distribution} (OOD) setting~\citep[e.g.,][]{cayton2007learning,jaiswal2022ood,hyvonen2024multilabel,chen2024roargraph,jaasaari2024lorann,mazare2025inference}, in which the corpus points and query points are drawn from different distributions. If there is a large difference between the query distribution and the corpus distribution, using a representative sample from the query distribution to adapt the index during construction can speed up ANN search significantly~\citep{jaiswal2022ood,hyvonen2024multilabel,chen2024roargraph}.

The OOD setting is common in multimodal search, where queries and corpus points have different modalities: for example, the queries may be text while the corpus consists of images, with both mapped into a shared embedding space. ANN search has also been used to accelerate attention computation in long-context LLM inference, in which the distributions of the keys and queries differ significantly~\citep{liu2024retrievalattention,mazare2025inference}. Asymmetric reductions for multi-vector retrieval can likewise produce OOD single-vector MIPS workloads by mapping queries and documents differently~\citep{jayaram2024muvera,jaasaari2026lemur}. \textsc{VIBE} includes datasets for all of these out-of-distribution settings (Section~\ref{sec:ood_datasets}).

\subsection{Taxonomy of ANN algorithms}

Broadly, approximate nearest neighbor search algorithms can be classified into graph-, clustering-, tree-, and hashing-based methods according to the index structure they use. We give a brief overview of each category below. \textsc{VIBE} includes methods from all four categories; Table~\ref{tab:algorithms} summarizes the algorithms included in the benchmark, and Table~\ref{tab:impdetails} in Appendix~\ref{app:impdetails} provides citations and implementation details. 

\paragraph{Graphs} Graph-based methods, such as HNSW~\citep{malkov2018efficient}, NSG~\citep{fu2017nsg}, Vamana~\citep{jayaram2019diskann}, and SymphonyQG~\citep{gou2025symphonyqg}, use a navigable proximity graph of the corpus as an index structure and retrieve the approximate nearest neighbors of the query point via greedy or bounded best-first graph traversal. 

\paragraph{Clustering} Clustering-based methods cluster the corpus points and evaluate the distances from the query point only to the corpus points in the clusters closest to the query point. The points in the selected clusters can be ranked using an approximate score-computation method, such as product quantization (PQ) in IVF-PQ~\citep{jegou2011product}, anisotropic vector quantization in ScaNN~\citep{guo2020accelerating}, or reduced-rank regression in LoRANN~\citep{jaasaari2024lorann}, to select a candidate set that may subsequently be reranked using exact scores.

\paragraph{Trees} Tree-based methods use space-partitioning index structures, including $k$-d trees~\citep{friedman1977algorithm}, principal axis trees~\citep{sproull1991refinements}, and random projection trees~\citep{dasgupta2008random}. Efficient tree-based methods, such as Annoy~\citep{annoy} and MRPT~\citep{hyvonen2016fast}, use ensembles of randomized trees.

\paragraph{Hashing} Hashing-based methods, such as PUFFINN~\citep{aumuller2019puffinn} and FALCONN++~\citep{pham2022falconn++}, partition the space using locality-sensitive hashing or filtering schemes and evaluate the distances from the query point to only those corpus points stored in buckets probed for the query. Hashing-based methods often use several randomized hash tables to improve the search accuracy, and these approaches can provide strong probabilistic correctness guarantees.

\section{Related work}

The ANN-Benchmarks project\,\footnote{\url{https://github.com/erikbern/ann-benchmarks}}~\citep{aumuller2020ann} is the most well-known and comprehensive benchmarking suite for vector indexes. However, its datasets are no longer representative of the datasets used in modern applications of ANN search, and the project does not contain a systematic evaluation of the OOD setting. Earlier performance evaluations~\citep{li2019approximate,aumuller2020ann} do not contain modern embedding datasets or the recent top-performing ANN methods. More recent performance evaluations~\citep{wang2021comprehensive,azizi2025graph} consider only graph-based ANN methods and do not provide an open and extensible benchmarking framework.

The big-ann-benchmarks project~\citep{simhadri2022results,simhadri2025results} has single-dataset tracks for benchmarking ANN algorithms on four specialized tasks, consisting of filtered search, streaming search, sparse search, and OOD search. Other recent studies focus specifically on streaming search~\citep{zeng2024candy,wang2026candor} or filtered search~\citep{iff2026benchmarking,lim2026revisiting,zhang2026vecbench}.

\citet{chen2026reveal} benchmark vector indexes from the perspective of end-to-end task performance. Some vector database companies have their own benchmarks, such as VectorDBBench\,\footnote{\url{https://github.com/zilliztech/VectorDBBench}}, and \citet{kang2025bigvectorbench} likewise compare vector database systems using an integrated benchmark that covers embedding latency and compound queries. These production-oriented systems combine a vector index with database and deployment functionality, and such benchmarks therefore address the complementary question of end-to-end system performance. In contrast, \textsc{VIBE} isolates open-source ANN implementations to support reproducible, algorithm-level comparisons.
\section{\textsc{VIBE}}
\label{sec:VIBE}

This section describes the datasets included in \textsc{VIBE}. It also presents features that support future research, including quantized datasets and broad hardware support.

\begin{table}[t]
    \centering
    \caption[]{\textsc{VIBE} in-distribution datasets. For references and details about the data sources and embedding models, see Appendix~\ref{sec:models-data-sources}. Embeddings with dissimilarity measure ``any'' are normalized to unit $\ell_2$ norm, and each algorithm can choose an appropriate measure. Here $m$ = number of corpus points, $d$ = embedding dimension.}
    \label{tab:id-datasets}
    \begin{tabular}{lllrrl}
        \toprule
        Data Source & Model & Type & $m$ ($\downarrow$) & $d$ & Measure \\
        \midrule
        DPR & Jina & Text & 20 969 760 & 768 & any \\
        MS MARCO & Qwen & Text & 8 840 823 & 1024 & any \\
        GooAQ & DistilRoBERTa & Text & 1 475 024 & 768 & any \\
        arXiv & Nomic Text & Text & 1 344 643 & 768 & any \\
        ImageNet & CLIP & Image & 1 281 167 & 512 & any \\
        Twitter & GloVe & Word & 1 192 514 & 200 & cosine \\
        AGNews & MXBAI & Text & 769 382 & 1024 & Euclidean \\
        Landmark & DINO & Image & 760 757 & 768 & cosine \\
        Landmark & Nomic Vision & Image & 760 757 & 768 & any \\
        Yahoo & MiniLM & Text & 677 305 & 384 & any \\
        iNaturalist & ResNet & Image & 499 000 & 2048 & cosine \\
        \bottomrule
    \end{tabular}
\end{table}

\subsection{In-distribution datasets}
\label{sec:data_sets}

The in-distribution datasets included in \textsc{VIBE} are listed in Table~\ref{tab:id-datasets}. We create embedding datasets using widely used models applied to common text and image data sources. We choose text embedding models that include the older but widely used models DistilRoBERTa~\citep{sanh2019distilbert,liu2019roberta,reimers2019sentencebert} and MiniLM~\citep{wang2020minilm,reimers2019sentencebert}, as well as recent top-performing~\citep{muennighoff2023mteb} industry models, including nomic-embed-text-v1.5~\citep{nussbaum2025nomic}, mxbai-embed-large-v1~\citep{emb2024mxbai}, jina-embeddings-v5-text-nano~\citep{akram2026jina}, and Qwen3-Embedding-0.6B~\citep{qwen3embedding}. For images, we similarly use embeddings from a ResNet-50 convolutional neural network~\citep{he2016deep}, as well as modern image embedding models: CLIP~\citep{radford2021learning}, DINO~\citep{caron2021emerging}, and nomic-embed-vision-v1.5~\citep{nussbaum2024nomic}. We also include the older GloVe~\citep{pennington2014glove} word embedding dataset due to its prevalence in earlier benchmarks.

To select the benchmark datasets, we start with this broad pool of widely used embedding models spanning different model families and output dimensionalities. For each model, we choose suitable public data sources (see Appendix~\ref{sec:models-data-sources}) whose document lengths fit the model's context window. We then select a small subset of model--dataset combinations by using relative contrast (see Section~\ref{sec:ann:search}) to guide selection toward a diverse set of datasets, while manually ensuring that the final collection includes both text and image datasets, different model families, and a broad range of output dimensionalities.  

Most of the resulting datasets contain on the order of one million corpus vectors, which keeps comprehensive evaluations across datasets, methods, and hyperparameters practical and allows using a common hyperparameter grid for each method throughout. We also include two larger datasets to help evaluate method performance with increasing corpus size.

\begin{figure}[!t]
    \centering
    \includegraphics[width=0.82\linewidth]{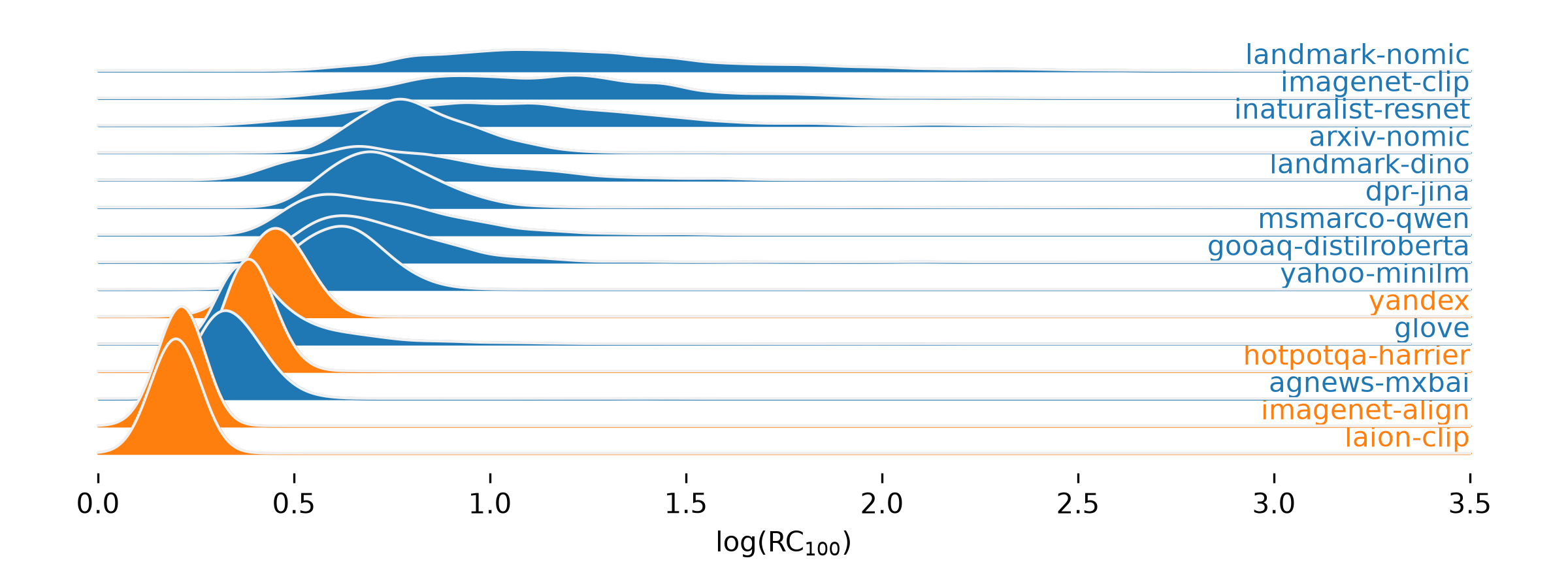}
    \caption[]{Distribution of the relative contrast query difficulty metric $\mathit{RC}_{100}$ for selected
    \ulc{in-distribution}{color_blue} and
    \ulc{out-of-distribution}{color_orange} datasets. (Note the logarithmic scale.)
    Density curves are arranged by their median relative contrast. Lower relative contrast corresponds to a harder nearest neighbor search problem, while a flatter curve represents greater variation in query difficulty. In general, queries in image embedding datasets tend to be easier and span a wider range of difficulty than those in text embedding datasets, while OOD datasets pose harder nearest neighbor search problems than in-distribution datasets. Inner product datasets are omitted because relative contrast is not meaningful for negative inner product dissimilarity.
    }
    \label{fig:dataset:difficulty}
\end{figure}

The datasets span a wide range of difficulty, as quantified by relative contrast. Figure~\ref{fig:dataset:difficulty} summarizes the distribution of relative contrast values across different datasets. Visually, datasets stemming from image embeddings have a wider range of relative contrast values than those stemming from text embeddings.

For most text embeddings, the model outputs are normalized to unit norm by default. In this case, cosine similarity, Euclidean distance, and inner product induce the same nearest-neighbor ranking, and each algorithm can choose an appropriate distance measure. Otherwise, we use Euclidean distance for text embeddings to represent a clustering task. For image embeddings that are not normalized, we use cosine distance. For each in-distribution dataset, we hold out 1000 points as test queries and use the remaining points as the corpus.

\subsection{Out-of-distribution datasets}
\label{sec:ood_datasets}

We choose the OOD datasets to cover a diverse array of retrieval tasks. The eight OOD datasets comprise one text-retrieval dataset, three text-to-image retrieval datasets, two single-vector MIPS datasets obtained by reducing multi-vector retrieval, and two MIPS datasets derived from approximate attention workloads (Table~\ref{tab:ood-datasets}).

We create the HotpotQA--Harrier text-retrieval dataset from HotpotQA~\citep{yang2018hotpotqa} using the harrier-oss-v1-270m model~\citep{microsoft2026harrier}. Its queries are out-of-distribution because the Harrier model applies a retrieval prompt only to queries, which are also typically short questions, whereas the documents are longer passages. For ImageNet--ALIGN, we use the ALIGN model~\citep{jia2021scaling} to embed ImageNet images as corpus vectors and text from the ImageNet-Captions dataset~\citep{fang2022data} as query vectors. We also include subsets of the existing LAION and Yandex text-to-image datasets previously used to benchmark OOD ANN algorithms~\citep{simhadri2025results,chen2024roargraph}.

\begin{table}[t]
    \centering
    \caption{\textsc{VIBE} out-of-distribution datasets. For details about the data sources and embedding models, see Appendix~\ref{sec:models-data-sources}. Embeddings with dissimilarity measure ``any'' are normalized to unit norm, and each algorithm can choose an appropriate measure. The two CQADupStack reductions and the \texttt{llama} and \texttt{yi} datasets are MIPS workloads and use (negative) inner product. Here $m$ = number of corpus points, $m_\text{learn}$ = number of query distribution samples, $d$ = embedding dimension.}
    \label{tab:ood-datasets}
    \begin{tabular}{lllrrrl}
        \toprule
        Data Source & Model & Type & $m$ ($\downarrow$) & $m_{\mathrm{learn}}$ & $d$ & Measure \\
        \midrule
        HotpotQA & Harrier & Text & 5 233 329 & 96 845 & 640 & any \\
        ImageNet & ALIGN & Text-To-Image & 1 281 167 & 315 648 & 640 & any \\
        LAION & CLIP & Text-To-Image & 1 000 448 & 999 448 & 512 & any \\
        Yandex & SE-ResNeXt & Text-To-Image & 999 000 & 999 000 & 200 & cosine \\
        CQADupStack & MUVERA & Multi-vector & 457 149 & 1 331 947 & 5120 & IP \\
        CQADupStack & LEMUR & Multi-vector & 457 149 & 1 331 947 & 2048 & IP \\
        Llama-3-8B & - & Attention & 256 921 & 255 921 & 128 & IP \\
        Yi-6B-200K & - & Attention & 187 843 & 186 843 & 128 & IP \\
        \bottomrule
    \end{tabular}
\end{table}
 
For the approximate attention workloads, we construct two datasets by extracting keys and queries from one selected attention head in each of two long-context language models: Yi-6B-200K and Llama-3-8B-Instruct-262k. These attention workloads are out of distribution because the models compute keys and queries using separate learned query and key projections~\citep{liu2024retrievalattention,mazare2025inference}. For the multi-vector retrieval workloads, we apply MUVERA~\citep{jayaram2024muvera} and LEMUR~\citep{jaasaari2026lemur} to 128-dimensional ColBERTv2~\citep{santhanam2022colbertv2} token embeddings of CQADupStack~\citep{hoogeveen2015cqadupstack}. These methods reduce multi-vector similarity search to single-vector MIPS, producing 5120-dimensional MUVERA vectors and 2048-dimensional LEMUR vectors. The resulting datasets are OOD because these reductions use different mappings for queries and documents.  For details, see Appendix~\ref{sec:models-data-sources}.

For each OOD dataset, we include a larger training sample from the query distribution, allowing query-distribution-aware methods~\citep{hyvonen2024multilabel,chen2024roargraph,jaasaari2024lorann} to use information about the query distribution during index construction. Some of these methods additionally require the exact nearest neighbors of the training queries in the corpus as input. We precompute and distribute these neighbors with the datasets.

For each OOD dataset, we use 1000 held-out test queries drawn from its query distribution. We quantify the distribution shift between each dataset's corpus and query distributions in Appendix~\ref{app:distribution_shift}; each dataset exhibits substantial distribution shift.

\subsection{Quantized datasets}

State-of-the-art embeddings are increasingly trained to perform well when quantized to 8-bit integer or even binary precision due to their high computational and storage costs\,\footnote{See, e.g., \url{https://huggingface.co/blog/embedding-quantization}}. For example, distance computations using the Hamming distance on binarized data are much faster to evaluate using bitwise XOR and specialized population-count instructions, such as those provided by the AVX-512 VPOPCNTDQ extension. \textsc{VIBE} provides quantized variants of embedding datasets and functions for creating them; a subset of the default algorithms supports these variants (see Appendix~\ref{app:binary-and-GPU}).

\subsection{Support for research}
\label{sec:research_support}

\textsc{VIBE} features additional components that support future research on vector indexes. Its evaluation pipeline builds on the evaluation code of ANN-Benchmarks~\citep{aumuller2020ann} and extends it to support a broader range of hardware platforms. Dropping the dependency on Docker, the benchmark pipeline supports high-performance computing (HPC) platforms through Apptainer (Singularity) containerization~\citep{kurtzer2017singularity} and NUMA-aware execution. The benchmark also supports GPU implementations (see Figure~\ref{fig:binary-and-GPU}, right panel).

The benchmark features a companion website\,\footnote{\raisebox{-0.18em}{\twemoji[height=1em]{globe_with_meridians}}\hspace{0.5mm} \url{https://vector-index-bench.github.io/}} that provides interactive plots (see Appendix~\ref{app:website}) to further explore benchmark results and dataset characteristics. In particular, the website allows users to compare algorithms at different recall levels and interactively examine every tested configuration, including configurations that do not lie on the Pareto frontier. The companion website and the benchmark suite are designed to make it easy to add new datasets and algorithms (see Appendix~\ref{app:datasets}).

\begin{table*}[!t]
\centering
\caption{Summary of the evaluated algorithms. `*' marks implementations that have not been included in earlier benchmarks. See Table~\ref{tab:impdetails} in Appendix~\ref{app:impdetails} for citations and implementation details.}
\label{tab:algorithms}
\begin{NiceTabular}{ccccc}
\toprule
\multicolumn{2}{@{\hskip 24pt}c@{\hskip 24pt}}{\Block[fill=light_blue]{*-2}{}\textbf{Graphs}} &
\multicolumn{1}{@{\hskip 24pt}c@{\hskip 24pt}}{\Block[fill=light_purple]{*-1}{}\textbf{Trees}} &
\multicolumn{1}{@{\hskip 24pt}c@{\hskip 24pt}}{\Block[fill=light_green]{*-1}{}\textbf{Clustering}} &
\multicolumn{1}{@{\hskip 24pt}c@{\hskip 24pt}}{\Block[fill=light_orange]{*-1}{}\textbf{Hashing}}\\
\midrule
\href{https://github.com/zilliztech/pyglass}{Glass} &
\href{https://github.com/lmcinnes/pynndescent}{PyNNDescent} &
\href{https://github.com/spotify/annoy}{ANNOY} &
\href{https://github.com/facebookresearch/faiss}{IVF} &
\href{https://github.com/puffinn/puffinn}{PUFFINN} \\
\href{https://github.com/facebookresearch/faiss}{NSG} &
\href{https://github.com/nmslib/hnswlib}{HNSW} &
\href{https://github.com/vioshyvo/mrpt}{MRPT} &
\href{https://github.com/facebookresearch/faiss}{IVF-PQ} &
\href{https://github.com/NinhPham/FalconnPP}{FALCONN++${}^*$} \\
\href{https://github.com/NGT-labs/NGT/}{NGT-QG} &
\href{https://github.com/VectorDB-NTU/RaBitQ-Library}{HNSW-RaBitQ${}^*$} &
\href{https://github.com/ejaasaari/mlann}{MLANN${}^*$} &
\href{https://github.com/VectorDB-NTU/RaBitQ-Library}{IVF-RaBitQ${}^*$} &
 \\
\href{https://github.com/microsoft/DiskANN}{Vamana} &
\href{https://github.com/BlaiseMuhirwa/flatnav}{FlatNav${}^*$} &
 &
\href{https://github.com/google-research/google-research/tree/master/scann}{ScaNN} &
 \\
\href{https://github.com/intel/ScalableVectorSearch}{LVQ${}^*$} &
\href{https://github.com/intel/ScalableVectorSearch}{LeanVec${}^*$} &
 &
\href{https://github.com/ejaasaari/lorann}{LoRANN${}^*$} &
 \\
\href{https://github.com/matchyc/RoarGraph}{RoarGraph${}^*$} &
\href{https://github.com/gouyt13/SymphonyQG}{SymphonyQG${}^*$} &
 &
 &
 \\
\bottomrule
\end{NiceTabular} \vspace*{2pt}
\end{table*}
 
\section{Performance evaluation}
\label{sec:results}

\subsection{Experimental setup}
\label{sec:experimental_setup}

Each experiment is run on a compute node with 384 GB of RAM and two Intel Xeon Gold 6230 (Cascade Lake) CPUs. Each CPU has 20 cores and supports AVX-512 instructions. All algorithms are benchmarked using a single worker thread on one physical CPU core, with simultaneous multithreading (SMT) disabled, and we use \texttt{numactl} to bind the algorithm execution to one CPU socket and constrain memory allocation to its associated local NUMA node. All of the compared implementations are written in C++ with the exception of PyNNDescent~\citep{wei2021pynndescent,pynndescent}, which uses the Numba library extensively for just-in-time compilation. All implementations are heavily optimized and use techniques such as vectorization to accelerate distance computations.

By default, we use $k = 100$ in all experiments because it is representative of real-world applications: retrieval systems commonly retrieve more candidates than they ultimately return, allowing the candidates to be filtered, diversified, or reranked using more expensive models. We use average recall to measure the effectiveness and queries per second (QPS) to measure the efficiency of the algorithms. Each query hyperparameter combination is run five times, and we use the minimum query time across these runs to compute QPS.

\paragraph{Algorithms}
We review the recent literature and the earlier benchmarking efforts~\citep[e.g.,][]{aumuller2020ann} to identify the top-performing ANN methods for our performance evaluation.
The inclusion criteria are as follows: (i) an implementation of the method is available as a library with Python bindings; (ii) the library is open source; (iii) the method is top-performing within its class or is widely used in practice.
See Table~\ref{tab:algorithms} for the included methods and Table~\ref{tab:impdetails} in Appendix~\ref{app:impdetails} for citations and implementation details.

\paragraph{Hyperparameter settings}
Most implementations expose several hyperparameters whose optimal values depend on the workload, and
\textsc{VIBE} allows for easy configuration of a grid search for these hyperparameters. 
Initial choices for older methods were obtained from the hyperparameter grids of the ANN-Benchmarks project~\citep{aumuller2020ann} and from the library documentation and the corresponding research papers for new methods. We then expanded these grids as necessary based on intermediate results.

In the figures in this section, we report the Pareto frontiers over all evaluated configurations. To keep the comparisons consistent and avoid dataset-specific hyperparameter tuning, we use the same hyperparameter grid for every dataset. The evaluated hyperparameter grids for all methods are available in our code repository\,\footnote{\raisebox{-0.18em}{\simpleicon{github}} \url{https://github.com/vector-index-bench/vibe/}}.

\begin{figure}[!t]
    \centering
    \begin{minipage}{0.82\linewidth}
    \includegraphics[width=\linewidth]{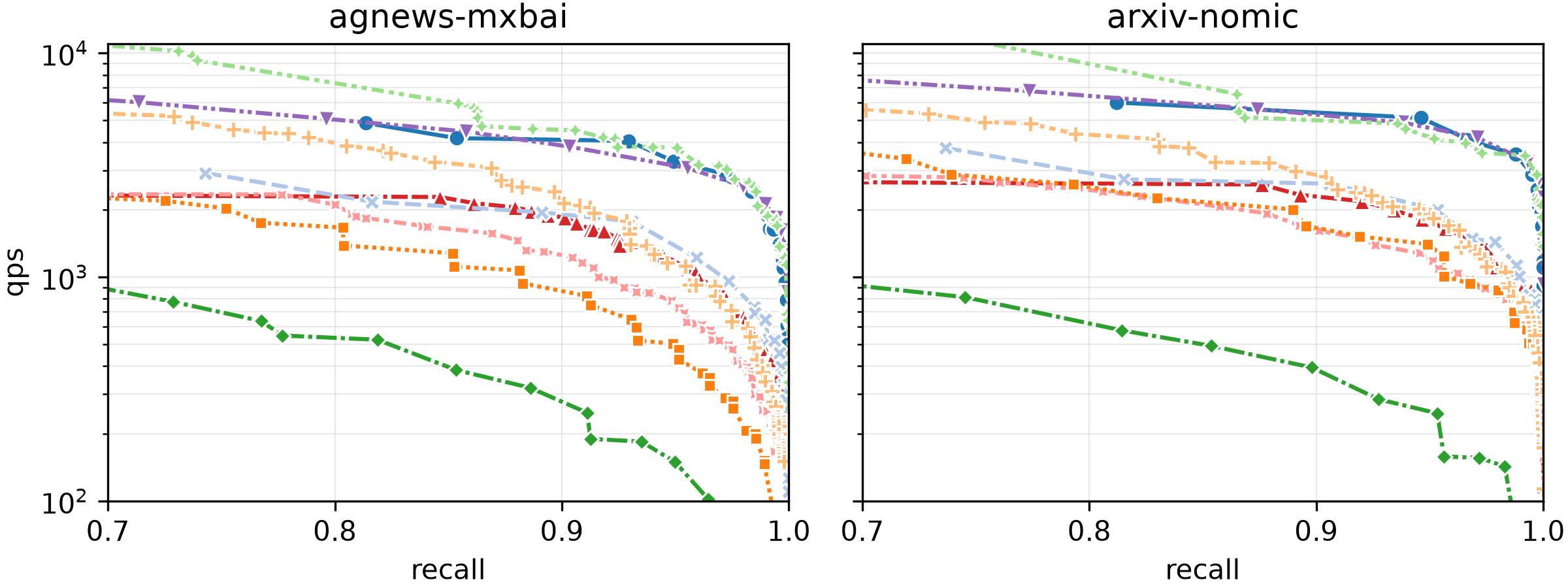}
    \end{minipage}
    \begin{minipage}{0.17\linewidth}
    \includegraphics[width=\linewidth]{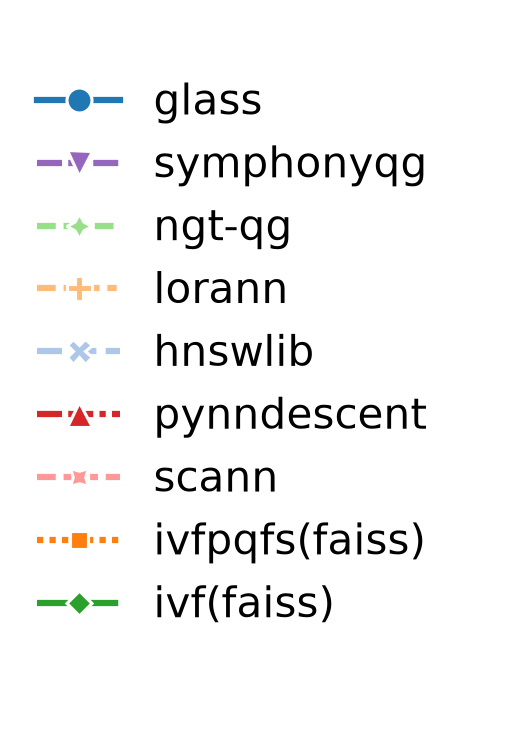}
    \end{minipage}
    \caption[]{Recall/throughput trade-off on two \textbf{text embedding} datasets (in-distribution queries). Among the methods shown, the graph-based methods Glass, NGT-QG, and SymphonyQG achieve the highest throughput.
    }
    \label{fig:in-distribution-text}
\end{figure}

\begin{figure}[!t]
    \centering
    \begin{minipage}{0.82\linewidth}
    \includegraphics[width=\linewidth]{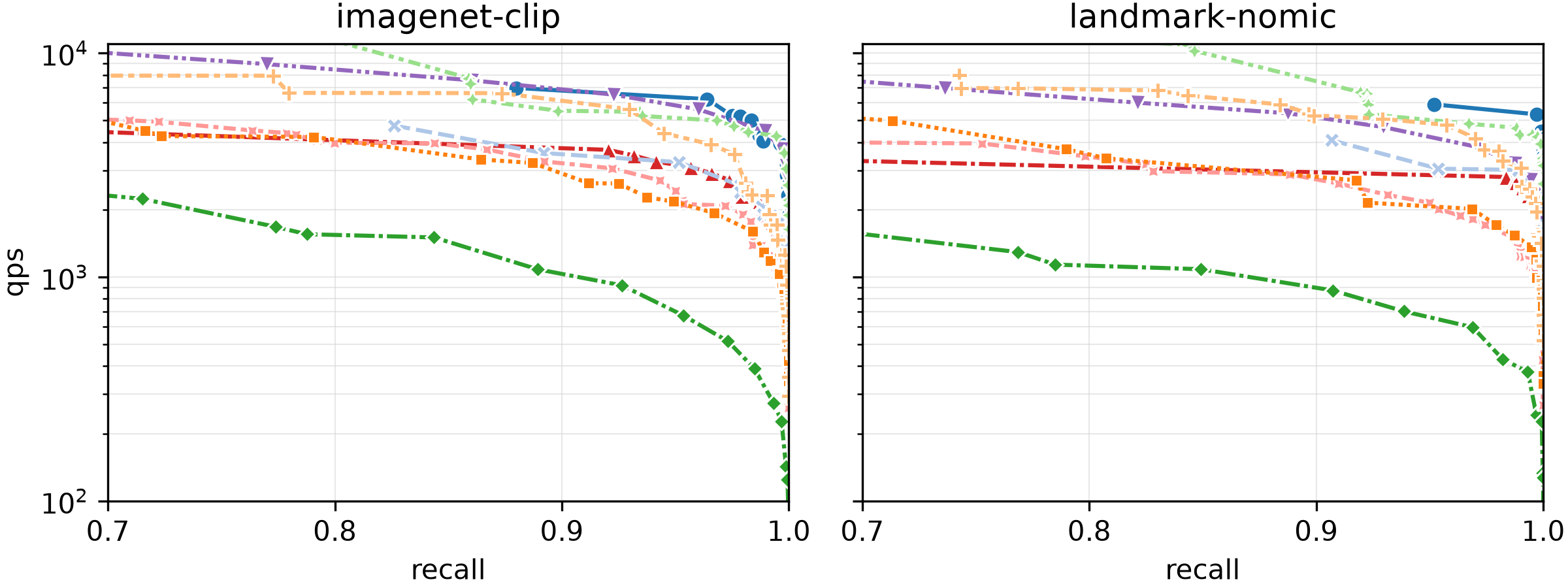}
    \end{minipage}
    \begin{minipage}{0.17\linewidth}
    \includegraphics[width=\linewidth]{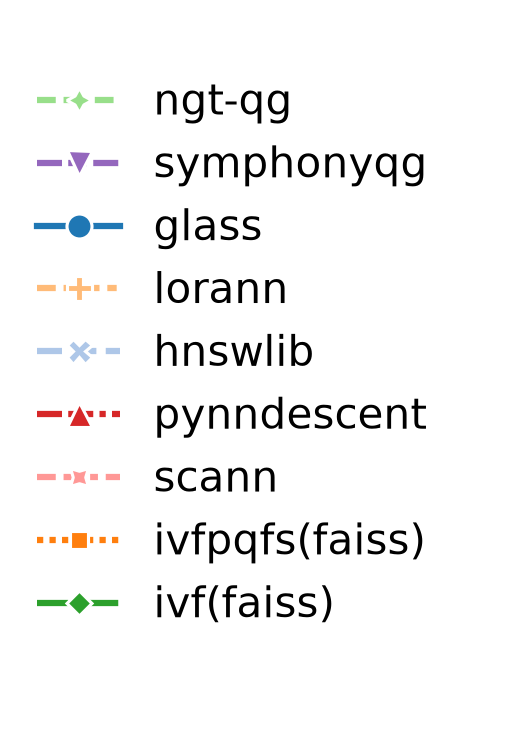}
    \end{minipage}
    \caption[]{Recall/throughput trade-off on two \textbf{image embedding} datasets (in-distribution queries). The graph-based methods Glass, NGT-QG, and SymphonyQG, and the clustering-based method LoRANN achieve the highest throughput.
    }
    \label{fig:in-distribution-image}
\end{figure}

\subsection{In-distribution setting}
\label{sec:in_distribution_setting}

We compare the ANN algorithms on eleven in-distribution datasets. Figure~\ref{fig:results-overview} presents an overview for seven representative in-distribution datasets. For selected algorithms, the recall--QPS curves are reported in Figure~\ref{fig:in-distribution-text} for two representative text-embedding datasets and in Figure~\ref{fig:in-distribution-image} for two representative image-embedding datasets. The results for all datasets can be found in Appendix~\ref{app:selected-algorithms}. For clarity, the figures in this section show only nine selected methods. Results for all methods are available on the companion website, while Appendix~\ref{app:all-algorithms} presents full results for two datasets. Appendix~\ref{app:hypervolume} additionally summarizes the recall--throughput trade-offs using normalized hypervolume.

The general trend is that graph-based and clustering-based methods significantly outperform hashing-based and tree-based methods. At 95\% average recall, the highest-throughput method on every dataset is one of the graph-based methods Glass, SymphonyQG, or NGT-QG. However, as shown in Appendix~\ref{app:construction_time}, graph-based methods generally have the disadvantage of longer index construction times.

\begin{figure}[!t]
    \centering
    \includegraphics[width=0.9\linewidth]{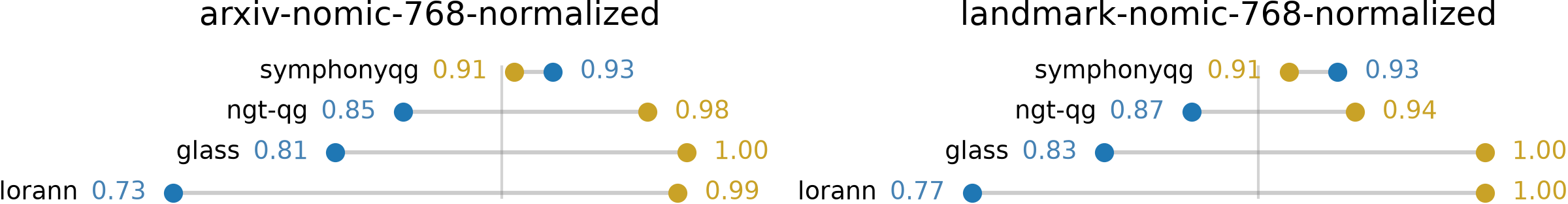}

    \caption[]{
      Average recall over the 100
      \ulc{hardest}{color_blue}
      and
      \ulc{easiest}{color_gold}
      queries (by the $\mathit{RC}_{100}$ metric) on two datasets, using each algorithm's fastest configuration achieving at least 90\% average recall over all test
      queries (threshold marked by the vertical line). SymphonyQG is robust with respect to query difficulty, while Glass and LoRANN show greater variability.
  }
    \label{fig:in-distribution-split-performance}
\end{figure}

\paragraph{Robustness to query difficulty}
The performance difference between the 100 hardest and the 100 easiest queries, as measured by relative contrast $\mathit{RC}_{100}$, is shown in Figure~\ref{fig:in-distribution-split-performance} for the four top-performing algorithms, using the fastest configurations achieving at least 90\% average recall over all test queries. Glass and LoRANN achieve nearly 100\% average recall on the easiest queries. On the hardest queries, Glass averages 81\% recall on \texttt{arxiv-nomic} and 83\% on \texttt{landmark-nomic}, while LoRANN averages 73\% and 77\%, respectively. SymphonyQG is robust, with a smaller easy--hard average-recall gap than Glass, LoRANN, and NGT-QG.

\begin{figure}[!t]
    \centering
    \includegraphics[width=0.86\linewidth]{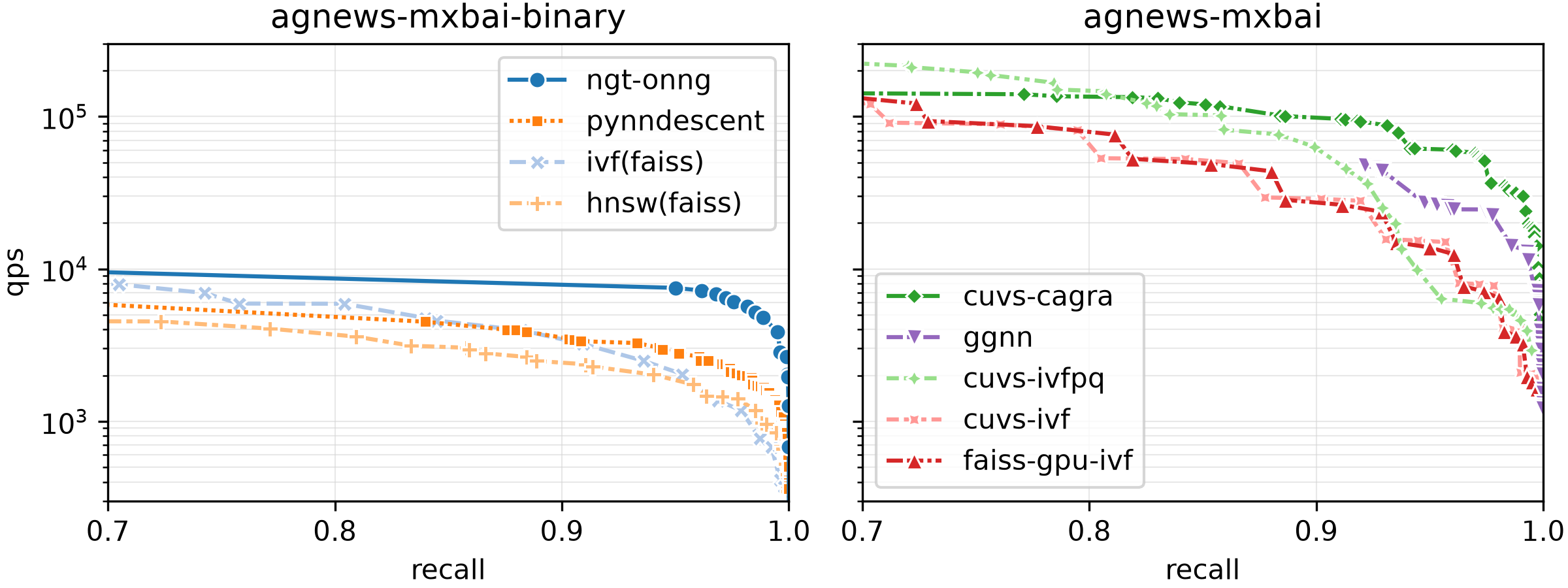}

    \caption[]{Left: recall/throughput trade-off on \textbf{binary data}. Right: recall/throughput trade-off for \textbf{GPU} algorithms. The graph-based NGT-ONNG has the highest throughput on the binary data, and the graph-based CAGRA is the fastest method in the GPU setting.
    }
    \label{fig:binary-and-GPU}
\end{figure}

\paragraph{Higher throughput with binary datasets and GPU deployment}
\textsc{VIBE} includes support for binary datasets (with Hamming distance) and GPU algorithms (see Section~\ref{sec:research_support}). Both settings can provide higher throughput. For example, on the float \texttt{agnews-mxbai} dataset at 90\% average recall, the best-performing method achieves a throughput of $4 \times 10^3$ QPS (Figure~\ref{fig:in-distribution-text}), whereas on the binarized version of the same dataset at 90\% average recall\,\footnote{However, note that for binary datasets, 90\% average recall means recovering 90\% of the exact Hamming nearest neighbors, so the recall levels are not directly comparable.}, the best-performing method achieves a throughput of $8 \times 10^3$ QPS (Figure~\ref{fig:binary-and-GPU}, left panel).

Using an NVIDIA V100 GPU, the best GPU algorithm achieves a throughput of $10^5$ QPS when all 1000 test queries are processed in a single batch (Figure~\ref{fig:binary-and-GPU}, right panel). More results are shown in Appendix~\ref{app:binary-and-GPU}: on binary datasets, the graph-based method NGT-ONNG~\citep{iwasaki2018optimization} is the fastest method, while the graph-based CAGRA~\citep{ootomo2024cagra} is the fastest GPU method. 

\subsection{Out-of-distribution setting}
\label{sec:OOD_setting}

\begin{figure}[!t]
    \centering
    \begin{minipage}{0.82\linewidth}
    \includegraphics[width=\linewidth]{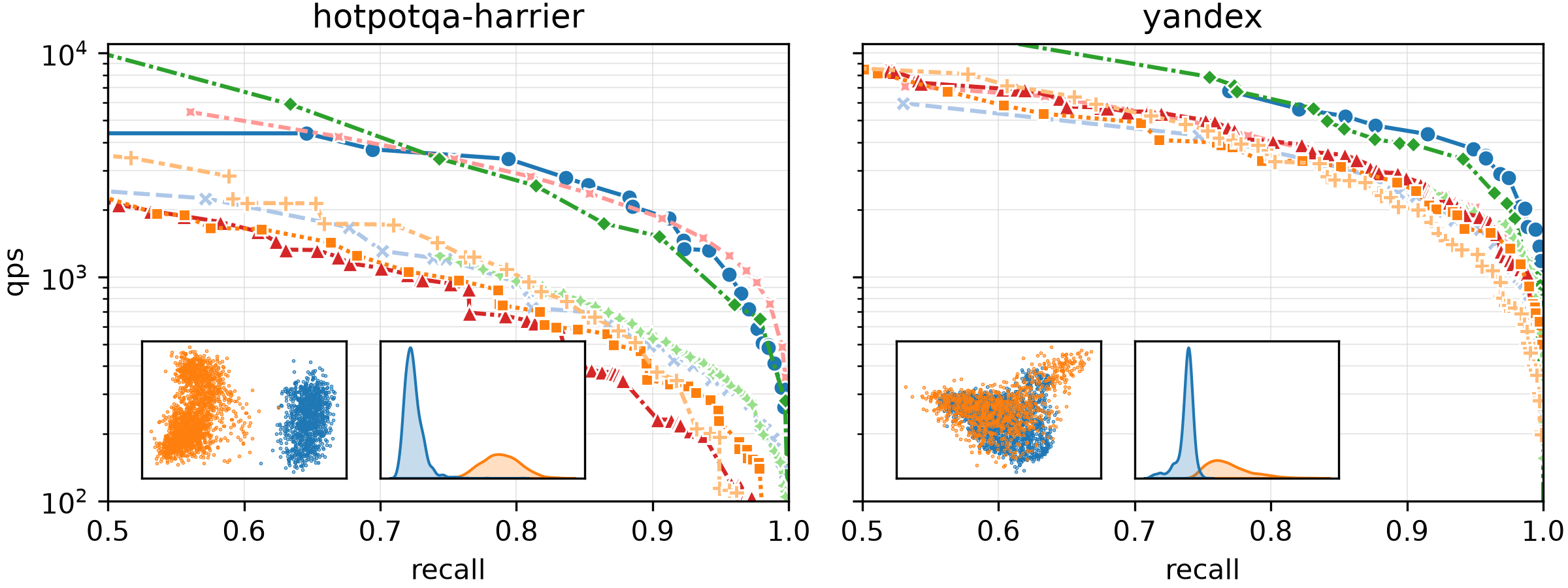}
    \end{minipage}
    \begin{minipage}{0.17\linewidth}
    \includegraphics[width=\linewidth]{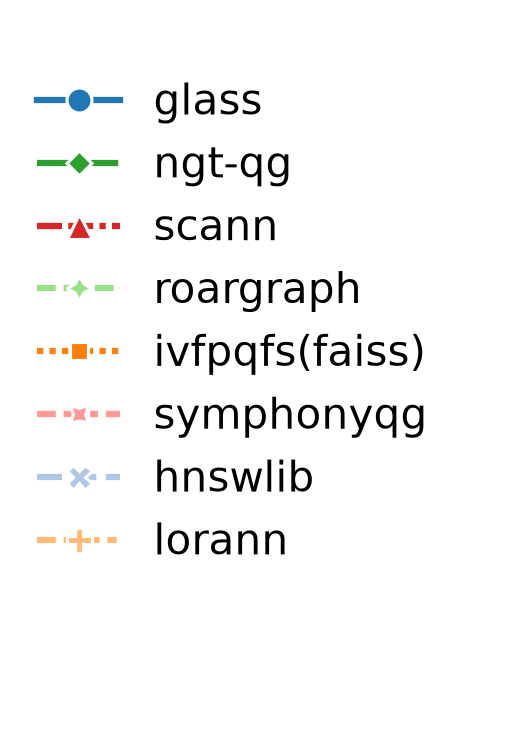}
    \end{minipage}
    \caption[]{Recall/throughput trade-off on the \textbf{text-retrieval} \texttt{hotpotqa} dataset (left) and the \textbf{text-to-image} \texttt{yandex} dataset (right), both with \textbf{out-of-distribution} queries. The left insets show PCA projections of the
    \ulc{corpus}{color_blue} points and the \ulc{queries}{color_orange}, and the right insets show the Mahalanobis-distance distributions for the corpus and query points. Less PCA overlap and a larger rightward shift of the query distance distribution indicate a stronger distribution shift. At high average recall, the general-purpose graph-based methods Glass, NGT-QG, and SymphonyQG achieve the highest throughput among the methods shown on both datasets, outperforming the evaluated OOD-specific methods, including RoarGraph.}
    \label{fig:OOD-text-to-image}
\end{figure}

\begin{figure}[!t]
    {\centering
    \begin{minipage}{0.82\linewidth}
    \includegraphics[width=\linewidth]{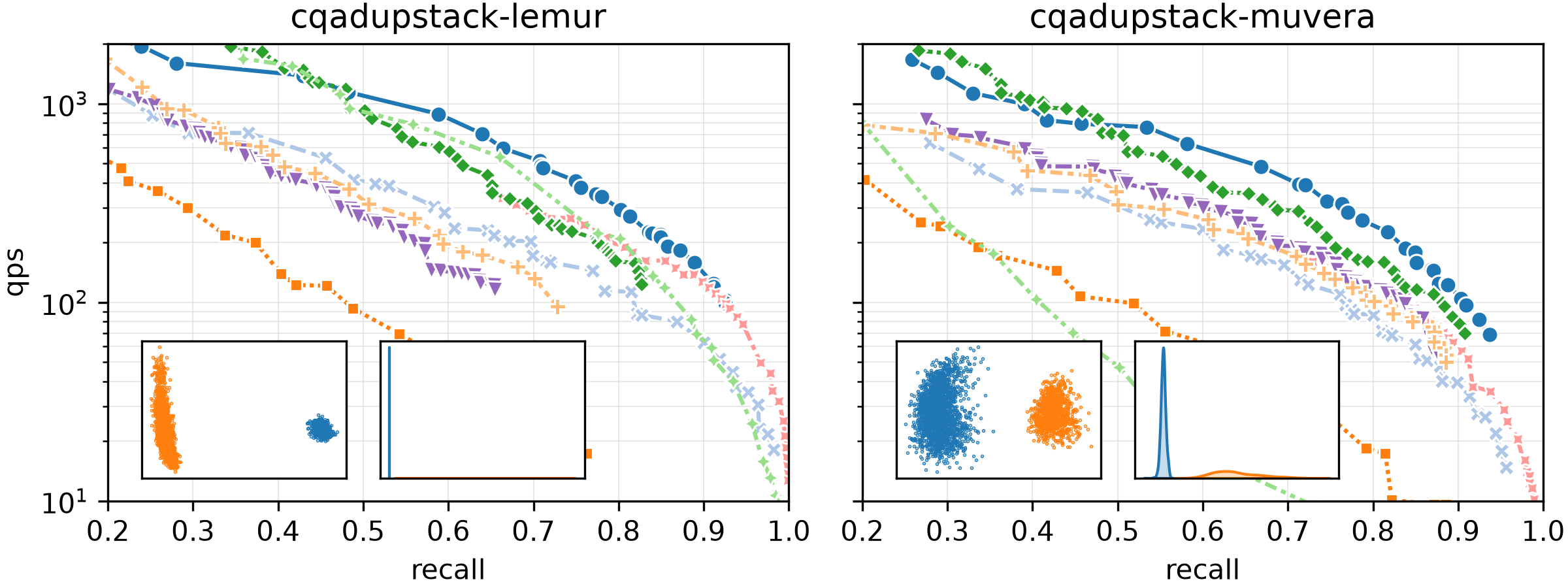}
    \end{minipage}
    \begin{minipage}{0.17\linewidth}
    \includegraphics[width=\linewidth]{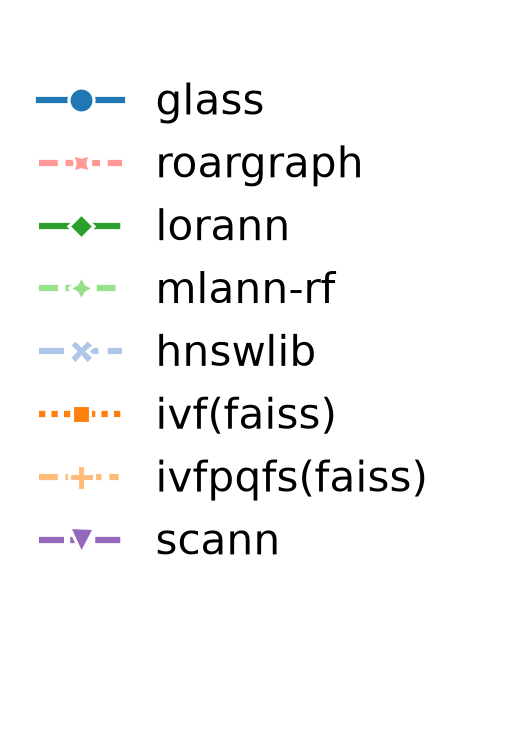}
    \end{minipage}
    \par}
    \caption{Recall/throughput trade-off on the \textbf{multi-vector-to-single-vector reduction} MIPS datasets \texttt{cqadupstack-lemur} (left) and \texttt{cqadupstack-muvera} (right), both with \textbf{out-of-distribution} queries. Methods generally need to scan substantially more candidates on the multi-vector reduction datasets than on the regular datasets, and many methods struggle to reach the highest recall levels with the common hyperparameter settings used across datasets.}
    \label{fig:OOD-multivector-reduction}
\end{figure}

\begin{figure}[!t]
    {\centering
    \begin{minipage}{0.82\linewidth}
    \includegraphics[width=\linewidth]{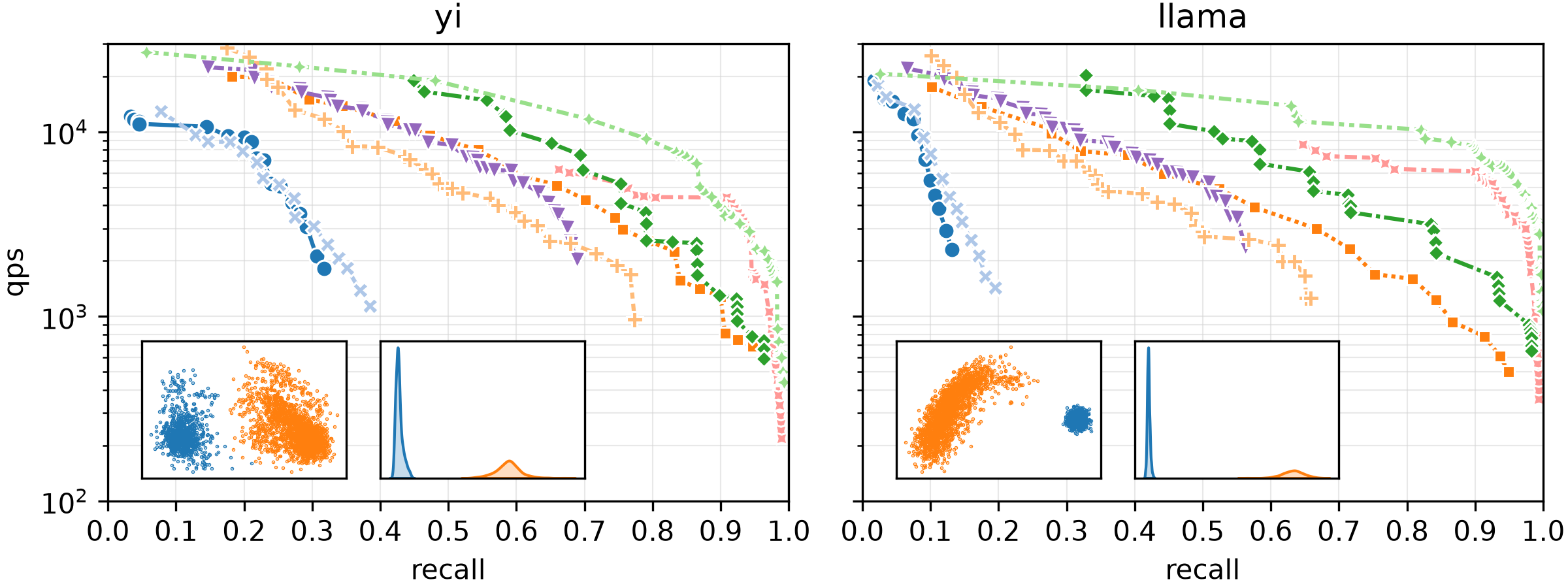}
    \end{minipage}
    \begin{minipage}{0.17\linewidth}
    \includegraphics[width=\linewidth]{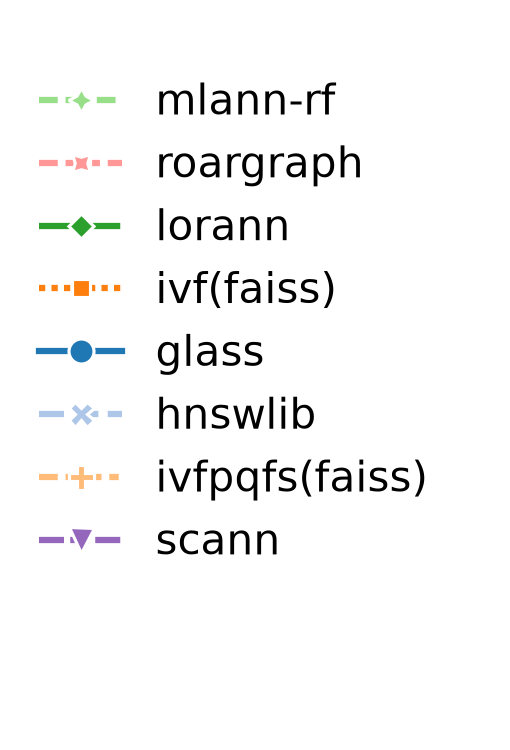}
    \end{minipage}
    \par}
    \caption{Recall/throughput trade-off on the \textbf{approximate attention computation} datasets \texttt{yi} (left) and \texttt{llama} (right), both with \textbf{out-of-distribution} queries. The specialized OOD methods MLANN (Random Forest), RoarGraph, and LoRANN outperform the regular ANN methods, such as Glass, that struggle to reach the highest recall levels.}
    \label{fig:OOD-approximate-attention-computation}
\end{figure}

We compare ANN algorithms on eight out-of-distribution (OOD) datasets. Figure~\ref{fig:results-overview} presents an overview of selected results at 95\% average recall. The results for one text-retrieval dataset and one text-to-image dataset are shown in Figure~\ref{fig:OOD-text-to-image}. Representative results for the two MIPS applications are shown in Figures~\ref{fig:OOD-multivector-reduction} and~\ref{fig:OOD-approximate-attention-computation}.

\begin{figure}[!t]
    \centering
    \includegraphics[width=0.95\linewidth]{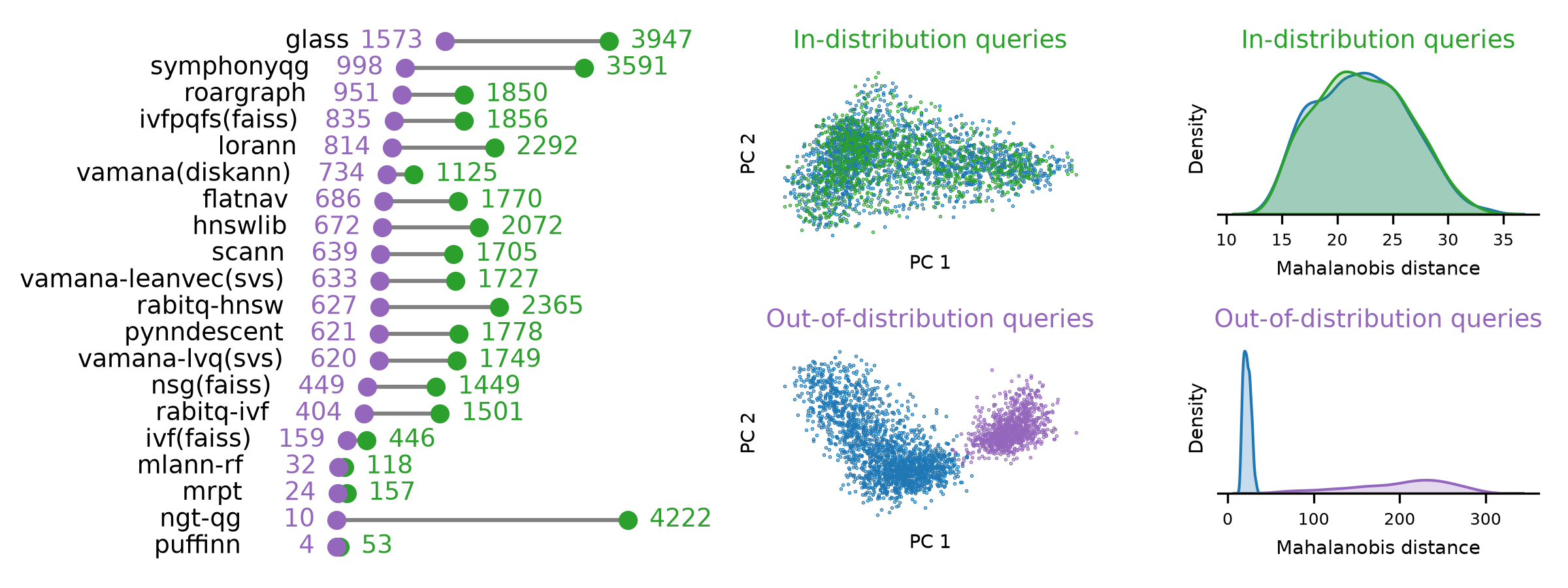}
    \caption[]{
    \emph{Left:} Throughput (QPS) for \ulc{out-of-distribution}{color_purple} and \ulc{in-distribution}{color_green} queries on the \texttt{laion-clip} dataset at 95\% average recall; \emph{middle:} PCA projections comparing the \ulc{corpus}{color_blue} with \ulc{in-distribution}{color_green} queries (top) and \ulc{out-of-distribution}{color_purple} queries (bottom); \emph{right:} the corresponding Mahalanobis distance distributions. Throughput degrades on the OOD queries for all methods shown.
    }
    \label{fig:performance-gap}
\end{figure}

\paragraph{Performance on text and text-to-image retrieval} On the text-retrieval and text-to-image OOD datasets, the most efficient general-purpose ANN methods outperform the ANN methods specifically tailored for OOD data (see Figure~\ref{fig:OOD-text-to-image}). The observed advantage of general-purpose methods may partly reflect differences in implementation maturity and optimization rather than index design alone~\citep{kriegel2017black}. However, as can be seen from Figure~\ref{fig:performance-gap}, the performance of the general-purpose ANN algorithms degrades significantly on OOD queries compared to in-distribution queries (see Appendix~\ref{app:additional_ood} for additional results). 

With in-distribution queries, a query typically lies near the indexed data cloud, and its nearest neighbors tend to be close together. An OOD query can instead be far from the corpus distribution, and its nearest corpus points can be far from both the query and one another. Reaching the same recall then requires more partition probes or longer graph traversal. The extent of the slowdown varies across methods because different routing and pruning rules respond differently to distribution shift, while approximate distance estimation schemes can lose accuracy by varying amounts. For example, if a quantizer is fitted to the corpus, for shifted queries, the quantized vectors can produce larger errors in the estimated candidate distances, steering graph traversal along less useful paths and increasing the number of examined candidates~\citep{gou2025symphonyqg}.

\paragraph{Multi-vector reduction} Figure~\ref{fig:OOD-multivector-reduction} shows results for the \texttt{lemur} and \texttt{muvera} workloads. Compared with the text-to-image workloads, methods generally need to scan substantially more candidates on these datasets, resulting in much slower search, and many methods struggle to reach the highest recall levels with the common hyperparameter settings used across datasets. RoarGraph reaches the highest recall levels and outperforms Glass on \texttt{lemur}, but still underperforms Glass on \texttt{muvera}. The OOD method MLANN is competitive on \texttt{lemur}, but substantially underperforms the other methods on \texttt{muvera}.

\paragraph{Approximate attention computation} General-purpose ANN methods struggle on the approximate attention computation datasets (see Figure~\ref{fig:OOD-approximate-attention-computation} for the \texttt{yi} and \texttt{llama} workloads): for example, the graph methods Glass and hnswlib do not reach 50\% average recall under any evaluated configuration. Clustering-based methods perform relatively better, but ScaNN and IVF-PQ perform worse than IVF. This is consistent with the results on the text-to-image datasets: their corpus-trained quantizers may distort query--candidate scores more severely under more pronounced distribution shift. In contrast, methods that are specifically designed for OOD data, namely RoarGraph, MLANN (Random Forest), and LoRANN, reach the highest recall levels and achieve the best performance.\footnote{Recall may not be the most appropriate quality metric for approximate attention computation because the relevant objective is the error in the resulting attention output rather than exact recovery of the top-$k$ keys~\citep{chen2025magicpig}. We report recall here to maintain consistency with the other ANN workloads.} The performance difference between the OOD methods and the general-purpose ANN methods is more pronounced on the \texttt{llama} dataset than on the \texttt{yi} dataset. The difference between the corpus and query distributions is also larger on \texttt{llama} than on \texttt{yi} (see Appendix~\ref{app:distribution_shift}).

\subsection{Significance tests}

\begin{figure}[!htbp]
    \centering
    \includegraphics[width=0.88\linewidth]{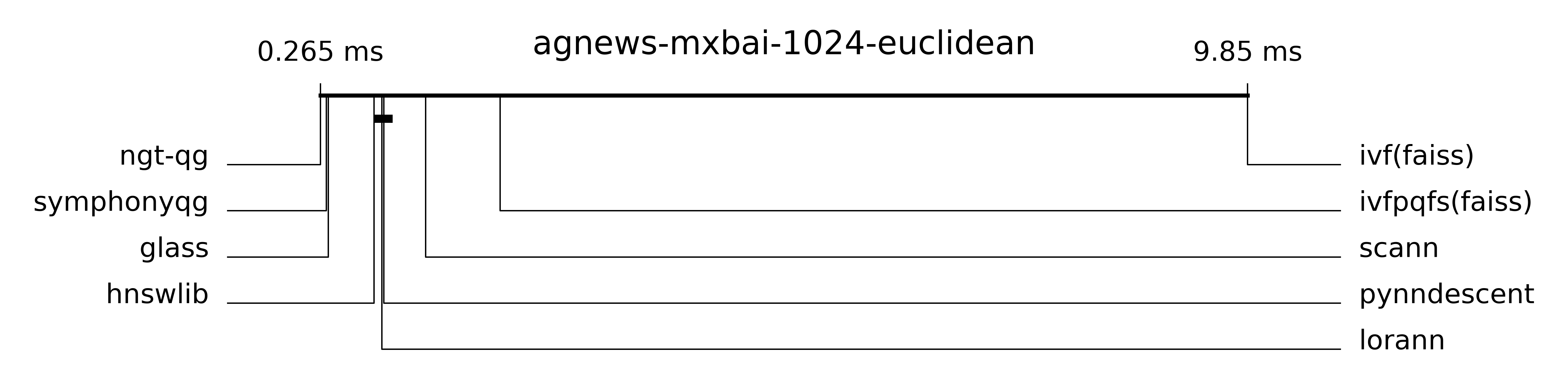}
    \includegraphics[width=0.88\linewidth]{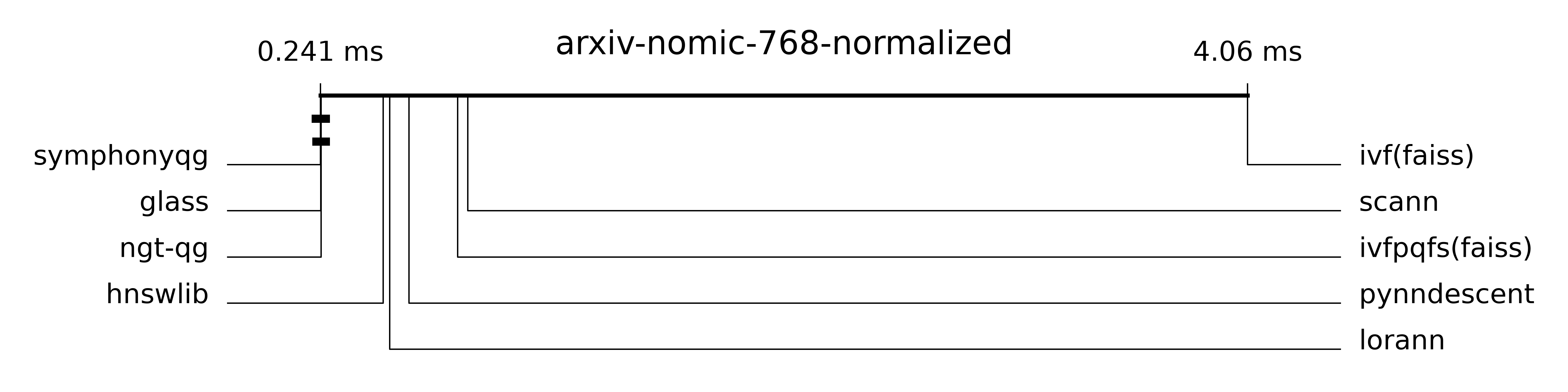}
    \caption[]{Critical difference diagrams visualizing the results of the significance tests for the \texttt{agnews-mxbai} and \texttt{arxiv-nomic} datasets. The bold horizontal black bars connect the algorithms whose paired per-query latency differences are not statistically significant at the level $\alpha = 0.01$ (with Holm-Bonferroni correction).
    }
    \label{fig:critical_difference}
\end{figure}

To assess the robustness of our results, for each dataset we select each algorithm's fastest configuration reaching at least 95\% average recall and compare algorithms using their paired per-query latencies over the 1000 test queries. We test all pairwise differences using Wilcoxon signed rank tests~\citep{wilcoxon1945individual} with the Holm-Bonferroni correction~\citep{holm1979simple}. Almost all (2399 out of 2450 pairwise comparisons\,\footnote{There are in total 2450 possible pairwise comparisons when we add up all the comparisons for both the in-distribution and the out-of-distribution datasets. We apply the Holm-Bonferroni correction over all of these 2450 comparisons.} of algorithms reaching 95\% average recall) of the differences are statistically significant at the significance level $\alpha = 0.01$. The critical difference diagrams~\citep{demvsar2006statistical} visualizing the results of the significance tests for \texttt{agnews-mxbai} and \texttt{arxiv-nomic} can be found in Figure \ref{fig:critical_difference}.

\section{Discussion}
\label{sec:discussion}

We introduce \textsc{VIBE}, a benchmark for evaluating ANN algorithms on modern embedding workloads with both in-distribution and OOD queries. Our findings demonstrate the importance of evaluating ANN algorithms across diverse, representative embedding workloads.

\subsection{Key findings}
\label{sec:key_findings}

\paragraph{Comparison across method classes} The graph-based methods Glass and SymphonyQG are the top-performing implementations. SymphonyQG is also the most robust method with respect to query difficulty (see Figure~\ref{fig:in-distribution-split-performance}). Graph-based and clustering-based indexes outperform both tree-based and hashing-based indexes.

\paragraph{Reduced precision} All of the top-performing graph-based methods (SymphonyQG, Glass, NGT-QG) and clustering-based methods (LoRANN, ScaNN, Faiss-IVF-PQ) use different types of quantization in at least some of their computations. SymphonyQG uses RaBitQ~\citep{gao2024rabitq}, while NGT-QG, ScaNN, and Faiss-IVF-PQ employ efficient 4-bit PQ implementations~\citep{andre2017accelerated}. LoRANN uses 8-bit scalar quantization to speed up vector-matrix multiplications, and Glass, an efficient HNSW implementation with scalar quantization, significantly outperforms the full-precision HNSW implementation.

\paragraph{OOD performance gap} In the out-of-distribution (OOD) setting, two key trends emerge: (1) On the approximate attention computation MIPS datasets, where the difference between the corpus and query distributions is the largest, general-purpose ANN methods struggle to reach the highest recall levels, and methods specifically tailored for the OOD setting (MLANN, RoarGraph, LoRANN) outperform them. (2) However, on the text-retrieval and text-to-image OOD datasets, our results suggest that while the performance of the general-purpose ANN methods degrades in the OOD setting, the current OOD-specific methods do not yield significant performance improvements over general-purpose methods.

\subsection{Limitations and future work}
\label{sec:limitations}

We do not record internal algorithm statistics, such as the number of corpus points accessed or distance computations performed, since most implementations do not expose such counters. Although \textsc{VIBE} measures index size, we do not report it here because implementations differ in whether they copy corpus vectors into the index, making the results incomparable.

All experiments in the paper are conducted on a single hardware platform. Because performance can depend on the interaction between CPU microarchitecture and index access patterns~\citep{kuffo2025bang}, relative implementation performance may differ on other hardware. Repeating the evaluation across multiple hardware architectures would therefore be valuable for assessing the generalizability of our findings.

Due to the high dimensionality of embeddings, we focus on million-scale datasets. This scale keeps the computational requirements manageable while allowing us to compare many algorithms across a broad range of embeddings.
However, evaluating the performance of ANN algorithms on larger, even billion-scale, datasets to determine whether our key findings also hold in this regime is an interesting direction for future research.

\impact{The article is foundational research that focuses on evaluating ANN search algorithms. ANN search is a primitive that is used as a component in machine learning pipelines. Measuring the computational cost associated with the choice of algorithm has the potential to reduce the carbon emissions of large-scale deployments. Beyond this potential positive environmental impact, our work has no direct societal impact.}

\acks{This work has been supported by the Research Council of Finland (grant \#361902 and the Flagship programme: Finnish Center for Artificial Intelligence FCAI) and the Jane and Aatos Erkko Foundation (BioDesign project, grant \#7001702). Martin Aumüller received funding from the Innovation Fund Denmark for the project DIREC (9142-00001B). The authors
acknowledge the research environment provided by ELLIS
Institute Finland, as well as CSC – IT Center for Science,
Finland, for computational resources.}

\vskip 0.2in
\bibliography{refs}

\begin{thebibliography}{110}
\providecommand{\natexlab}[1]{#1}
\providecommand{\url}[1]{\texttt{#1}}
\expandafter\ifx\csname urlstyle\endcsname\relax
  \providecommand{\doi}[1]{doi: #1}\else
  \providecommand{\doi}{doi: \begingroup \urlstyle{rm}\Url}\fi

\bibitem[Aguerrebere et~al.(2023)Aguerrebere, Bhati, Hildebrand, Tepper, and
  Willke]{aguerrebere2023similarity}
Cecilia Aguerrebere, Ishwar~Singh Bhati, Mark Hildebrand, Mariano Tepper, and
  Theodore~L. Willke.
\newblock Similarity search in the blink of an eye with compressed indices.
\newblock \emph{Proceedings of the VLDB Endowment}, 16\penalty0 (11):\penalty0
  3433--3446, 2023.

\bibitem[Akram et~al.(2026)Akram, Sturua, Havriushenko, Herreros, G{\"u}nther,
  Werk, and Xiao]{akram2026jina}
Mohammad~Kalim Akram, Saba Sturua, Nastia Havriushenko, Quentin Herreros,
  Michael G{\"u}nther, Maximilian Werk, and Han Xiao.
\newblock jina-embeddings-v5-text: Task-targeted embedding distillation.
\newblock \emph{arXiv preprint arXiv:2602.15547}, 2026.

\bibitem[Andr{\'e} et~al.(2017)Andr{\'e}, Kermarrec, and
  Le~Scouarnec]{andre2017accelerated}
Fabien Andr{\'e}, Anne-Marie Kermarrec, and Nicolas Le~Scouarnec.
\newblock Accelerated nearest neighbor search with {Quick ADC}.
\newblock In \emph{Proceedings of the 2017 ACM International Conference on
  Multimedia Retrieval}, pages 159--166, 2017.

\bibitem[Aum{\"{u}}ller and Ceccarello(2021)]{DBLP:journals/is/AumullerC21}
Martin Aum{\"{u}}ller and Matteo Ceccarello.
\newblock The role of local dimensionality measures in benchmarking nearest
  neighbor search.
\newblock \emph{Information Systems}, 101:\penalty0 101807, 2021.

\bibitem[Aum{\"{u}}ller and Ceccarello(2023)]{DBLP:journals/debu/AumullerC23}
Martin Aum{\"{u}}ller and Matteo Ceccarello.
\newblock Recent approaches and trends in approximate nearest neighbor search,
  with remarks on benchmarking.
\newblock \emph{{IEEE} Data Engineering Bulletin}, 47\penalty0 (3):\penalty0
  89--105, 2023.

\bibitem[Aum{\"u}ller et~al.(2019)Aum{\"u}ller, Christiani, Pagh, and
  Vesterli]{aumuller2019puffinn}
Martin Aum{\"u}ller, Tobias Christiani, Rasmus Pagh, and Michael Vesterli.
\newblock {PUFFINN}: Parameterless and universally fast finding of nearest
  neighbors.
\newblock In \emph{27th Annual European Symposium on Algorithms (ESA 2019)},
  volume 144 of \emph{Leibniz International Proceedings in Informatics
  (LIPIcs)}, pages 10:1--10:16, 2019.

\bibitem[Aum{\"u}ller et~al.(2020)Aum{\"u}ller, Bernhardsson, and
  Faithfull]{aumuller2020ann}
Martin Aum{\"u}ller, Erik Bernhardsson, and Alexander Faithfull.
\newblock {ANN-Benchmarks}: A benchmarking tool for approximate nearest
  neighbor algorithms.
\newblock \emph{Information Systems}, 87:\penalty0 101374, 2020.

\bibitem[Azizi et~al.(2025)Azizi, Echihabi, and Palpanas]{azizi2025graph}
Ilias Azizi, Karima Echihabi, and Themis Palpanas.
\newblock Graph-based vector search: An experimental evaluation of the
  state-of-the-art.
\newblock \emph{Proceedings of the ACM on Management of Data}, 3\penalty0
  (1):\penalty0 1--31, 2025.

\bibitem[Bajaj et~al.(2016)Bajaj, Campos, Craswell, Deng, Gao, Liu, Majumder,
  McNamara, Mitra, Nguyen, Rosenberg, Song, Stoica, Tiwary, and
  Wang]{bajaj2016ms}
Payal Bajaj, Daniel Campos, Nick Craswell, Li~Deng, Jianfeng Gao, Xiaodong Liu,
  Rangan Majumder, Andrew McNamara, Bhaskar Mitra, Tri Nguyen, Mir Rosenberg,
  Xia Song, Alina Stoica, Saurabh Tiwary, and Tong Wang.
\newblock {MS MARCO}: A human generated machine reading comprehension dataset.
\newblock \emph{arXiv preprint arXiv:1611.09268}, 2016.

\bibitem[Bertsch et~al.(2023)Bertsch, Alon, Neubig, and
  Gormley]{bertsch2023unlimiformer}
Amanda Bertsch, Uri Alon, Graham Neubig, and Matthew~R. Gormley.
\newblock Unlimiformer: Long-range transformers with unlimited length input.
\newblock \emph{Advances in Neural Information Processing Systems},
  36:\penalty0 35522--35543, 2023.

\bibitem[Borgeaud et~al.(2022)Borgeaud, Mensch, Hoffmann, Cai, Rutherford,
  Millican, Van Den~Driessche, Lespiau, Damoc, Clark, De~Las~Casas, Guy,
  Menick, Ring, Hennigan, Huang, Maggiore, Jones, Cassirer, Brock, Paganini,
  Irving, Vinyals, Osindero, Simonyan, Rae, Elsen, and
  Sifre]{borgeaud2022improving}
Sebastian Borgeaud, Arthur Mensch, Jordan Hoffmann, Trevor Cai, Eliza
  Rutherford, Katie Millican, George~Bm Van Den~Driessche, Jean-Baptiste
  Lespiau, Bogdan Damoc, Aidan Clark, Diego De~Las~Casas, Aurelia Guy, Jacob
  Menick, Roman Ring, Tom Hennigan, Saffron Huang, Loren Maggiore, Chris Jones,
  Albin Cassirer, Andy Brock, Michela Paganini, Geoffrey Irving, Oriol Vinyals,
  Simon Osindero, Karen Simonyan, Jack Rae, Erich Elsen, and Laurent Sifre.
\newblock Improving language models by retrieving from trillions of tokens.
\newblock In \emph{Proceedings of the 39th International Conference on Machine
  Learning}, volume 162 of \emph{Proceedings of Machine Learning Research},
  pages 2206--2240, 2022.

\bibitem[Bruch(2024)]{bruch2024foundations}
Sebastian Bruch.
\newblock \emph{Foundations of vector retrieval}.
\newblock Springer, 2024.

\bibitem[Caron et~al.(2021)Caron, Touvron, Misra, J{\'e}gou, Mairal,
  Bojanowski, and Joulin]{caron2021emerging}
Mathilde Caron, Hugo Touvron, Ishan Misra, Herv{\'e} J{\'e}gou, Julien Mairal,
  Piotr Bojanowski, and Armand Joulin.
\newblock Emerging properties in self-supervised vision transformers.
\newblock In \emph{Proceedings of the IEEE/CVF International Conference on
  Computer Vision}, pages 9650--9660, 2021.

\bibitem[Cayton and Dasgupta(2007)]{cayton2007learning}
Lawrence Cayton and Sanjoy Dasgupta.
\newblock A learning framework for nearest neighbor search.
\newblock \emph{Advances in Neural Information Processing Systems},
  20:\penalty0 233--240, 2007.

\bibitem[Ceccarello et~al.(2025)Ceccarello, Levchenko, Ileana, and
  Palpanas]{ceccarello2025evaluating}
Matteo Ceccarello, Alexandra Levchenko, Ioana Ileana, and Themis Palpanas.
\newblock Evaluating and generating query workloads for high dimensional vector
  similarity search.
\newblock In \emph{Proceedings of the 31st ACM SIGKDD Conference on Knowledge
  Discovery and Data Mining V.2}, pages 5299--5310, 2025.

\bibitem[Chen et~al.(2024)Chen, Zhang, He, Jing, and Wang]{chen2024roargraph}
Meng Chen, Kai Zhang, Zhenying He, Yinan Jing, and X.~Sean Wang.
\newblock {RoarGraph}: A projected bipartite graph for efficient cross-modal
  approximate nearest neighbor search.
\newblock \emph{Proceedings of the VLDB Endowment}, 17\penalty0 (11):\penalty0
  2735--2749, 2024.

\bibitem[Chen et~al.(2026)Chen, Fu, Wu, Wu, Fan, Ke, Gao, Ni, and
  Zeng]{chen2026reveal}
Tingyang Chen, Cong Fu, Jiahua Wu, Haotian Wu, Hua Fan, Xiangyu Ke, Yunjun Gao,
  Yabo Ni, and Anxiang Zeng.
\newblock Reveal hidden pitfalls and navigate next generation of vector
  similarity search from task-centric views: [experiments \& analysis].
\newblock \emph{Proceedings of the ACM on Management of Data}, 4\penalty0
  (1):\penalty0 1--25, 2026.

\bibitem[Chen et~al.(2025)Chen, Sadhukhan, Ye, Zhou, Zhang, Nolte, Tian, Douze,
  Bottou, Jia, and Chen]{chen2025magicpig}
Zhuoming Chen, Ranajoy Sadhukhan, Zihao Ye, Yang Zhou, Jianyu Zhang, Niklas
  Nolte, Yuandong Tian, Matthijs Douze, Leon Bottou, Zhihao Jia, and Beidi
  Chen.
\newblock {MagicPIG}: {LSH} sampling for efficient {LLM} generation.
\newblock In \emph{International Conference on Learning Representations}, 2025.

\bibitem[Covington et~al.(2016)Covington, Adams, and Sargin]{covington2016deep}
Paul Covington, Jay Adams, and Emre Sargin.
\newblock Deep neural networks for {YouTube} recommendations.
\newblock In \emph{Proceedings of the 10th ACM Conference on Recommender
  Systems}, pages 191--198, 2016.

\bibitem[Dasgupta and Freund(2008)]{dasgupta2008random}
Sanjoy Dasgupta and Yoav Freund.
\newblock Random projection trees and low dimensional manifolds.
\newblock In \emph{Proceedings of the Fortieth Annual ACM Symposium on Theory
  of Computing}, pages 537--546, 2008.

\bibitem[Dem{\v{s}}ar(2006)]{demvsar2006statistical}
Janez Dem{\v{s}}ar.
\newblock Statistical comparisons of classifiers over multiple data sets.
\newblock \emph{Journal of Machine Learning Research}, 7:\penalty0 1--30, 2006.

\bibitem[Deng et~al.(2009)Deng, Dong, Socher, Li, Li, and
  Fei-Fei]{deng2009imagenet}
Jia Deng, Wei Dong, Richard Socher, Li-Jia Li, Kai Li, and Li~Fei-Fei.
\newblock {ImageNet}: A large-scale hierarchical image database.
\newblock In \emph{2009 IEEE Conference on Computer Vision and Pattern
  Recognition}, pages 248--255. IEEE, 2009.

\bibitem[Dhulipala et~al.(2024)Dhulipala, Hadian, Jayaram, Lee, and
  Mirrokni]{jayaram2024muvera}
Laxman Dhulipala, Majid Hadian, Rajesh Jayaram, Jason Lee, and Vahab Mirrokni.
\newblock {MUVERA}: Multi-vector retrieval via fixed dimensional encoding.
\newblock \emph{Advances in Neural Information Processing Systems},
  37:\penalty0 101042--101073, 2024.

\bibitem[Dong et~al.(2011)Dong, Charikar, and Li]{wei2021pynndescent}
Wei Dong, Moses Charikar, and Kai Li.
\newblock Efficient $k$-nearest neighbor graph construction for generic
  similarity measures.
\newblock In \emph{Proceedings of the 20th International Conference on World
  Wide Web}, pages 577--586, 2011.

\bibitem[Douze et~al.(2026)Douze, Guzhva, Deng, Johnson, Szilvasy, Mazar{\'e},
  Lomeli, Hosseini, and J{\'e}gou]{douze2024faiss}
Matthijs Douze, Alexandr Guzhva, Chengqi Deng, Jeff Johnson, Gergely Szilvasy,
  Pierre-Emmanuel Mazar{\'e}, Maria Lomeli, Lucas Hosseini, and Herv{\'e}
  J{\'e}gou.
\newblock The {Faiss} library.
\newblock \emph{IEEE Transactions on Big Data}, 12\penalty0 (2):\penalty0
  346--361, 2026.

\bibitem[Dowson and Landau(1982)]{dowson1982frechet}
D.~C. Dowson and B.~V. Landau.
\newblock The {Fr{\'e}chet} distance between multivariate normal distributions.
\newblock \emph{Journal of Multivariate Analysis}, 12\penalty0 (3):\penalty0
  450--455, 1982.

\bibitem[Fang et~al.(2022)Fang, Ilharco, Wortsman, Wan, Shankar, Dave, and
  Schmidt]{fang2022data}
Alex Fang, Gabriel Ilharco, Mitchell Wortsman, Yuhao Wan, Vaishaal Shankar,
  Achal Dave, and Ludwig Schmidt.
\newblock Data determines distributional robustness in contrastive language
  image pre-training ({CLIP}).
\newblock In \emph{Proceedings of the 39th International Conference on Machine
  Learning}, volume 162 of \emph{Proceedings of Machine Learning Research},
  pages 6216--6234, 2022.

\bibitem[Friedman et~al.(1977)Friedman, Bentley, and
  Finkel]{friedman1977algorithm}
Jerome~H. Friedman, Jon~Louis Bentley, and Raphael~Ari Finkel.
\newblock An algorithm for finding best matches in logarithmic expected time.
\newblock \emph{ACM Transactions on Mathematical Software}, 3\penalty0
  (3):\penalty0 209--226, 1977.

\bibitem[Fu et~al.(2019)Fu, Xiang, Wang, and Cai]{fu2017nsg}
Cong Fu, Chao Xiang, Changxu Wang, and Deng Cai.
\newblock Fast approximate nearest neighbor search with the navigating
  spreading-out graph.
\newblock \emph{Proceedings of the VLDB Endowment}, 12\penalty0 (5):\penalty0
  461--474, 2019.

\bibitem[Gao and Long(2024)]{gao2024rabitq}
Jianyang Gao and Cheng Long.
\newblock {RaBitQ}: Quantizing high-dimensional vectors with a theoretical
  error bound for approximate nearest neighbor search.
\newblock \emph{Proceedings of the ACM on Management of Data}, 2\penalty0
  (3):\penalty0 1--27, 2024.

\bibitem[Gao et~al.(2025{\natexlab{a}})Gao, Gou, Xu, Shi, Yang, and
  Long]{gao2025rabitq}
Jianyang Gao, Yutong Gou, Yuexuan Xu, Jifan Shi, Zhonghao Yang, and Cheng Long.
\newblock The {RaBitQ} library.
\newblock In \emph{Proceedings of the 1st Workshop on Vector Databases at
  International Conference on Machine Learning}, 2025{\natexlab{a}}.

\bibitem[Gao et~al.(2025{\natexlab{b}})Gao, Gou, Xu, Yang, Long, and
  Wong]{gao2025practical}
Jianyang Gao, Yutong Gou, Yuexuan Xu, Yongyi Yang, Cheng Long, and Raymond
  Chi-Wing Wong.
\newblock Practical and asymptotically optimal quantization of high-dimensional
  vectors in {Euclidean} space for approximate nearest neighbor search.
\newblock \emph{Proceedings of the ACM on Management of Data}, 3\penalty0
  (3):\penalty0 1--26, 2025{\natexlab{b}}.

\bibitem[Gou et~al.(2025)Gou, Gao, Xu, and Long]{gou2025symphonyqg}
Yutong Gou, Jianyang Gao, Yuexuan Xu, and Cheng Long.
\newblock {SymphonyQG}: Towards symphonious integration of quantization and
  graph for approximate nearest neighbor search.
\newblock \emph{Proceedings of the ACM on Management of Data}, 3\penalty0
  (1):\penalty0 1--26, 2025.

\bibitem[Groh et~al.(2023)Groh, Ruppert, Wieschollek, and Lensch]{groh2022ggnn}
Fabian Groh, Lukas Ruppert, Patrick Wieschollek, and Hendrik P.~A. Lensch.
\newblock {GGNN}: Graph-based {GPU} nearest neighbor search.
\newblock \emph{IEEE Transactions on Big Data}, 9\penalty0 (1):\penalty0
  267--279, 2023.

\bibitem[Guo et~al.(2020)Guo, Sun, Lindgren, Geng, Simcha, Chern, and
  Kumar]{guo2020accelerating}
Ruiqi Guo, Philip Sun, Erik Lindgren, Quan Geng, David Simcha, Felix Chern, and
  Sanjiv Kumar.
\newblock Accelerating large-scale inference with anisotropic vector
  quantization.
\newblock In \emph{Proceedings of the 37th International Conference on Machine
  Learning}, volume 119 of \emph{Proceedings of Machine Learning Research},
  pages 3887--3896, 2020.

\bibitem[Guu et~al.(2020)Guu, Lee, Tung, Pasupat, and Chang]{guu2020retrieval}
Kelvin Guu, Kenton Lee, Zora Tung, Panupong Pasupat, and Mingwei Chang.
\newblock {REALM}: Retrieval-augmented language model pre-training.
\newblock In \emph{Proceedings of the 37th International Conference on Machine
  Learning}, volume 119 of \emph{Proceedings of Machine Learning Research},
  pages 3929--3938, 2020.

\bibitem[He et~al.(2012)He, Kumar, and Chang]{he2012difficulty}
Junfeng He, Sanjiv Kumar, and Shih-Fu Chang.
\newblock On the difficulty of nearest neighbor search.
\newblock In \emph{Proceedings of the 29th International Conference on Machine
  Learning}, pages 1127--1134, 2012.

\bibitem[He et~al.(2016)He, Zhang, Ren, and Sun]{he2016deep}
Kaiming He, Xiangyu Zhang, Shaoqing Ren, and Jian Sun.
\newblock Deep residual learning for image recognition.
\newblock In \emph{Proceedings of the IEEE Conference on Computer Vision and
  Pattern Recognition}, pages 770--778, 2016.

\bibitem[Holm(1979)]{holm1979simple}
Sture Holm.
\newblock A simple sequentially rejective multiple test procedure.
\newblock \emph{Scandinavian Journal of Statistics}, 6\penalty0 (2):\penalty0
  65--70, 1979.

\bibitem[Hoogeveen et~al.(2015)Hoogeveen, Verspoor, and
  Baldwin]{hoogeveen2015cqadupstack}
Doris Hoogeveen, Karin~M. Verspoor, and Timothy Baldwin.
\newblock {CQADupStack}: A benchmark data set for community question-answering
  research.
\newblock In \emph{Proceedings of the 20th Australasian Document Computing
  Symposium}, pages 3:1--3:8, 2015.

\bibitem[Hyv{\"o}nen et~al.(2016)Hyv{\"o}nen, Pitk{\"a}nen, Tasoulis,
  J{\"a}{\"a}saari, Tuomainen, Wang, Corander, and Roos]{hyvonen2016fast}
Ville Hyv{\"o}nen, Teemu Pitk{\"a}nen, Sotiris Tasoulis, Elias
  J{\"a}{\"a}saari, Risto Tuomainen, Liang Wang, Jukka Corander, and Teemu
  Roos.
\newblock Fast nearest neighbor search through sparse random projections and
  voting.
\newblock In \emph{Proceedings of the 2016 IEEE International Conference on Big
  Data}, pages 881--888. IEEE, 2016.

\bibitem[Hyv{\"o}nen et~al.(2022)Hyv{\"o}nen, J{\"a}{\"a}saari, and
  Roos]{hyvonen2022multilabel}
Ville Hyv{\"o}nen, Elias J{\"a}{\"a}saari, and Teemu Roos.
\newblock A multilabel classification framework for approximate nearest
  neighbor search.
\newblock \emph{Advances in Neural Information Processing Systems},
  35:\penalty0 35741--35754, 2022.

\bibitem[Hyv{\"o}nen et~al.(2024)Hyv{\"o}nen, J{\"a}{\"a}saari, and
  Roos]{hyvonen2024multilabel}
Ville Hyv{\"o}nen, Elias J{\"a}{\"a}saari, and Teemu Roos.
\newblock A multilabel classification framework for approximate nearest
  neighbor search.
\newblock \emph{Journal of Machine Learning Research}, 25\penalty0
  (46):\penalty0 1--51, 2024.

\bibitem[Iff et~al.(2026)Iff, Br{\"u}gger, Chrapek, Kochergin, Besta, and
  Hoefler]{iff2026benchmarking}
Patrick Iff, Paul Br{\"u}gger, Marcin Chrapek, David Kochergin, Maciej Besta,
  and Torsten Hoefler.
\newblock Benchmarking filtered approximate nearest neighbor search algorithms
  on transformer-based embedding vectors.
\newblock In \emph{Proceedings of the 49th International ACM SIGIR Conference
  on Research and Development in Information Retrieval}, pages 610--622, 2026.

\bibitem[Indyk and Motwani(1998)]{indyk1998approximate}
Piotr Indyk and Rajeev Motwani.
\newblock Approximate nearest neighbors: Towards removing the curse of
  dimensionality.
\newblock In \emph{Proceedings of the Thirtieth Annual ACM Symposium on Theory
  of Computing}, pages 604--613, 1998.

\bibitem[Iwasaki and Miyazaki(2018)]{iwasaki2018optimization}
Masajiro Iwasaki and Daisuke Miyazaki.
\newblock Optimization of indexing based on $k$-nearest neighbor graph for
  proximity search in high-dimensional data.
\newblock \emph{arXiv preprint arXiv:1810.07355}, 2018.

\bibitem[Izacard et~al.(2022)Izacard, Caron, Hosseini, Riedel, Bojanowski,
  Joulin, and Grave]{izacard2022unsupervised}
Gautier Izacard, Mathilde Caron, Lucas Hosseini, Sebastian Riedel, Piotr
  Bojanowski, Armand Joulin, and Edouard Grave.
\newblock Unsupervised dense information retrieval with contrastive learning.
\newblock \emph{Transactions on Machine Learning Research}, 2022.

\bibitem[J{\"a}{\"a}saari et~al.(2019)J{\"a}{\"a}saari, Hyv{\"o}nen, and
  Roos]{jaasaari2019efficient}
Elias J{\"a}{\"a}saari, Ville Hyv{\"o}nen, and Teemu Roos.
\newblock Efficient autotuning of hyperparameters in approximate nearest
  neighbor search.
\newblock In \emph{Pacific-Asia Conference on Knowledge Discovery and Data
  Mining}, pages 590--602. Springer, 2019.

\bibitem[J{\"a}{\"a}saari et~al.(2024)J{\"a}{\"a}saari, Hyv{\"o}nen, and
  Roos]{jaasaari2024lorann}
Elias J{\"a}{\"a}saari, Ville Hyv{\"o}nen, and Teemu Roos.
\newblock {LoRANN}: Low-rank matrix factorization for approximate nearest
  neighbor search.
\newblock \emph{Advances in Neural Information Processing Systems},
  37:\penalty0 102121--102153, 2024.

\bibitem[J{\"a}{\"a}saari et~al.(2026)J{\"a}{\"a}saari, Hyv{\"o}nen, and
  Roos]{jaasaari2026lemur}
Elias J{\"a}{\"a}saari, Ville Hyv{\"o}nen, and Teemu Roos.
\newblock {LEMUR}: Learned multi-vector retrieval.
\newblock In \emph{Proceedings of the 43rd International Conference on Machine
  Learning}, volume 306 of \emph{Proceedings of Machine Learning Research},
  2026.

\bibitem[Jaiswal et~al.(2022)Jaiswal, Krishnaswamy, Garg, Simhadri, and
  Agrawal]{jaiswal2022ood}
Shikhar Jaiswal, Ravishankar Krishnaswamy, Ankit Garg, Harsha~Vardhan Simhadri,
  and Sheshansh Agrawal.
\newblock {OOD-DiskANN}: Efficient and scalable graph {ANNS} for
  out-of-distribution queries.
\newblock \emph{arXiv preprint arXiv:2211.12850}, 2022.

\bibitem[J{\'e}gou et~al.(2011)J{\'e}gou, Douze, and Schmid]{jegou2011product}
Herv{\'e} J{\'e}gou, Matthijs Douze, and Cordelia Schmid.
\newblock Product quantization for nearest neighbor search.
\newblock \emph{IEEE Transactions on Pattern Analysis and Machine
  Intelligence}, 33\penalty0 (1):\penalty0 117--128, 2011.

\bibitem[Jia et~al.(2021)Jia, Yang, Xia, Chen, Parekh, Pham, Le, Sung, Li, and
  Duerig]{jia2021scaling}
Chao Jia, Yinfei Yang, Ye~Xia, Yi-Ting Chen, Zarana Parekh, Hieu Pham, Quoc~V.
  Le, Yun-Hsuan Sung, Zhen Li, and Tom Duerig.
\newblock Scaling up visual and vision-language representation learning with
  noisy text supervision.
\newblock In \emph{Proceedings of the 38th International Conference on Machine
  Learning}, volume 139 of \emph{Proceedings of Machine Learning Research},
  pages 4904--4916, 2021.

\bibitem[Johnson et~al.(2021)Johnson, Douze, and J{\'e}gou]{johnson2019billion}
Jeff Johnson, Matthijs Douze, and Herv{\'e} J{\'e}gou.
\newblock Billion-scale similarity search with {GPUs}.
\newblock \emph{IEEE Transactions on Big Data}, 7\penalty0 (3):\penalty0
  535--547, 2021.

\bibitem[Kang et~al.(2025)Kang, Ge, Hu, Zhang, Wang, and
  Zhan]{kang2025bigvectorbench}
Guoxin Kang, Zhongxin Ge, Jingpei Hu, Xueya Zhang, Lei Wang, and Jianfeng Zhan.
\newblock {BigVectorBench}: Heterogeneous data embedding and compound queries
  are essential in evaluating vector databases.
\newblock \emph{Proceedings of the VLDB Endowment}, 18\penalty0 (5):\penalty0
  1536--1550, 2025.

\bibitem[Karpukhin et~al.(2020)Karpukhin, O{\u{g}}uz, Min, Lewis, Wu, Edunov,
  Chen, and Yih]{karpukhin2020dense}
Vladimir Karpukhin, Barlas O{\u{g}}uz, Sewon Min, Patrick Lewis, Ledell Wu,
  Sergey Edunov, Danqi Chen, and Wen-tau Yih.
\newblock Dense passage retrieval for open-domain question answering.
\newblock In \emph{Proceedings of the 2020 Conference on Empirical Methods in
  Natural Language Processing (EMNLP)}, pages 6769--6781, 2020.

\bibitem[Khashabi et~al.(2021)Khashabi, Ng, Khot, Sabharwal, Hajishirzi, and
  Callison-Burch]{khashabi2021gooaq}
Daniel Khashabi, Amos Ng, Tushar Khot, Ashish Sabharwal, Hannaneh Hajishirzi,
  and Chris Callison-Burch.
\newblock {GooAQ}: Open question answering with diverse answer types.
\newblock In \emph{Findings of the Association for Computational Linguistics:
  {EMNLP} 2021}, pages 421--433, 2021.

\bibitem[Khattab and Zaharia(2020)]{khattab2020colbert}
Omar Khattab and Matei Zaharia.
\newblock {ColBERT}: Efficient and effective passage search via contextualized
  late interaction over {BERT}.
\newblock In \emph{Proceedings of the 43rd International ACM SIGIR Conference
  on Research and Development in Information Retrieval}, pages 39--48, 2020.

\bibitem[Kolouri et~al.(2019)Kolouri, Nadjahi, Simsekli, Badeau, and
  Rohde]{kolouri2019generalized}
Soheil Kolouri, Kimia Nadjahi, Umut Simsekli, Roland Badeau, and Gustavo~K.
  Rohde.
\newblock Generalized sliced {Wasserstein} distances.
\newblock \emph{Advances in Neural Information Processing Systems},
  32:\penalty0 261--272, 2019.

\bibitem[Kriegel et~al.(2017)Kriegel, Schubert, and Zimek]{kriegel2017black}
Hans-Peter Kriegel, Erich Schubert, and Arthur Zimek.
\newblock The (black) art of runtime evaluation: Are we comparing algorithms or
  implementations?
\newblock \emph{Knowledge and Information Systems}, 52:\penalty0 341--378,
  2017.

\bibitem[Kuffo and Boncz(2025)]{kuffo2025bang}
Leonardo Kuffo and Peter Boncz.
\newblock Bang for the buck: Vector search on cloud {CPUs}.
\newblock In \emph{Proceedings of the 21st International Workshop on Data
  Management on New Hardware}, DaMoN '25. Association for Computing Machinery,
  2025.

\bibitem[Kurtzer et~al.(2017)Kurtzer, Sochat, and
  Bauer]{kurtzer2017singularity}
Gregory~M. Kurtzer, Vanessa Sochat, and Michael~W. Bauer.
\newblock Singularity: Scientific containers for mobility of compute.
\newblock \emph{PLOS ONE}, 12\penalty0 (5):\penalty0 e0177459, 2017.

\bibitem[Lee et~al.(2024)Lee, Shakir, Koenig, and Lipp]{emb2024mxbai}
Sean Lee, Aamir Shakir, Darius Koenig, and Julius Lipp.
\newblock Open source strikes bread - new fluffy embedding model, 2024.
\newblock URL \url{https://www.mixedbread.com/blog/mxbai-embed-large-v1}.

\bibitem[Lewis et~al.(2020)Lewis, Perez, Piktus, Petroni, Karpukhin, Goyal,
  K{\"u}ttler, Lewis, Yih, Rockt{\"a}schel, Riedel, and
  Kiela]{lewis2020retrieval}
Patrick Lewis, Ethan Perez, Aleksandra Piktus, Fabio Petroni, Vladimir
  Karpukhin, Naman Goyal, Heinrich K{\"u}ttler, Mike Lewis, Wen-tau Yih, Tim
  Rockt{\"a}schel, Sebastian Riedel, and Douwe Kiela.
\newblock Retrieval-augmented generation for knowledge-intensive {NLP} tasks.
\newblock \emph{Advances in Neural Information Processing Systems},
  33:\penalty0 9459--9474, 2020.

\bibitem[Lewis et~al.(2021)Lewis, Wu, Liu, Minervini, K{\"u}ttler, Piktus,
  Stenetorp, and Riedel]{lewis2021paq}
Patrick Lewis, Yuxiang Wu, Linqing Liu, Pasquale Minervini, Heinrich
  K{\"u}ttler, Aleksandra Piktus, Pontus Stenetorp, and Sebastian Riedel.
\newblock {PAQ}: 65 million probably-asked questions and what you can do with
  them.
\newblock \emph{Transactions of the Association for Computational Linguistics},
  9:\penalty0 1098--1115, 2021.

\bibitem[Li et~al.(2020)Li, Zhang, Sun, Wang, Li, Zhang, and
  Lin]{li2019approximate}
Wen Li, Ying Zhang, Yifang Sun, Wei Wang, Mingjie Li, Wenjie Zhang, and Xuemin
  Lin.
\newblock Approximate nearest neighbor search on high dimensional
  data---experiments, analyses, and improvement.
\newblock \emph{IEEE Transactions on Knowledge and Data Engineering},
  32\penalty0 (8):\penalty0 1475--1488, 2020.

\bibitem[Lim et~al.(2026)Lim, Kim, Kim, and Do]{lim2026revisiting}
Mintaek Lim, Dogeun Kim, Minwoo Kim, and Jaeyoung Do.
\newblock Revisiting filtered {ANN} benchmarks: A hardness-controlled benchmark
  generator for realistic evaluation.
\newblock \emph{arXiv preprint arXiv:2606.14193}, 2026.

\bibitem[Liu et~al.(2025)Liu, Chen, Lu, Jiang, Han, Zhang, Chen, Zhang, Ding,
  Zhang, Chen, Yang, Yang, and Qiu]{liu2024retrievalattention}
Di~Liu, Meng Chen, Baotong Lu, Huiqiang Jiang, Zhenhua Han, Qianxi Zhang,
  Qi~Chen, Chengruidong Zhang, Bailu Ding, Kai Zhang, Chen Chen, Fan Yang,
  Yuqing Yang, and Lili Qiu.
\newblock {RetrievalAttention}: Accelerating long-context {LLM} inference via
  vector retrieval.
\newblock \emph{Advances in Neural Information Processing Systems},
  38:\penalty0 54358--54385, 2025.

\bibitem[Liu et~al.(2019)Liu, Ott, Goyal, Du, Joshi, Chen, Levy, Lewis,
  Zettlemoyer, and Stoyanov]{liu2019roberta}
Yinhan Liu, Myle Ott, Naman Goyal, Jingfei Du, Mandar Joshi, Danqi Chen, Omer
  Levy, Mike Lewis, Luke Zettlemoyer, and Veselin Stoyanov.
\newblock {RoBERTa}: A robustly optimized {BERT} pretraining approach.
\newblock \emph{arXiv preprint arXiv:1907.11692}, 2019.

\bibitem[Malkov and Yashunin(2020)]{malkov2018efficient}
Yu~A. Malkov and Dmitry~A. Yashunin.
\newblock Efficient and robust approximate nearest neighbor search using
  hierarchical navigable small world graphs.
\newblock \emph{IEEE Transactions on Pattern Analysis and Machine
  Intelligence}, 42\penalty0 (4):\penalty0 824--836, 2020.

\bibitem[Mazar{\'e} et~al.(2025)Mazar{\'e}, Szilvasy, Lomeli, Massa, Murray,
  J{\'e}gou, and Douze]{mazare2025inference}
Pierre-Emmanuel Mazar{\'e}, Gergely Szilvasy, Maria Lomeli, Francisco Massa,
  Naila Murray, Herv{\'e} J{\'e}gou, and Matthijs Douze.
\newblock Inference-time sparse attention with asymmetric indexing.
\newblock \emph{arXiv preprint arXiv:2502.08246}, 2025.

\bibitem[McInnes(2018)]{pynndescent}
Leland McInnes.
\newblock {PyNNDescent}: A {Python} library for approximate nearest neighbors,
  2018.
\newblock URL \url{https://github.com/lmcinnes/pynndescent/}.

\bibitem[{Microsoft}(2026)]{microsoft2026harrier}
{Microsoft}.
\newblock {Harrier-OSS-v1-270M}, 2026.
\newblock URL \url{https://huggingface.co/microsoft/harrier-oss-v1-270m}.

\bibitem[Muennighoff et~al.(2023)Muennighoff, Tazi, Magne, and
  Reimers]{muennighoff2023mteb}
Niklas Muennighoff, Nouamane Tazi, Lo{\"i}c Magne, and Nils Reimers.
\newblock {MTEB}: Massive text embedding benchmark.
\newblock In \emph{Proceedings of the 17th Conference of the European Chapter
  of the Association for Computational Linguistics}, pages 2014--2037, 2023.

\bibitem[Munyampirwa et~al.(2026)Munyampirwa, Lakshman, and
  Coleman]{munyampirwa2024down}
Blaise Munyampirwa, Vihan Lakshman, and Benjamin Coleman.
\newblock Down with the hierarchy: The `{H}' in {HNSW} stands for ``hubs''.
\newblock In \emph{Proceedings of the 48th European Conference on Information
  Retrieval (ECIR 2026), Part III}, pages 33--48, 2026.

\bibitem[Nussbaum et~al.(2024)Nussbaum, Duderstadt, and
  Mulyar]{nussbaum2024nomic}
Zach Nussbaum, Brandon Duderstadt, and Andriy Mulyar.
\newblock Nomic embed vision: Expanding the latent space.
\newblock \emph{arXiv preprint arXiv:2406.18587}, 2024.

\bibitem[Nussbaum et~al.(2025)Nussbaum, Morris, Mulyar, and
  Duderstadt]{nussbaum2025nomic}
Zach Nussbaum, John~Xavier Morris, Andriy Mulyar, and Brandon Duderstadt.
\newblock {Nomic Embed}: Training a reproducible long context text embedder.
\newblock \emph{Transactions on Machine Learning Research}, 2025.

\bibitem[NVIDIA(2024)]{cuvs}
NVIDIA.
\newblock {cuVS}: Vector search and clustering on the {GPU}, 2024.
\newblock URL \url{https://github.com/rapidsai/cuvs}.

\bibitem[Ootomo et~al.(2024)Ootomo, Naruse, Nolet, Wang, Feher, and
  Wang]{ootomo2024cagra}
Hiroyuki Ootomo, Akira Naruse, Corey Nolet, Ray Wang, Tamas Feher, and Yong
  Wang.
\newblock {CAGRA}: Highly parallel graph construction and approximate nearest
  neighbor search for {GPUs}.
\newblock In \emph{2024 IEEE 40th International Conference on Data Engineering
  (ICDE)}, pages 4236--4247. IEEE, 2024.

\bibitem[Pennington et~al.(2014)Pennington, Socher, and
  Manning]{pennington2014glove}
Jeffrey Pennington, Richard Socher, and Christopher~D. Manning.
\newblock {GloVe}: Global vectors for word representation.
\newblock In \emph{Proceedings of the 2014 Conference on Empirical Methods in
  Natural Language Processing (EMNLP)}, pages 1532--1543, 2014.

\bibitem[Pham and Liu(2022)]{pham2022falconn++}
Ninh Pham and Tao Liu.
\newblock {FALCONN++}: A locality-sensitive filtering approach for approximate
  nearest neighbor search.
\newblock \emph{Advances in Neural Information Processing Systems},
  35:\penalty0 31186--31198, 2022.

\bibitem[Radford et~al.(2021)Radford, Kim, Hallacy, Ramesh, Goh, Agarwal,
  Sastry, Askell, Mishkin, Clark, Krueger, and Sutskever]{radford2021learning}
Alec Radford, Jong~Wook Kim, Chris Hallacy, Aditya Ramesh, Gabriel Goh,
  Sandhini Agarwal, Girish Sastry, Amanda Askell, Pamela Mishkin, Jack Clark,
  Gretchen Krueger, and Ilya Sutskever.
\newblock Learning transferable visual models from natural language
  supervision.
\newblock In \emph{Proceedings of the 38th International Conference on Machine
  Learning}, volume 139 of \emph{Proceedings of Machine Learning Research},
  pages 8748--8763, 2021.

\bibitem[Reimers and Gurevych(2019)]{reimers2019sentencebert}
Nils Reimers and Iryna Gurevych.
\newblock {Sentence-BERT}: Sentence embeddings using {S}iamese {BERT}-networks.
\newblock In \emph{Proceedings of the 2019 Conference on Empirical Methods in
  Natural Language Processing and the 9th International Joint Conference on
  Natural Language Processing (EMNLP-IJCNLP)}, pages 3982--3992, 2019.

\bibitem[Sanh et~al.(2019)Sanh, Debut, Chaumond, and Wolf]{sanh2019distilbert}
Victor Sanh, Lysandre Debut, Julien Chaumond, and Thomas Wolf.
\newblock {DistilBERT}, a distilled version of {BERT}: Smaller, faster, cheaper
  and lighter.
\newblock \emph{arXiv preprint arXiv:1910.01108}, 2019.

\bibitem[Santhanam et~al.(2022{\natexlab{a}})Santhanam, Khattab, Potts, and
  Zaharia]{santhanam2022plaid}
Keshav Santhanam, Omar Khattab, Christopher Potts, and Matei Zaharia.
\newblock {PLAID}: An efficient engine for late interaction retrieval.
\newblock In \emph{Proceedings of the 31st ACM International Conference on
  Information \& Knowledge Management}, pages 1747--1756, 2022{\natexlab{a}}.

\bibitem[Santhanam et~al.(2022{\natexlab{b}})Santhanam, Khattab, Saad-Falcon,
  Potts, and Zaharia]{santhanam2022colbertv2}
Keshav Santhanam, Omar Khattab, Jon Saad-Falcon, Christopher Potts, and Matei
  Zaharia.
\newblock {ColBERTv2}: Effective and efficient retrieval via lightweight late
  interaction.
\newblock In \emph{Proceedings of the 2022 Conference of the North American
  Chapter of the Association for Computational Linguistics: Human Language
  Technologies}, pages 3715--3734, 2022{\natexlab{b}}.

\bibitem[Seo et~al.(2019)Seo, Lee, Kwiatkowski, Parikh, Farhadi, and
  Hajishirzi]{seo2019real}
Minjoon Seo, Jinhyuk Lee, Tom Kwiatkowski, Ankur Parikh, Ali Farhadi, and
  Hannaneh Hajishirzi.
\newblock Real-time open-domain question answering with dense-sparse phrase
  index.
\newblock In \emph{Proceedings of the 57th Annual Meeting of the Association
  for Computational Linguistics}, pages 4430--4441. Association for
  Computational Linguistics, 2019.

\bibitem[Simhadri et~al.(2022)Simhadri, Williams, Aum{\"u}ller, Douze, Babenko,
  Baranchuk, Chen, Hosseini, Krishnaswamny, Srinivasa, Subramanya, and
  Wang]{simhadri2022results}
Harsha~Vardhan Simhadri, George Williams, Martin Aum{\"u}ller, Matthijs Douze,
  Artem Babenko, Dmitry Baranchuk, Qi~Chen, Lucas Hosseini, Ravishankar
  Krishnaswamny, Gopal Srinivasa, Suhas~Jayaram Subramanya, and Jingdong Wang.
\newblock Results of the {NeurIPS’21} challenge on billion-scale approximate
  nearest neighbor search.
\newblock In \emph{Proceedings of the NeurIPS 2021 Competitions and
  Demonstrations Track}, volume 176 of \emph{Proceedings of Machine Learning
  Research}, pages 177--189, 2022.

\bibitem[Simhadri et~al.(2025)Simhadri, Aum{\"u}ller, Douze, Baranchuk, Ingber,
  Liberty, Williams, Landrum, Manohar, Karjikar, Dhulipala, Chen, Chen, Ma,
  Zhang, Cai, Shi, Zheng, Chen, Yin, and Huang]{simhadri2025results}
Harsha~Vardhan Simhadri, Martin Aum{\"u}ller, Matthijs Douze, Dmitry Baranchuk,
  Amir Ingber, Edo Liberty, George Williams, Ben Landrum, Magdalen Manohar,
  Mazin Karjikar, Laxman Dhulipala, Meng Chen, Yue Chen, Rui Ma, Kai Zhang,
  Yuzheng Cai, Jiayang Shi, Weiguo Zheng, Yizhuo Chen, Jie Yin, and Ben Huang.
\newblock Results of the {Big ANN}: {NeurIPS’23} competition.
\newblock \emph{Advances in Neural Information Processing Systems}, 38, 2025.

\bibitem[Spotify(2013)]{annoy}
Spotify.
\newblock Annoy: Approximate nearest neighbors oh yeah, 2013.
\newblock URL \url{https://github.com/spotify/annoy}.

\bibitem[Sproull(1991)]{sproull1991refinements}
Robert~F. Sproull.
\newblock Refinements to nearest-neighbor searching in $k$-dimensional trees.
\newblock \emph{Algorithmica}, 6:\penalty0 579--589, 1991.

\bibitem[Su et~al.(2023)Su, Shi, Kasai, Wang, Hu, Ostendorf, Yih, Smith,
  Zettlemoyer, and Yu]{su2023one}
Hongjin Su, Weijia Shi, Jungo Kasai, Yizhong Wang, Yushi Hu, Mari Ostendorf,
  Wen-tau Yih, Noah~A. Smith, Luke Zettlemoyer, and Tao Yu.
\newblock One embedder, any task: Instruction-finetuned text embeddings.
\newblock In \emph{Findings of the Association for Computational Linguistics:
  ACL 2023}, pages 1102--1121, 2023.

\bibitem[Subramanya et~al.(2019)Subramanya, Devvrit, Simhadri, Krishnawamy, and
  Kadekodi]{jayaram2019diskann}
Suhas~Jayaram Subramanya, Fnu Devvrit, Harsha~Vardhan Simhadri, Ravishankar
  Krishnawamy, and Rohan Kadekodi.
\newblock {DiskANN}: Fast accurate billion-point nearest neighbor search on a
  single node.
\newblock \emph{Advances in Neural Information Processing Systems},
  32:\penalty0 13771--13781, 2019.

\bibitem[Tepper et~al.(2024)Tepper, Bhati, Aguerrebere, Hildebrand, and
  Willke]{tepper2024leanvec}
Mariano Tepper, Ishwar~Singh Bhati, Cecilia Aguerrebere, Mark Hildebrand, and
  Theodore~L. Willke.
\newblock {LeanVec}: Searching vectors faster by making them fit.
\newblock \emph{Transactions on Machine Learning Research}, 2024.

\bibitem[Van~Horn et~al.(2021)Van~Horn, Cole, Beery, Wilber, Belongie, and
  Mac~Aodha]{vanhorn2021benchmarking}
Grant Van~Horn, Elijah Cole, Sara Beery, Kimberly Wilber, Serge Belongie, and
  Oisin Mac~Aodha.
\newblock Benchmarking representation learning for natural world image
  collections.
\newblock In \emph{Proceedings of the IEEE/CVF Conference on Computer Vision
  and Pattern Recognition}, pages 12884--12893, 2021.

\bibitem[Wang et~al.(2024)Wang, Ping, {McAfee}, Xu, Li, Shoeybi, and
  Catanzaro]{wang2024instructretro}
Boxin Wang, Wei Ping, Lawrence {McAfee}, Peng Xu, Bo~Li, Mohammad Shoeybi, and
  Bryan Catanzaro.
\newblock {InstructRetro}: Instruction tuning post retrieval-augmented
  pretraining.
\newblock In \emph{Proceedings of the 41st International Conference on Machine
  Learning}, volume 235 of \emph{Proceedings of Machine Learning Research},
  pages 51255--51272, 2024.

\bibitem[Wang et~al.(2021)Wang, Xu, Yue, and Wang]{wang2021comprehensive}
Mengzhao Wang, Xiaoliang Xu, Qiang Yue, and Yuxiang Wang.
\newblock A comprehensive survey and experimental comparison of graph-based
  approximate nearest neighbor search.
\newblock \emph{Proceedings of the VLDB Endowment}, 14\penalty0 (11):\penalty0
  1964--1978, 2021.

\bibitem[Wang et~al.(2026)Wang, Dong, Wu, Liu, Zhang, Zhao, Wan, Lei, Zhang,
  Zheng, Liu, Liao, and Jin]{wang2026candor}
Mingqi Wang, Junyao Dong, Zhuoyan Wu, Jun Liu, Ruicheng Zhang, Jianjun Zhao,
  Ruipeng Wan, Xinyan Lei, Shuhao Zhang, Bolong Zheng, Haikun Liu, Xiaofei
  Liao, and Hai Jin.
\newblock {CANDOR-Bench}: Benchmarking in-memory continuous {ANNS} under
  dynamic open-world streams [experiments \& analysis].
\newblock \emph{Proceedings of the ACM on Management of Data}, 4\penalty0
  (1):\penalty0 1--27, 2026.

\bibitem[Wang et~al.(2020)Wang, Wei, Dong, Bao, Yang, and Zhou]{wang2020minilm}
Wenhui Wang, Furu Wei, Li~Dong, Hangbo Bao, Nan Yang, and Ming Zhou.
\newblock {MiniLM}: Deep self-attention distillation for task-agnostic
  compression of pre-trained transformers.
\newblock \emph{Advances in Neural Information Processing Systems},
  33:\penalty0 5776--5788, 2020.

\bibitem[Wang(2025)]{PyGlass}
Zihao Wang.
\newblock Graph library for approximate similarity search, 2025.
\newblock URL \url{https://github.com/zilliztech/pyglass}.

\bibitem[Weyand et~al.(2020)Weyand, Araujo, Cao, and Sim]{weyand2020google}
Tobias Weyand, Andr{\'e} Araujo, Bingyi Cao, and Jack Sim.
\newblock Google landmarks dataset v2 - a large-scale benchmark for
  instance-level recognition and retrieval.
\newblock In \emph{Proceedings of the IEEE/CVF Conference on Computer Vision
  and Pattern Recognition}, pages 2575--2584, 2020.

\bibitem[Wilcoxon(1945)]{wilcoxon1945individual}
Frank Wilcoxon.
\newblock Individual comparisons by ranking methods.
\newblock \emph{Biometrics Bulletin}, 1\penalty0 (6):\penalty0 80--83, 1945.

\bibitem[Wolf et~al.(2019)Wolf, Debut, Sanh, Chaumond, Delangue, Moi, Cistac,
  Rault, Louf, Funtowicz, Davison, Shleifer, von Platen, Ma, Jernite, Plu, Xu,
  Le~Scao, Gugger, Drame, Lhoest, and Rush]{wolf2019huggingface}
Thomas Wolf, Lysandre Debut, Victor Sanh, Julien Chaumond, Clement Delangue,
  Anthony Moi, Pierric Cistac, Tim Rault, R{\'e}mi Louf, Morgan Funtowicz, Joe
  Davison, Sam Shleifer, Patrick von Platen, Clara Ma, Yacine Jernite, Julien
  Plu, Canwen Xu, Teven Le~Scao, Sylvain Gugger, Mariama Drame, Quentin Lhoest,
  and Alexander~M. Rush.
\newblock {HuggingFace's Transformers}: State-of-the-art natural language
  processing.
\newblock \emph{arXiv preprint arXiv:1910.03771}, 2019.

\bibitem[Xiong et~al.(2021)Xiong, Xiong, Li, Tang, Liu, Bennett, Ahmed, and
  Overwijk]{xiong2021approximate}
Lee Xiong, Chenyan Xiong, Ye~Li, Kwok-Fung Tang, Jialin Liu, Paul~N. Bennett,
  Junaid Ahmed, and Arnold Overwijk.
\newblock Approximate nearest neighbor negative contrastive learning for dense
  text retrieval.
\newblock In \emph{International Conference on Learning Representations}, 2021.

\bibitem[{Yahoo Japan}(2015)]{ngt}
{Yahoo Japan}.
\newblock {NGT}: Neighborhood graph and tree for indexing high-dimensional
  data, 2015.
\newblock URL \url{https://github.com/NGT-labs/NGT}.

\bibitem[Yang et~al.(2020)Yang, Yi, Zhiyuan~Cheng, Hong, Li, Xiaoming~Wang, Xu,
  and Chi]{yang2020mixed}
Ji~Yang, Xinyang Yi, Derek Zhiyuan~Cheng, Lichan Hong, Yang Li, Simon
  Xiaoming~Wang, Taibai Xu, and Ed~H. Chi.
\newblock Mixed negative sampling for learning two-tower neural networks in
  recommendations.
\newblock In \emph{Companion Proceedings of the Web Conference 2020}, pages
  441--447, 2020.

\bibitem[Yang et~al.(2018)Yang, Qi, Zhang, Bengio, Cohen, Salakhutdinov, and
  Manning]{yang2018hotpotqa}
Zhilin Yang, Peng Qi, Saizheng Zhang, Yoshua Bengio, William~W. Cohen, Ruslan
  Salakhutdinov, and Christopher~D. Manning.
\newblock {HotpotQA}: A dataset for diverse, explainable multi-hop question
  answering.
\newblock In \emph{Proceedings of the 2018 Conference on Empirical Methods in
  Natural Language Processing}, pages 2369--2380, 2018.

\bibitem[Zeng et~al.(2024)Zeng, Wu, Hu, Shi, Sun, and Zhang]{zeng2024candy}
Xianzhi Zeng, Zhuoyan Wu, Xinjing Hu, Xuanhua Shi, Shixuan Sun, and Shuhao
  Zhang.
\newblock {CANDY}: A benchmark for continuous approximate nearest neighbor
  search with dynamic data ingestion.
\newblock \emph{arXiv preprint arXiv:2406.19651}, 2024.

\bibitem[Zhang et~al.(2026)Zhang, Zhang, Fan, Li, and Du]{zhang2026vecbench}
Xiang Zhang, Chao Zhang, Ju~Fan, Guoliang Li, and Xiaoyong Du.
\newblock {VecBench}: A controllable benchmark for filtered vector search:
  [experiments \& analysis].
\newblock \emph{Proceedings of the ACM on Management of Data}, 4\penalty0
  (3):\penalty0 1--27, 2026.

\bibitem[Zhang et~al.(2025)Zhang, Li, Long, Zhang, Lin, Yang, Xie, Yang, Liu,
  Lin, Huang, and Zhou]{qwen3embedding}
Yanzhao Zhang, Mingxin Li, Dingkun Long, Xin Zhang, Huan Lin, Baosong Yang,
  Pengjun Xie, An~Yang, Dayiheng Liu, Junyang Lin, Fei Huang, and Jingren Zhou.
\newblock {Qwen3 Embedding}: Advancing text embedding and reranking through
  foundation models.
\newblock \emph{arXiv preprint arXiv:2506.05176}, 2025.

\end{thebibliography}

\clearpage
\appendix

\section{Benchmark and dataset description}
\label{app:datasets}

\subsection{Access}

The \textsc{VIBE} benchmark and its associated datasets are publicly available and hosted on GitHub and Hugging Face, ensuring continued availability. The datasets can also be easily recreated from scratch using the provided framework. The companion website, the benchmark code, and the precomputed datasets can be accessed via the following links:

\begin{itemize}
    \setlength\parskip{0pt}
    \setlength\parsep{0pt}
    \item[] \raisebox{-0.16em}{\twemoji[height=1em]{globe_with_meridians}}\hspace{1mm} \url{https://vector-index-bench.github.io/}
    \item[] \raisebox{-0.18em}{\simpleicon{github}}\hspace{1mm} \url{https://github.com/vector-index-bench/vibe/}
    \item[] \raisebox{-0.20em}{\includegraphics[height=1em]{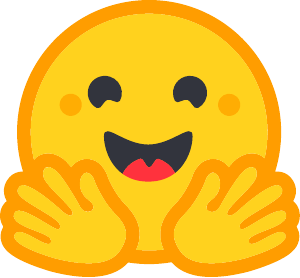}}\hspace{1mm} \url{https://huggingface.co/datasets/vector-index-bench/vibe}
\end{itemize}

\subsection{Maintenance}

\textsc{VIBE} is an ongoing effort, and we allow community contributions on GitHub for adding new methods to the benchmark. To contribute a new method, a contributor opens a pull request. After review and merging, the maintainers update the companion website with the new results. The results will be updated regularly using computing infrastructure provided by CSC – IT Center for Science, Finland. Instructions for adding new methods and datasets to the benchmark are provided in the repository README:
\begin{itemize}
    \setlength\itemsep{0pt}
    \setlength\parskip{0pt}
    \setlength\parsep{0pt}
    \item[] \raisebox{-0.18em}{\simpleicon{github}}\hspace{1mm} \url{https://github.com/vector-index-bench/vibe/blob/main/README.md}
\end{itemize}

\subsection{Reproducibility}

The benchmark code can be found in our GitHub repository. The README in the repository describes the requirements and instructions for running the benchmark. The benchmark framework automatically downloads the precomputed datasets, but the datasets can also be recreated from scratch using the provided instructions. After running the benchmark, all figures of this paper can be reproduced using the command \verb|./plot.sh --plot-type paper|

\subsection{License}

The \textsc{VIBE} benchmarking framework is licensed under the MIT license. Most of the datasets in \textsc{VIBE} are created by the present authors using the embeddings and data sources detailed in Section~\ref{sec:models-data-sources}, and our embedding datasets are distributed under CC BY 4.0. However, as part of \textsc{VIBE} we also redistribute the following embedding datasets: 
\begin{itemize}
    \setlength\itemsep{0pt}
    \setlength\parskip{0pt}
    \setlength\parsep{0pt}
    \item GloVe\,\footnote{\url{https://nlp.stanford.edu/projects/glove/}}~\citep{pennington2014glove} (under PDDL 1.0)
    \item Yandex T2I\,\footnote{\url{https://big-ann-benchmarks.com/neurips23.html}}~\citep{simhadri2025results} (a subset of 1 million vectors, under CC BY 4.0)
    \item LAION\,\footnote{\url{https://laion.ai/blog/laion-400-open-dataset/}} (a subset of 1 million vectors, under CC BY 4.0)
\end{itemize}

\subsection{Rights and responsibilities}

The authors bear all responsibility in case of violation of rights associated with the datasets.

\subsection{Intended uses}

The \textsc{VIBE} benchmark and its associated datasets are intended to be used only for benchmarking vector indexes. The datasets contain only raw embedding vectors and not the original documents or images that were used to compute the embeddings. The datasets are designed primarily for use with the \textsc{VIBE} framework, although they are also straightforward to use without the framework.

\subsection{Framework structure}

\noindent Each algorithm in the VIBE benchmark is provided as a self-contained module within the \texttt{vibe/algorithms/} directory, where each algorithm directory contains three files: 
\begin{itemize}
    \item \texttt{image.def}: Apptainer container definition that specifies the algorithm's dependencies and build instructions, inheriting from a common base image
    \item \texttt{config.yml}: defines algorithm variants with their hyperparameters, organized by data type and distance metric
    \item \texttt{module.py}: implements one or more Python classes with the required methods \verb|fit()| for index building and \verb|query()| for nearest neighbor search
\end{itemize}

\noindent New algorithms can be added by creating a new directory following this structure: implementing the interface with methods for fitting data and querying neighbors, defining algorithm variants with their hyperparameters in YAML format, and providing a container definition for reproducible installation and execution.

\subsection{Dataset structure}

Each dataset is distributed as a single HDF5 file.

\paragraph{File attributes} The HDF5 files contain the following attributes:
\begin{description}[itemsep=2.2pt]
  \item[\texttt{dimension}] The dimensionality of the data
  \item[\texttt{distance}] The distance metric to use; one of \texttt{euclidean}, \texttt{cosine}, \texttt{ip}, \texttt{hamming}, or \texttt{normalized}
  \item[\texttt{point\_type}] The precision of the vectors; one of \texttt{float}, \texttt{int8}, \texttt{uint8}, or \texttt{binary}
\end{description}

\paragraph{HDF5 datasets} The HDF5 files contain the following HDF5 datasets:
\begin{description}[itemsep=2.2pt]
  \item[\texttt{train}] Array of size \((m_{\text{corpus}}, \mathrm{dim})\): the embeddings used to build the vector index
  \item[\texttt{test}] Array of size \((m_{\text{test}}, \mathrm{dim})\): the test query embeddings
  \item[\texttt{neighbors}] Array of size \((m_{\text{test}}, 100)\): the IDs of the true 100 nearest neighbors of each query
  \item[\texttt{distances}] Array of size \((m_{\text{test}}, 100)\): the distances to the \texttt{neighbors}
  \item[\texttt{avg\_distances}] Array of size \(m_{\text{test}}\): the average distance from each test query to the corpus
\end{description}

\noindent Additionally, the HDF5 files of OOD datasets contain the following HDF5 datasets:
\begin{description}[itemsep=2.2pt]
  \item[\texttt{learn}] Array of size \((m_{\text{learn}}, \mathrm{dim})\): a larger sample from the query distribution
  \item[\texttt{learn\_neighbors}] Array of size \((m_{\text{learn}}, 100)\): the IDs of the true 100 nearest neighbors for each point in \texttt{learn}
\end{description}

\clearpage
\subsection{Models and data sources}
\label{sec:models-data-sources}

The embedding models used to produce the new datasets (see Tables~\ref{tab:id-datasets} and~\ref{tab:ood-datasets}) in \textsc{VIBE} are shown in Table~\ref{tab:embeddings}. For discussion, we refer to Sections~\ref{sec:data_sets} and~\ref{sec:ood_datasets}. The used data sources are shown in Table~\ref{tab:datasources}. Most of the data sources are under custom non-commercial licenses; for details, we refer to the provided references for each data source.

\paragraph{Approximate attention} To construct our approximate attention computation datasets, we use Hugging Face Transformers~\citep{wolf2019huggingface} with two long-context LLMs:
\begin{itemize}
    \item \texttt{llama-128-ip}: Extracted from Layer 13, Head 16 of the gradientai/Llama-3-8B-Instruct-262k\footnote{\url{https://huggingface.co/gradientai/Llama-3-8B-Instruct-262k}} model using a copy of \emph{The Picture of Dorian Gray}\footnote{\url{https://www.gutenberg.org/ebooks/174}} as the prompt
    \item \texttt{yi-128-ip}: Extracted from Layer 32, Head 14 of the 01-ai/Yi-6B-200K\footnote{\url{https://huggingface.co/01-ai/Yi-6B-200K}} model using a copy of the book \emph{Pride and Prejudice}\footnote{\url{https://www.gutenberg.org/ebooks/1342}} as the prompt
\end{itemize}

Both models use grouped-query attention and have 32 query heads, but Yi has 4 key/value heads and Llama has 8. Thus, each Yi key head is shared by 8 consecutive query heads, whereas each Llama key head is shared by 4. Within each layer, at each token position, we therefore split the query projection into 32 query-head vectors and the key projection into 4 vectors for Yi or 8 vectors for Llama. All of these head vectors are 128-dimensional. This produces a separate query matrix $Q\in\mathbb{R}^{L\times128}$ and key matrix $K\in\mathbb{R}^{L\times128}$ for every layer--query-head pair, where $L$ is the tokenized prompt length.

For a selected layer and head, the rows of $K$ form the indexed corpus and the rows of $Q$ form the queries. Following~\cite{mazare2025inference}, we additionally discard the keys and queries corresponding to the first token and the last 2047 tokens, as typically they have higher inner products with the queries and should be computed exactly.

To select the final layer--head pair, we estimate the sliced 2-Wasserstein distance~\citep{kolouri2019generalized} between $K$ and $Q$ for all layer and head combinations, and choose the pair with median distance as the final dataset. The benchmark results were similar across different prompts and candidate attention datasets with varying Wasserstein distances, so the selected pairs are representative of the approximate attention workload. The \texttt{create\_dataset.sh} script in the code repository can be used to download other layer--head datasets.

\paragraph{Multi-vector reduction} For the multi-vector reduction datasets \texttt{cqadupstack-muvera} and \texttt{cqadupstack-lemur}, we first use the ColBERTv2~\citep{santhanam2022colbertv2} model\footnote{\url{https://huggingface.co/colbert-ir/colbertv2.0}} to compute the token embeddings from CQADupStack~\citep{hoogeveen2015cqadupstack}. For each corpus document, we use a maximum of 220 token embeddings, and for each query document, we use exactly 32 token embeddings. To construct the single-vector reductions, for MUVERA\footnote{\url{https://github.com/google/graph-mining/tree/main/sketching/point_cloud}}~\citep{jayaram2024muvera}, we set $R_{\mathrm{reps}} = 40$, $k_{\mathrm{sim}} = 6$, and $d_{\mathrm{proj}} = 128$, and then apply a final projection to generate $5120$-dimensional fixed-dimensional encodings (FDEs). For LEMUR\footnote{\url{https://github.com/ejaasaari/lemur}}~\citep{jaasaari2026lemur}, we set the hidden-layer size to $d' = 2048$.

\begingroup
\footnotesize
\begin{table*}[!htbp]
\centering
\caption{Embedding models used to produce the \textsc{VIBE} datasets. The license for each embedding applies to the model; none of the models claim rights over their outputs.}
\label{tab:embeddings}
\setlength{\tabcolsep}{0pt}
\begin{tabular*}{\textwidth}{@{\extracolsep{\fill}}lllll@{}}
    \toprule
    Name & Model & Citation & License \\
    \midrule
    ALIGN & \href{https://huggingface.co/kakaobrain/align-base}{align-base} & \cite{jia2021scaling} & Custom \\
    CLIP & \href{https://huggingface.co/openai/clip-vit-base-patch32}{clip-vit-base-patch32} & \cite{radford2021learning} & MIT \\
    DINO & \href{https://huggingface.co/timm/vit_base_patch16_224.dino}{vit\_base\_patch16\_224.dino} & \cite{caron2021emerging} & Apache 2.0 \\
    DistilRoBERTa & \href{https://huggingface.co/sentence-transformers/all-distilroberta-v1}{all-distilroberta-v1}{} & \cite{sanh2019distilbert} & Apache 2.0 \\
    Harrier & \href{https://huggingface.co/microsoft/harrier-oss-v1-270m}{harrier-oss-v1-270m} & \cite{microsoft2026harrier} & MIT \\
    Jina & \href{https://huggingface.co/jinaai/jina-embeddings-v5-text-nano}{jina-embeddings-v5-text-nano} & \cite{akram2026jina} & CC-BY-NC-4.0 \\
    MiniLM & \href{https://huggingface.co/sentence-transformers/all-MiniLM-L6-v2}{all-MiniLM-L6-v2} & \cite{wang2020minilm} & Apache 2.0 \\
    MXBAI & \href{https://huggingface.co/mixedbread-ai/mxbai-embed-large-v1}{mxbai-embed-large-v1} & \cite{emb2024mxbai} & Apache 2.0 \\
    Nomic Text & \href{https://huggingface.co/nomic-ai/nomic-embed-text-v1.5}{nomic-embed-text-v1.5} & \cite{nussbaum2025nomic} & Apache 2.0 \\
    Nomic Vision & \href{https://huggingface.co/nomic-ai/nomic-embed-vision-v1.5}{nomic-embed-vision-v1.5} & \cite{nussbaum2024nomic} & Apache 2.0 \\
    Qwen & \href{https://huggingface.co/Qwen/Qwen3-Embedding-0.6B}{Qwen3-Embedding-0.6B} & \cite{qwen3embedding} & Apache 2.0 \\
    ResNet & \href{https://docs.pytorch.org/vision/main/models/generated/torchvision.models.resnet50.html}{resnet50} & \cite{he2016deep} & BSD 3 \\
    \bottomrule
\end{tabular*}

\let\originalTPTnoteSettings\TPTnoteSettings
\renewcommand{\TPTnoteSettings}{\originalTPTnoteSettings\footnotesize}

\begin{threeparttable}
\begin{center}
\caption{Data sources used to produce the \textsc{VIBE} embedding datasets.}
\label{tab:datasources}
    \setlength{\tabcolsep}{0pt}
    \begin{tabular*}{\textwidth}{@{\extracolsep{\fill}}llrrl@{}}
        \toprule
        Name & Type & $n$ & Reference & Citation \\
        \midrule
        AGNews & Text & 770 382 & \tnote{1} & N/A \\
        arXiv abstracts & Text & 1 345 643 & \tnote{2} & N/A \\
        CQADupStack & Text & 457 149 & \tnote{3} & \cite{hoogeveen2015cqadupstack} \\
        DPR & Text & 20 970 760 & \tnote{4} & \cite{karpukhin2020dense} \\
        GooAQ & Text & 1 476 024 & \tnote{5} & \cite{khashabi2021gooaq} \\
        HotpotQA & Text & 5 233 329 & \tnote{6} & \cite{yang2018hotpotqa} \\
        ImageNet & Image & 1 281 167 & \tnote{7} & \cite{deng2009imagenet} \\
        ImageNet-Captions & Text & 316 648 & \tnote{8} & \cite{fang2022data} \\
        iNaturalist Mini & Image & 500 000 & \tnote{9} & \cite{vanhorn2021benchmarking} \\
        Landmark & Image & 761 757 & \tnote{10} & \cite{weyand2020google} \\
        MS MARCO & Text & 8 841 823 & \tnote{11} & \cite{bajaj2016ms} \\
        Yahoo Answers & Text & 678 305 & \tnote{12} & N/A \\
        \bottomrule
    \end{tabular*}
\begin{tablenotes}
    \item [1] \url{http://groups.di.unipi.it/~gulli/AG_corpus_of_news_articles.html}
    \item [2] \url{https://www.kaggle.com/datasets/Cornell-University/arxiv}
    \item [3] \url{http://nlp.cis.unimelb.edu.au/resources/cqadupstack/}
    \item [4] \url{https://github.com/facebookresearch/DPR}
    \item [5] \url{https://github.com/allenai/gooaq}
    \item [6] \url{https://hotpotqa.github.io/}
    \item [7] \url{https://www.image-net.org/}
    \item [8] \url{https://github.com/mlfoundations/imagenet-captions}
    \item [9] \url{https://github.com/visipedia/inat_comp/tree/master/2021}
    \item [10] \url{https://github.com/cvdfoundation/google-landmark}
    \item [11] \url{https://microsoft.github.io/msmarco/}
    \item [12] \url{https://www.kaggle.com/datasets/soumikrakshit/yahoo-answers-dataset}
\end{tablenotes}
\end{center}
\end{threeparttable}
\end{table*}
\clearpage
\endgroup

\clearpage
\section{Interactive website}
\label{app:website}

\begin{figure}[!htpb]
    \centering
    \includegraphics[width=0.88\linewidth]{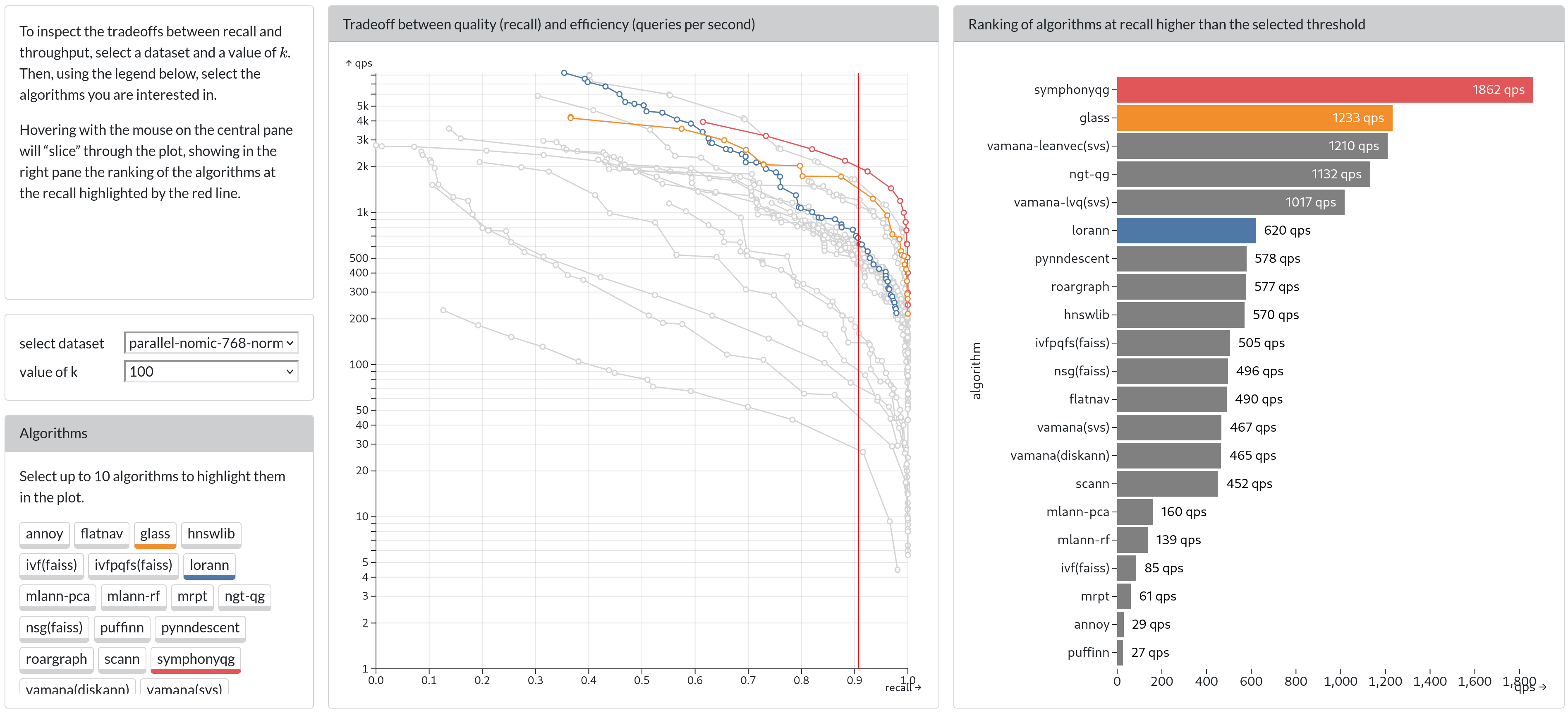}
    \caption{Screenshot of the webpage that allows users to explore the recall/throughput trade-off interactively.
    }
    \label{fig:screenshot-tradeoff}
    \includegraphics[width=0.88\linewidth]{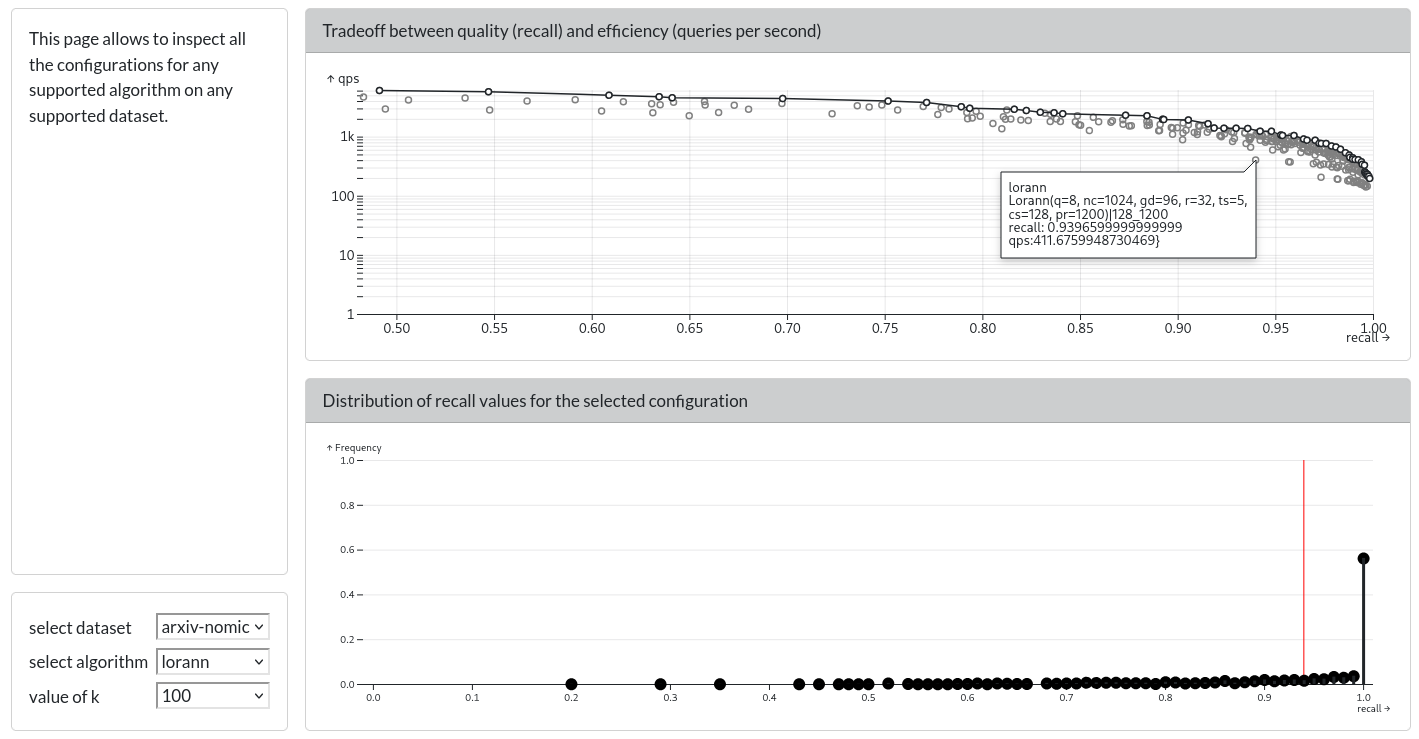}
    \caption{Screenshot of the webpage that allows users to explore the behavior of all the tested configurations of a given algorithm interactively.
    }
    \label{fig:screenshot-algorithm-focus}
\end{figure}

Our interactive website is available at \url{https://vector-index-bench.github.io/}. The website can also be run locally by following the instructions at \url{https://github.com/vector-index-bench/vector-index-bench.github.io}.
It provides pages that allow users to (1) compare all algorithms at a given recall level; (2) explore the distribution of queries and data of different datasets; (3) investigate the trade-offs of different implementations (Figure~\ref{fig:screenshot-tradeoff}); (4) assess the performance of all configurations for each algorithm (Figure~\ref{fig:screenshot-algorithm-focus}); (5) evaluate the robustness of algorithms on easy and difficult queries.

\clearpage
\section{Distribution shift in OOD datasets}
\label{app:distribution_shift}

We quantify the distribution shift in the OOD datasets using three metrics: Multiscale maximum mean discrepancy (MMD) captures general differences between distributions across multiple kernel length scales. The Fréchet distance measures differences in the means and covariance structures of the embedding distributions~\citep{dowson1982frechet}, while sliced 2-Wasserstein measures discrepancies between their one-dimensional projections and therefore captures broad geometric changes in their mass distributions~\citep{kolouri2019generalized}. 

In Table~\ref{table:distribution-distances}, we show these three metrics for each OOD dataset using data-versus-data ($D \leftrightarrow D$) (within-distribution reference) and data-versus-query ($D \leftrightarrow Q$) metrics. The query distribution is represented by the larger query sample provided with each OOD dataset rather than by the 1000 held-out test queries. All evaluated datasets are clearly out of distribution, as their data-versus-query distances consistently exceed the corresponding data-versus-data reference distances across all three metrics. The \texttt{llama} and \texttt{yi} datasets exhibit the most severe overall shifts, while \texttt{hotpotqa-harrier} and \texttt{yandex} show the weakest shifts.

\begin{table*}[ht!]
\centering
\caption{Distribution distances for the OOD datasets. For each metric, $D \leftrightarrow D$ measures the within-distribution distance between two corpus samples, while $D \leftrightarrow Q$ measures the distance between the corpus and query distributions. Larger $D \leftrightarrow Q$ values relative to $D \leftrightarrow D$ indicate a stronger distribution shift.}
\label{table:distribution-distances}
\begin{NiceTabular}{@{}l rr rr rr@{}}
\toprule
dataset & \multicolumn{2}{c}{multiscale MMD} & \multicolumn{2}{c}{Fréchet distance} & \multicolumn{2}{c}{sliced $2$-Wasserstein} \\
 & $D \leftrightarrow D$ & $D \leftrightarrow Q$ &  $D \leftrightarrow D$ & $D \leftrightarrow Q$ & $D \leftrightarrow D$ & $D \leftrightarrow Q$ \\
\midrule
hotpotqa-harrier & 0.0022 & 0.2084 & 0.0380 & 0.5322 & 0.0004 & 0.0176 \\
laion-clip & 0.0009 & 0.3298 & 0.0309 & 0.8818 & 0.0004 & 0.0328 \\
imagenet-align & 0.0013 & 0.4269 & 0.0371 & 1.0477 & 0.0004 & 0.0374 \\
yandex & 0.0010 & 0.0850 & 0.0277 & 0.5073 & 0.0008 & 0.0150 \\
cqadupstack-muvera & 0.0011 & 0.2335 & 6.1337 & 47.3362 & 0.0075 & 0.5478 \\
cqadupstack-lemur & 0.0014 & 0.5180 & 0.7450 & 9.6105 & 0.0024 & 0.1802 \\
yi & 0.0018 & 0.6612 & 0.2156 & 26.4129 & 0.0113 & 2.2717 \\
llama & 0.0014 & 0.8017 & 0.1931 & 32.7253 & 0.0107 & 2.7928 \\
\bottomrule
\end{NiceTabular}
\end{table*}

\clearpage
\section{Algorithms}
\label{app:impdetails}

In Table~\ref{tab:impdetails}, we give a detailed list of versions of each algorithm used in the experiments. The name of each method links to the corresponding library used in the benchmark. \ulc{Clustering-based}{color_green} methods are highlighted in green, \ulc{tree-based}{color_purple} in purple, \ulc{hashing-based}{color_orange} in orange, and \ulc{graph-based}{color_blue} in blue.

\begin{table}[!htbp]
\centering
\caption{Details of the algorithm implementations included in the benchmark.}
\label{tab:impdetails}
\begin{NiceTabular}{lll}
\toprule
Method & Reference & Version\\\midrule
\rowcolor{light_purple} \href{https://github.com/spotify/annoy}{ANNOY} & \cite{annoy} & 1.17.3\\
\rowcolor{light_orange} \href{https://github.com/NinhPham/FalconnPP}{FALCONN++} & \cite{pham2022falconn++}& git+5fd3f17\\
\rowcolor{light_blue} \href{https://github.com/BlaiseMuhirwa/flatnav}{FlatNav} & \cite{munyampirwa2024down}& 0.1.2\\
\rowcolor{light_blue} \href{https://github.com/zilliztech/pyglass}{Glass} & \cite{PyGlass} & git+d2296ec\\
\rowcolor{light_blue} \href{https://github.com/nmslib/hnswlib}{HNSW} & \cite{malkov2018efficient}& 0.8.0\\
\rowcolor{light_blue} \href{https://github.com/VectorDB-NTU/RaBitQ-Library}{HNSW-RaBitQ} & \cite{gao2024rabitq,gao2025practical,gao2025rabitq} & git+5ea4df0\\
\rowcolor{light_green} \href{https://github.com/facebookresearch/faiss}{IVF} & \cite{jegou2011product,douze2024faiss}& 1.14.3\\
\rowcolor{light_green} \href{https://github.com/facebookresearch/faiss}{IVF-PQ} & \cite{jegou2011product,douze2024faiss}& 1.14.3\\
\rowcolor{light_green} \href{https://github.com/VectorDB-NTU/RaBitQ-Library}{IVF-RaBitQ} & \cite{gao2024rabitq,gao2025rabitq,gao2025practical} & git+5ea4df0\\
\rowcolor{light_blue} \href{https://github.com/intel/ScalableVectorSearch}{LVQ} & \cite{aguerrebere2023similarity}& 0.4.0\\
\rowcolor{light_blue} \href{https://github.com/intel/ScalableVectorSearch}{LeanVec} & \cite{tepper2024leanvec}& 0.4.0\\
\rowcolor{light_green} \href{https://github.com/ejaasaari/lorann}{LoRANN} & \cite{jaasaari2024lorann}& 0.4.5\\
\rowcolor{light_purple} \href{https://github.com/ejaasaari/mlann}{MLANN} & \cite{hyvonen2022multilabel}& git+ba141b4\\
\rowcolor{light_purple} \href{https://github.com/vioshyvo/mrpt}{MRPT} & \cite{hyvonen2016fast,jaasaari2019efficient}& 2.0.4\\
\rowcolor{light_blue} \href{https://github.com/NGT-labs/NGT/}{NGT-ONNG} & \cite{iwasaki2018optimization,ngt} & 2.7.4\\
\rowcolor{light_blue} \href{https://github.com/NGT-labs/NGT/}{NGT-QG} & \cite{ngt} & 2.7.4\\
\rowcolor{light_blue} \href{https://github.com/facebookresearch/faiss}{NSG} & \cite{fu2017nsg,douze2024faiss}& 1.14.3\\
\rowcolor{light_orange} \href{https://github.com/puffinn/puffinn}{PUFFINN} & \cite{aumuller2019puffinn}& git+fd86b0d\\
\rowcolor{light_blue} \href{https://github.com/lmcinnes/pynndescent}{PyNNDescent} & \cite{wei2021pynndescent,pynndescent}& 0.6.0\\
\rowcolor{light_blue} \href{https://github.com/matchyc/RoarGraph}{RoarGraph} & \cite{chen2024roargraph}& git+f2b49b6\\
\rowcolor{light_green} \href{https://github.com/google-research/google-research/tree/master/scann}{ScaNN} & \cite{guo2020accelerating}& 1.4.2\\
\rowcolor{light_blue} \href{https://github.com/gouyt13/SymphonyQG}{SymphonyQG} & \cite{gou2025symphonyqg}& git+32a0019\\
\rowcolor{light_blue} \href{https://github.com/microsoft/DiskANN}{Vamana} & \cite{jayaram2019diskann}& 0.7.0\\
\bottomrule
\end{NiceTabular}
 \vspace*{6pt}
\end{table}
 
\noindent The list of algorithms used in the GPU experiments is given in Table~\ref{tab:gpu-algorithms}.

\begin{table}[!htbp]
\centering
\caption{Details of the GPU algorithm implementations included in the benchmark.}
\label{tab:gpu-algorithms}
\begin{NiceTabular}{lll}
\toprule
Method & Reference & Version\\\midrule
\rowcolor{light_blue} \href{https://github.com/rapidsai/cuvs}{CAGRA (cuVS)} & \cite{ootomo2024cagra,cuvs} & 26.04.00\\
\rowcolor{light_blue} \href{https://github.com/cgtuebingen/ggnn}{GGNN} & \cite{groh2022ggnn} & 0.9\\
\rowcolor{light_green} \href{https://github.com/rapidsai/cuvs}{IVF-PQ (cuVS)} & \cite{cuvs} & 26.04.00\\
\rowcolor{light_green} \href{https://github.com/rapidsai/cuvs}{IVF (cuVS)} & \cite{cuvs} & 26.04.00\\
\rowcolor{light_green} \href{https://github.com/facebookresearch/faiss}{IVF (Faiss)} & \cite{johnson2019billion,douze2024faiss} & 1.14.3\\
\bottomrule
\end{NiceTabular}
\vspace*{6pt}
\end{table}
 
\clearpage
\section{Index construction time}
\label{app:construction_time}

We measure the index construction time on a single CPU core. Table~\ref{table:construction} shows index construction times for all algorithms on 10 datasets at $95\%$ average recall for $k = 100$ at optimal hyperparameters (a dash means that the algorithm did not reach $95\%$ average recall).

\begin{table*}[!htbp]
\centering
\caption{Index construction times (seconds) with throughput-optimized hyperparameters at $95\%$ average recall. The final column reports the mean $\pm$ standard deviation of construction time relative to the fastest construction time on each dataset. Algorithms are ordered by mean relative build time. \protect\ulc{Clustering}{color_green} methods are highlighted in green, \protect\ulc{tree}{color_purple} in purple, \protect\ulc{hashing}{color_orange} in orange, and \protect\ulc{graph}{color_blue} in blue.}
\label{table:construction}
\newcommand{\relbuildtime}[2]{$#1$\,{\fontsize{7}{8}\selectfont\scalebox{0.72}{$\pm$}$#2$}}
\setlength{\tabcolsep}{2.2pt}
\begin{NiceTabular}{>{\small}l r r r r r r r r r r !{\hspace{2pt}} | >{\footnotesize}r}
\toprule
\multicolumn{1}{l}{algorithm} &
\multicolumn{1}{c}{\rotatebox{90}{\small agnews-mxbai}} &
\multicolumn{1}{c}{\rotatebox{90}{\small arxiv-nomic}} &
\multicolumn{1}{c}{\rotatebox{90}{\small glove}} &
\multicolumn{1}{c}{\rotatebox{90}{\small gooaq-distilroberta}} &
\multicolumn{1}{c}{\rotatebox{90}{\small imagenet-clip}} &
\multicolumn{1}{c}{\rotatebox{90}{\small inaturalist-resnet}} &
\multicolumn{1}{c}{\rotatebox{90}{\small landmark-dino}} &
\multicolumn{1}{c}{\rotatebox{90}{\small landmark-nomic}} &
\multicolumn{1}{c}{\rotatebox{90}{\small msmarco-qwen}} &
\multicolumn{1}{c@{\hspace{3.8pt}}|}{\rotatebox{90}{\small yahoo-minilm}} &
\multicolumn{1}{c}{\rotatebox{90}{\footnotesize \shortstack{relative build time\\(mean $\pm$ sd)}}} \\
\midrule
\rowcolor{light_green} ivfpqfs(faiss) & 92 & 81 & 71 & 82 & 35 & 145 & 69 & 35 & 561 & 38 & \relbuildtime{1.4}{0.6} \\
\rowcolor{light_green} rabitq-ivf & 78 & 77 & 235 & 79 & 56 & 47 & 70 & 58 & 395 & 23 & \relbuildtime{1.4}{0.8} \\
\rowcolor{light_green} ivf(faiss) & 214 & 571 & 63 & 563 & 432 & 108 & 50 & 443 & 394 & 246 & \relbuildtime{5.8}{4.5} \\
\rowcolor{light_green} scann & 689 & 761 & 412 & 764 & 479 & 1004 & 468 & 248 & 5497 & 241 & \relbuildtime{11.1}{4.1} \\
\rowcolor{light_blue} glass & 987 & 929 & - & 1045 & 658 & 535 & 797 & 375 & 11674 & 562 & \relbuildtime{16.5}{6.1} \\
\rowcolor{light_blue} pynndescent & 327 & 427 & 2524 & 965 & 288 & 2136 & 551 & 230 & 11404 & 384 & \relbuildtime{17.9}{14.2} \\
\rowcolor{light_blue} vamana-leanvec(svs) & 1052 & 1499 & - & 1698 & 912 & 501 & 596 & 427 & 10369 & 774 & \relbuildtime{19.4}{7.5} \\
\rowcolor{light_purple} mrpt & 1263 & 1872 & 870 & 1966 & 1424 & 1037 & 1023 & 988 & - & 623 & \relbuildtime{24.1}{7.3} \\
\rowcolor{light_orange} puffinn & - & 2106 & 676 & 2363 & 1147 & - & 1258 & 1231 & - & 689 & \relbuildtime{27.2}{7.3} \\
\rowcolor{light_blue} vamana-lvq(svs) & 1823 & 2219 & - & 2323 & 1070 & - & 1133 & 1017 & 14797 & 832 & \relbuildtime{29.6}{4.8} \\
\rowcolor{light_purple} annoy & 446 & 2797 & 2125 & 3072 & 487 & 2197 & 2341 & 354 & 22873 & 948 & \relbuildtime{33.1}{16.6} \\
\rowcolor{light_green} lorann & 1149 & 1656 & - & 3526 & 2706 & 1086 & 1045 & 1773 & 19055 & 1539 & \relbuildtime{40.7}{20.7} \\
\rowcolor{light_blue} flatnav & 1585 & 3021 & 5509 & 3231 & 1362 & 1890 & 2075 & 909 & 16878 & 1542 & \relbuildtime{44.3}{18.3} \\
\rowcolor{light_blue} nsg(faiss) & 2256 & 2863 & - & 2796 & 2485 & 1626 & 2069 & 1801 & 20204 & 1465 & \relbuildtime{46.0}{13.3} \\
\rowcolor{light_blue} symphonyqg & 2207 & 3152 & - & 4085 & 2013 & - & 1760 & 1589 & 23779 & 1437 & \relbuildtime{47.6}{11.4} \\
\rowcolor{light_orange} falconnpp & - & 2843 & 1906 & 1573 & 3152 & 1762 & 927 & 2994 & 28190 & 1343 & \relbuildtime{49.6}{25.7} \\
\rowcolor{light_blue} hnswlib & 2211 & 3090 & 7591 & 3521 & 1627 & 2069 & 2005 & 863 & 30935 & 1165 & \relbuildtime{51.8}{26.8} \\
\rowcolor{light_blue} vamana(diskann) & 2393 & 4407 & 4237 & 3597 & 2378 & 1644 & 1546 & 2178 & 31148 & 1205 & \relbuildtime{52.7}{16.0} \\
\rowcolor{light_blue} rabitq-hnsw & 4610 & 4592 & 11013 & 7820 & 3138 & 3350 & 3375 & 2218 & 33884 & 2814 & \relbuildtime{89.1}{34.2} \\
\rowcolor{light_blue} ngt-qg & 14735 & 7352 & 40140 & 12047 & 3898 & 7977 & 5188 & 3486 & - & 7665 & \relbuildtime{209.7}{166.3} \\
\bottomrule
\end{NiceTabular}
\end{table*}
 
Comparing index construction times, it is worth noting that the hyperparameters in \textsc{VIBE} have been tuned for optimal query throughput, and index construction times may vary significantly depending on hyperparameter choices. We also note that, in general, the index construction of clustering-based and tree-based methods can be parallelized more easily than the index construction of graph-based methods.

\clearpage
\section{Additional results}
\label{app:additional_results}

\subsection{Selected algorithms}
\label{app:selected-algorithms}

\begin{figure}[!hbpt]
    \centering
    \begin{minipage}{0.82\linewidth}
    \includegraphics[width=\linewidth]{results/agnews-mxbai-1024-euclidean__arxiv-nomic-768-normalized-qps-recall.png}
    \end{minipage}
    \begin{minipage}{0.17\linewidth}
    \includegraphics[width=\linewidth]{results/agnews-mxbai-1024-euclidean__arxiv-nomic-768-normalized-qps-recall-legend.png}
    \end{minipage}

    \vspace{2mm}

    \begin{minipage}{0.82\linewidth}
    \includegraphics[width=\linewidth]{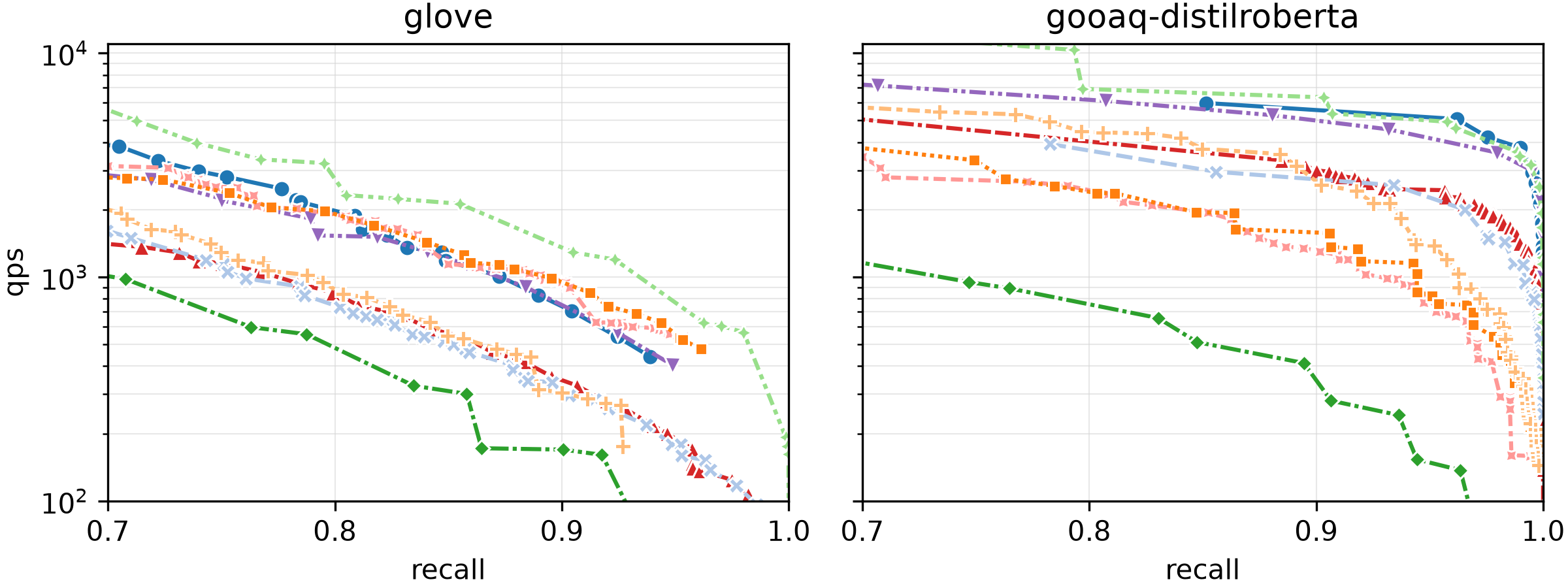}
    \end{minipage}
    \begin{minipage}{0.17\linewidth}
    \includegraphics[width=\linewidth]{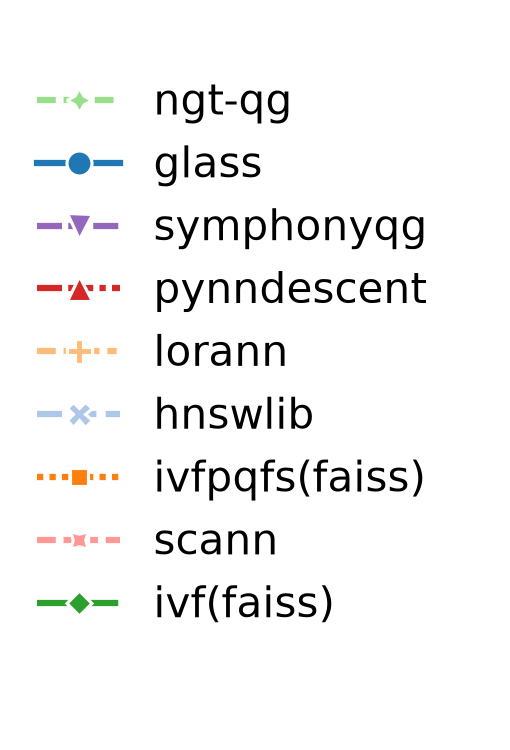}
    \end{minipage}

    \vspace{2mm}

    \begin{minipage}{0.82\linewidth}
    \includegraphics[width=\linewidth]{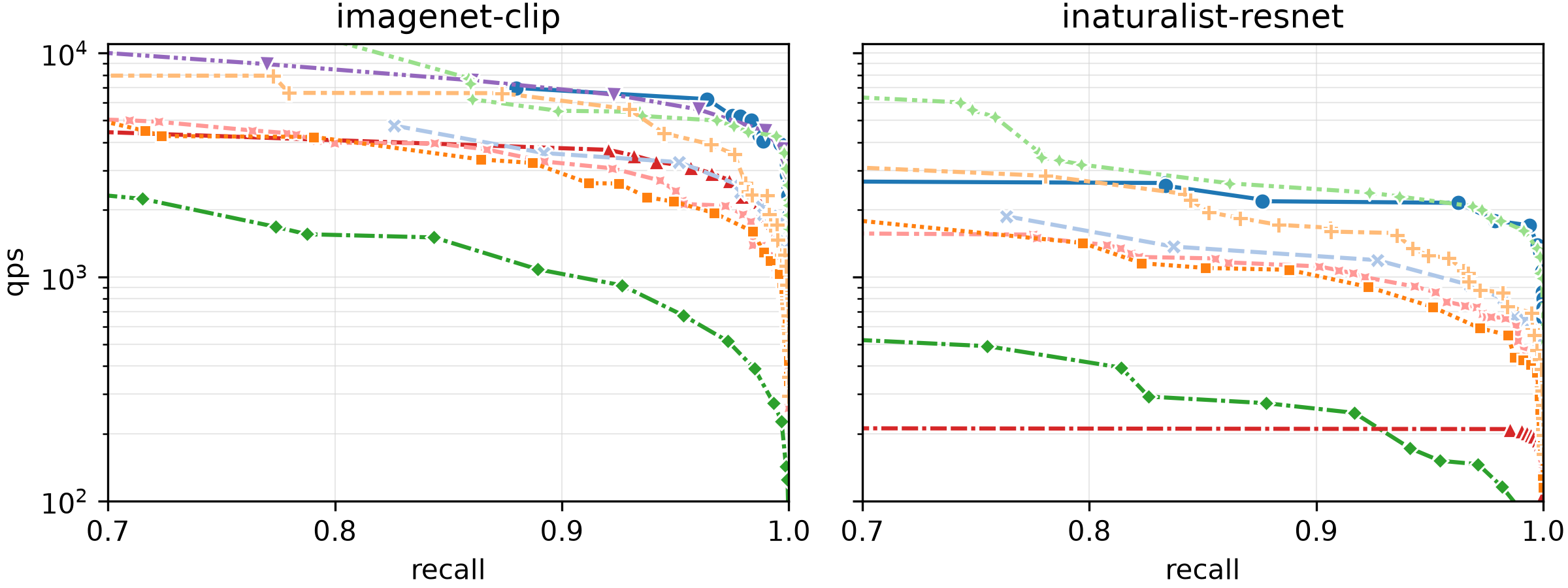}
    \end{minipage}
    \begin{minipage}{0.17\linewidth}
    \includegraphics[width=\linewidth]{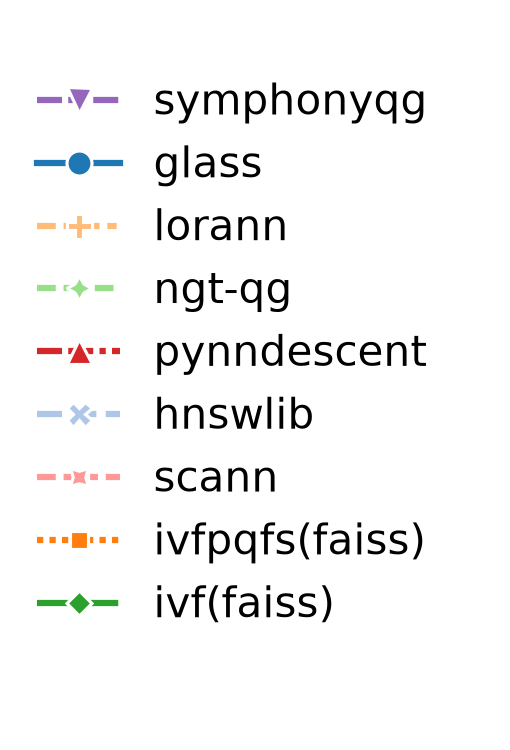}
    \end{minipage}

    \vspace{2mm}

    \begin{minipage}{0.82\linewidth}
    \includegraphics[width=\linewidth]{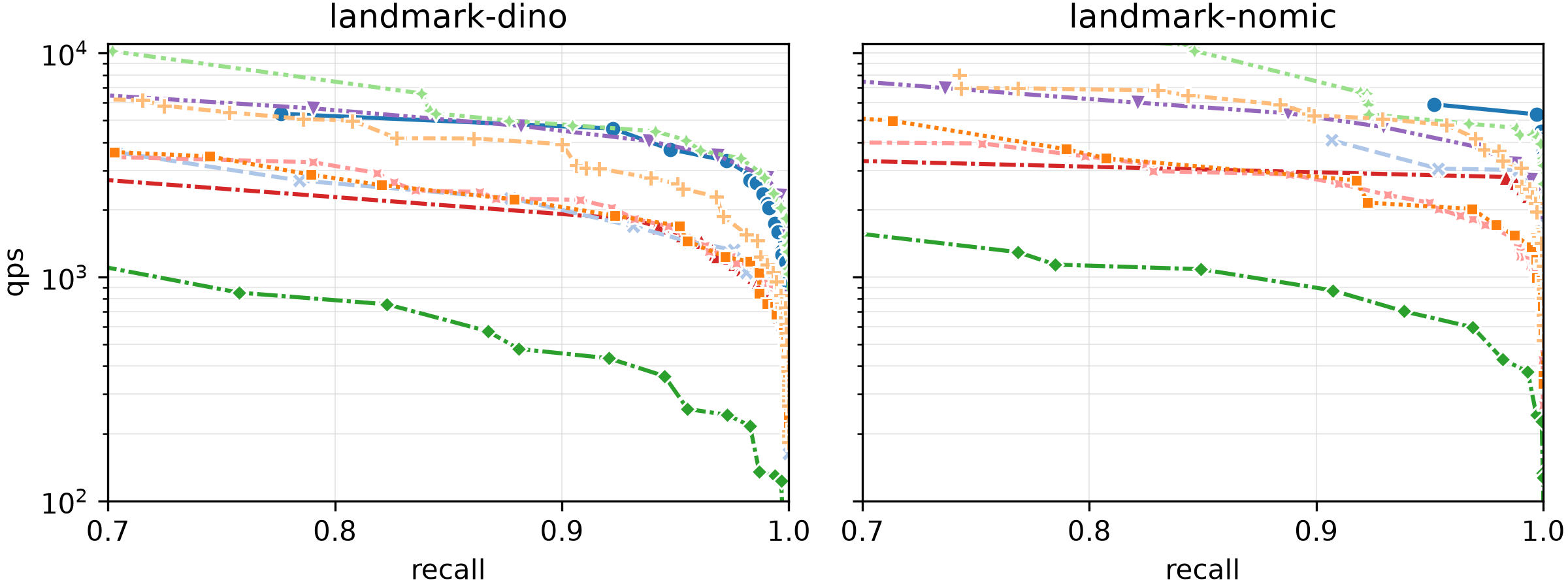}
    \end{minipage}
    \begin{minipage}{0.17\linewidth}
    \includegraphics[width=\linewidth]{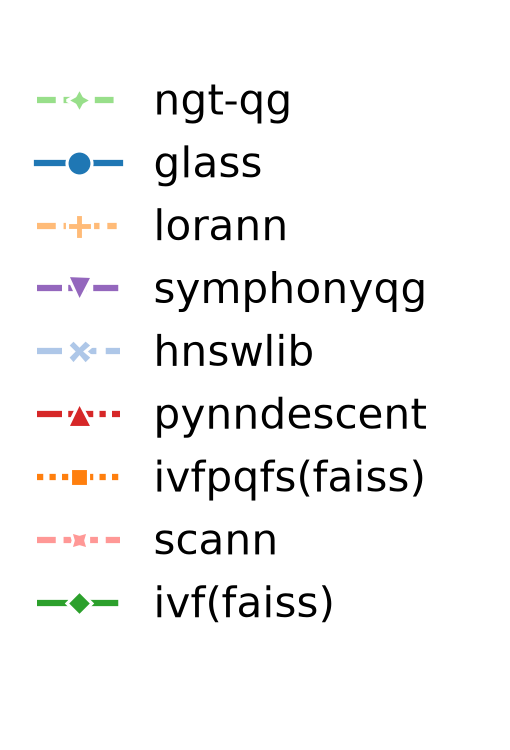}
    \end{minipage}

\end{figure}

\clearpage
\begingroup
\centering

    \begin{minipage}{0.82\linewidth}
    \includegraphics[width=\linewidth]{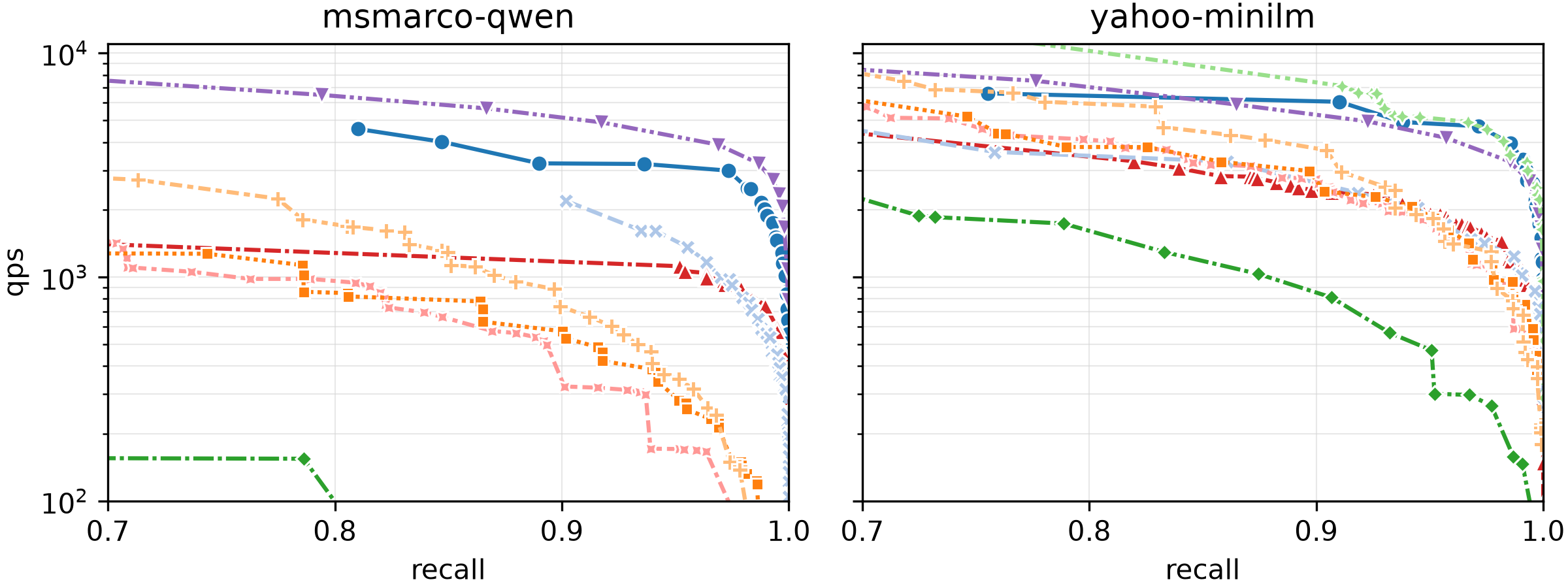}
    \end{minipage}
    \begin{minipage}{0.17\linewidth}
    \includegraphics[width=\linewidth]{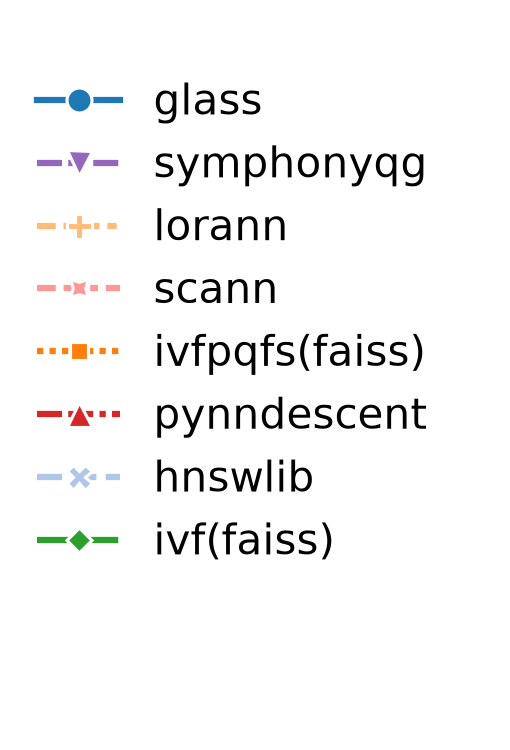}
    \end{minipage}

    \vspace{2mm}

    \begin{minipage}{0.82\linewidth}
    \includegraphics[width=\linewidth]{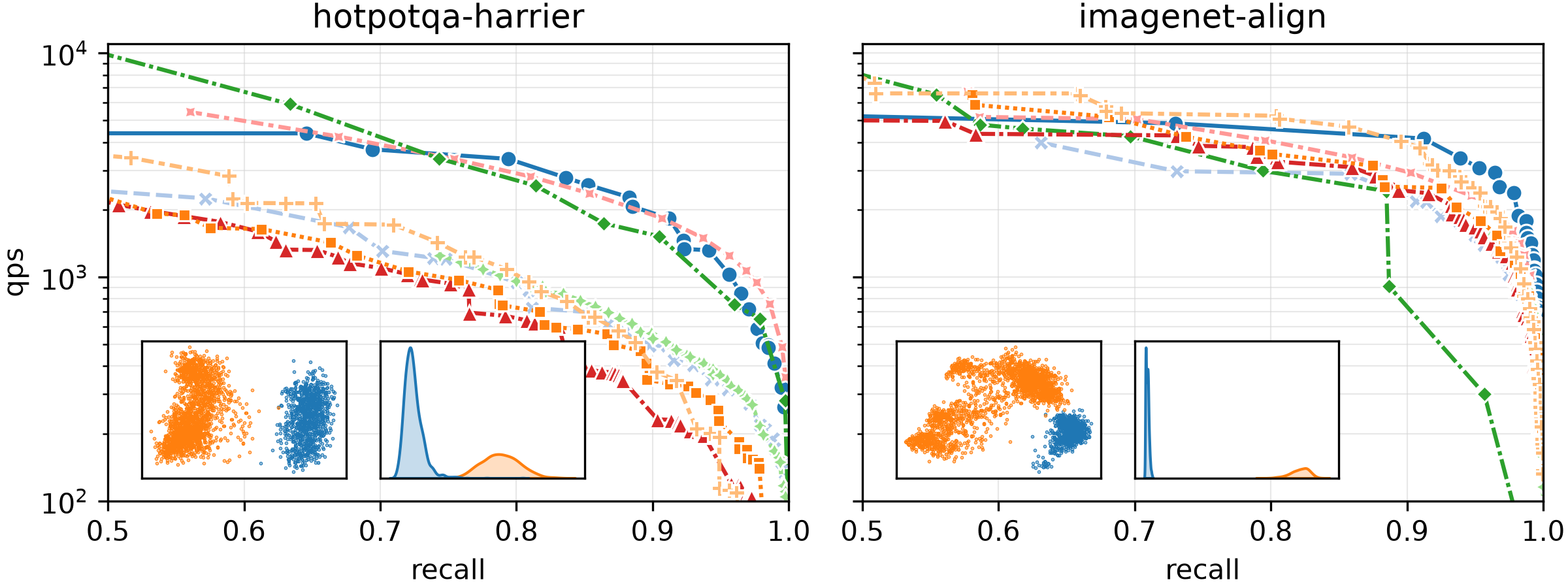}
    \end{minipage}
    \begin{minipage}{0.17\linewidth}
    \includegraphics[width=\linewidth]{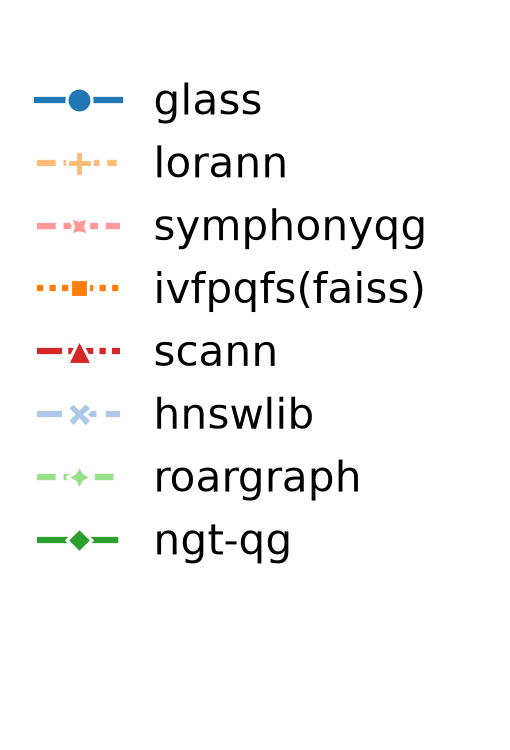}
    \end{minipage}

    \vspace{2mm}

    \begin{minipage}{0.82\linewidth}
    \includegraphics[width=\linewidth]{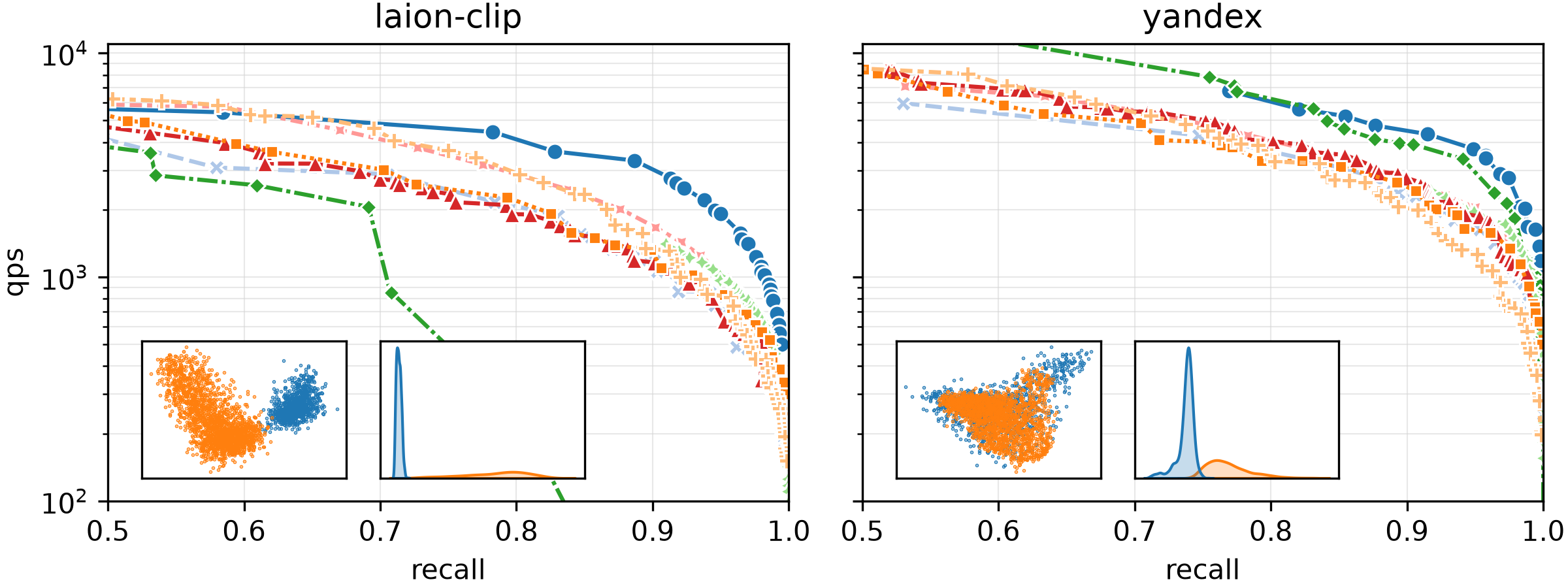}
    \end{minipage}
    \begin{minipage}{0.17\linewidth}
    \includegraphics[width=\linewidth]{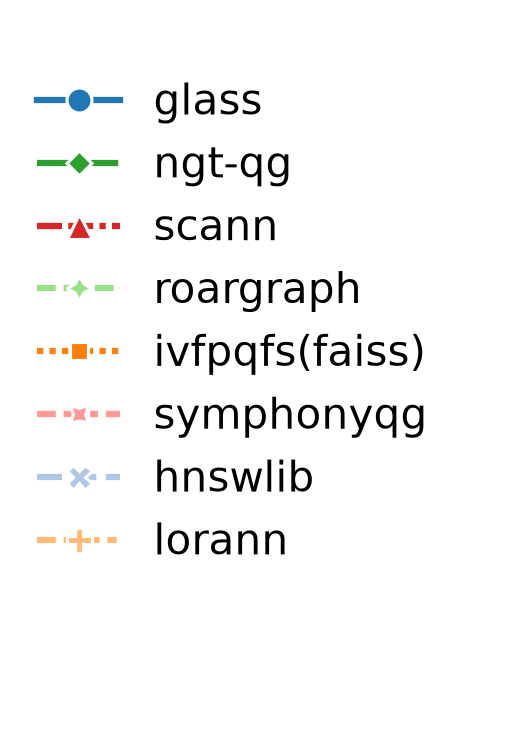}
    \end{minipage}
\par
\endgroup

\clearpage
\subsection{All algorithms}
\label{app:all-algorithms}

In this section, we present benchmarking results for all algorithms on two datasets. For full results on all datasets, we refer to our website: \url{https://vector-index-bench.github.io/}

\begin{figure}[!htpb]
\centering
\includegraphics[width=\linewidth]{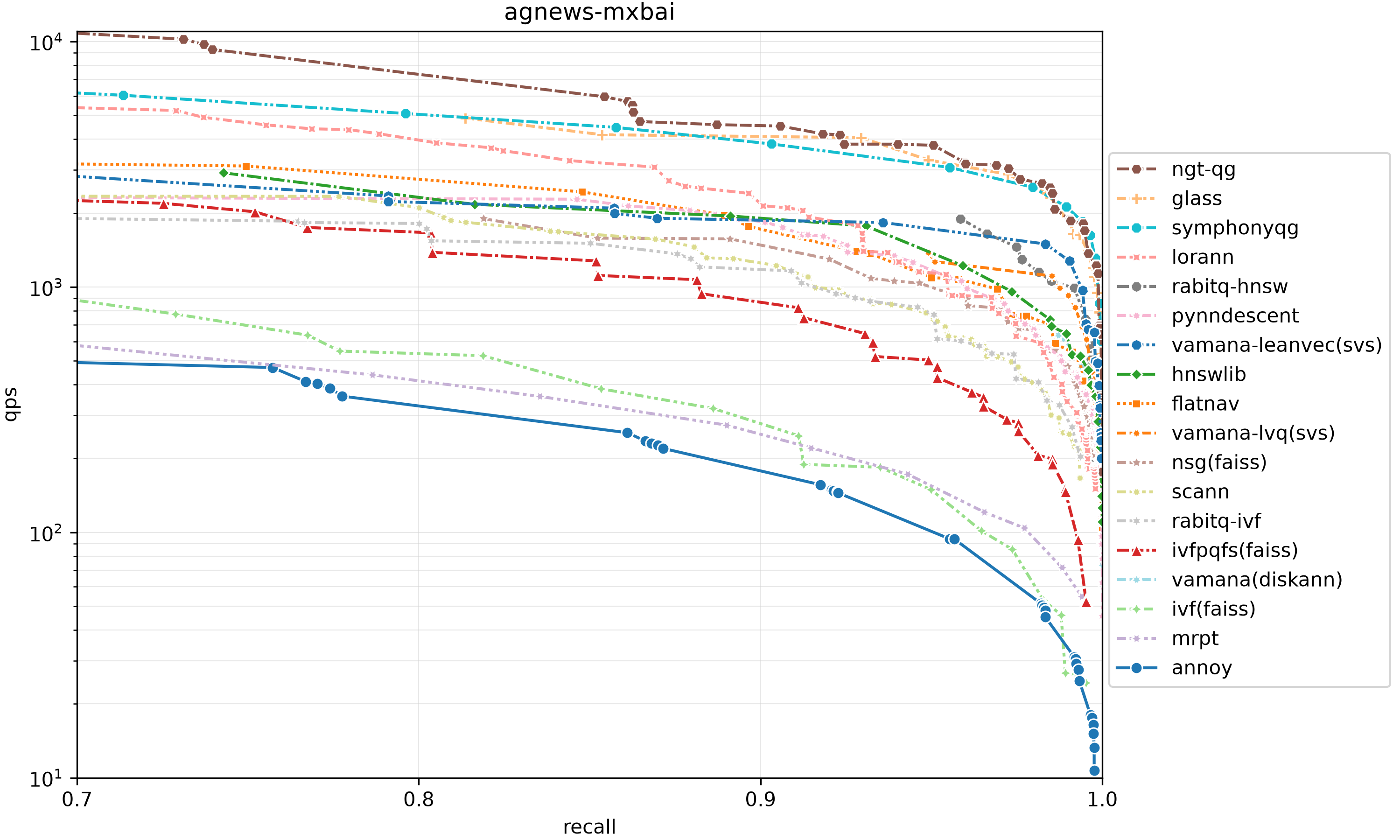}
\end{figure}

\begin{figure}[!htpb]
\centering
\includegraphics[width=\linewidth]{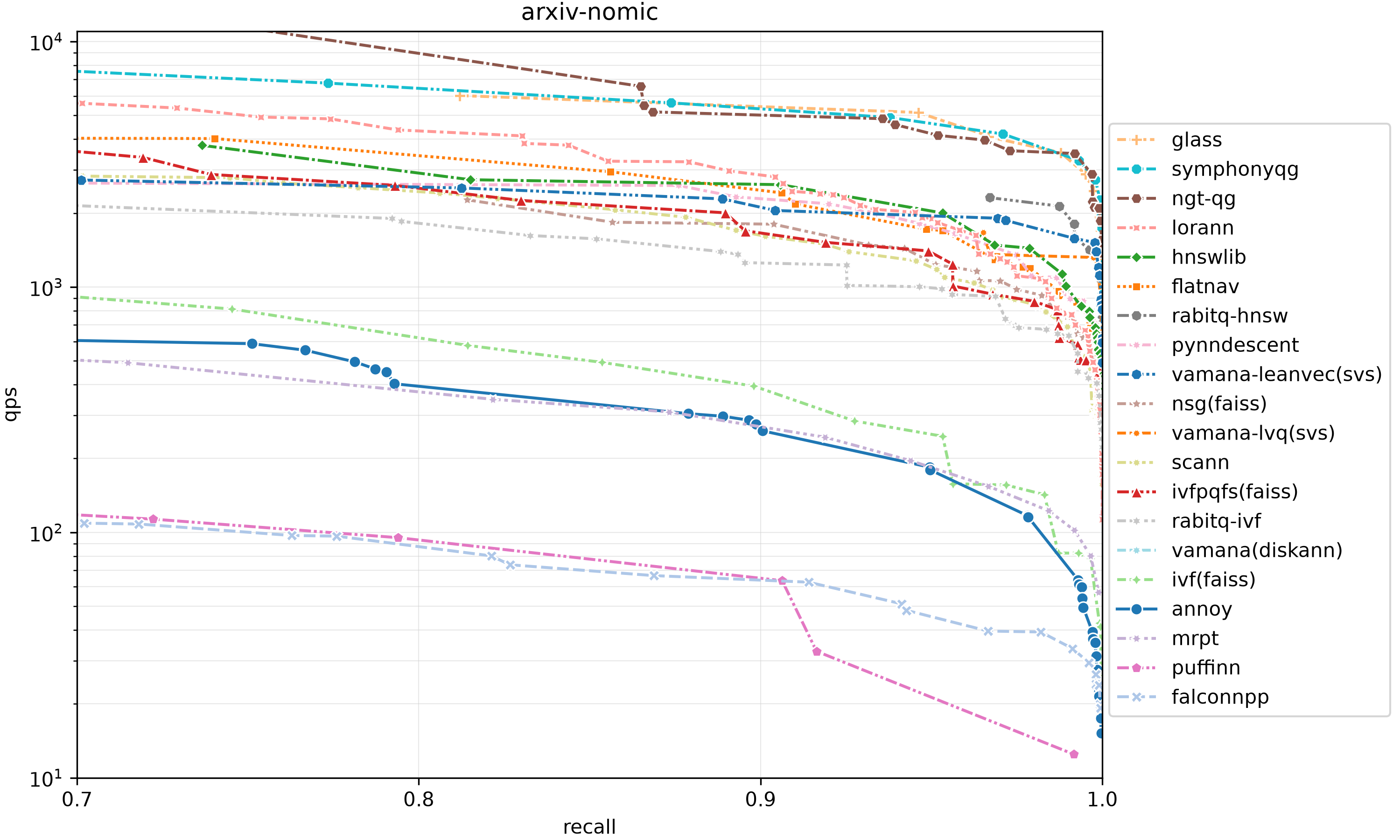}
\end{figure}

\clearpage
\subsection{Binary embeddings and GPU algorithms}
\label{app:binary-and-GPU}

In this section, we present additional results on achieving a higher throughput through binary embeddings and GPU algorithms. For details on the GPU algorithms, see Appendix~\ref{app:impdetails}. For discussion, we refer to Section~\ref{sec:in_distribution_setting}.

\vspace*{1.5\baselineskip}

\begin{figure}[!hbpt]
    \centering
    \begin{minipage}{\linewidth}
    \includegraphics[width=0.87\linewidth]{results/agnews-mxbai-1024-hamming-binary__agnews-mxbai-1024-euclidean-qps-recall-gpu.png}
    \end{minipage}

    \vspace{1mm}

    \begin{minipage}{\linewidth}
    \includegraphics[width=0.87\linewidth]{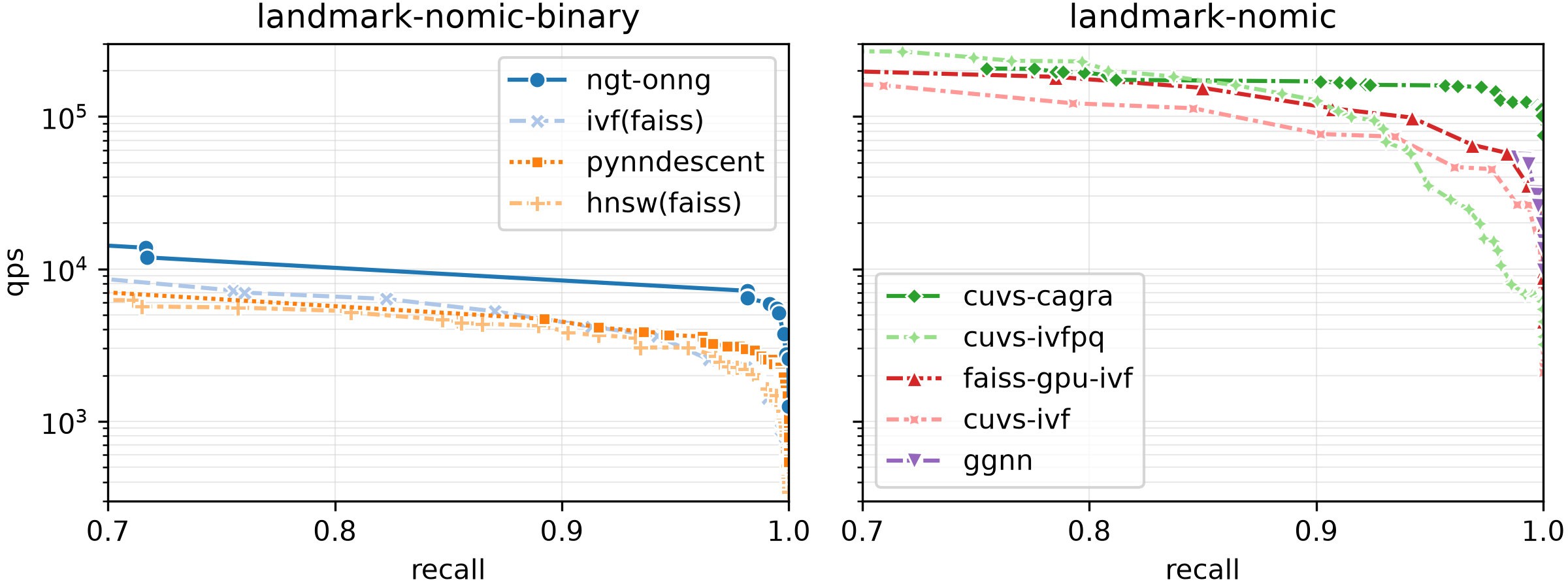}
    \end{minipage}

\end{figure}

\clearpage
\section{Hypervolume}
\label{app:hypervolume}

In Table~\ref{table:hypervolume}, we summarize the recall--throughput trade-off results on 10 datasets using normalized hypervolume over the average recall interval $[0.9, 1]$.
Let $\mathcal C_{a,d} = \{(r_{a,d,i}, q_{a,d,i})\}_i$ denote the (average recall, QPS) pairs for algorithm $a$ on dataset $d$. For a threshold $\tau$, define
\[
Q_{a,d}(\tau)
:=
\max\left(\{q:(r,q)\in\mathcal C_{a,d},\,r\ge\tau\} \cup \{0\} \right),
\qquad
\tilde Q_{a,d}(\tau)
:=
\frac{Q_{a,d}(\tau)}{\max_{a'} Q_{a',d}(0.9)}.
\]
That is, $\tilde Q_{a,d}(\tau)$ is the normalized best throughput among configurations achieving average recall at least $\tau$. The normalized hypervolume is given by
\[
\mathrm{HV}_{a,d}
=
\mathbb E_{\tau\sim \mathrm{Unif}[0.9, 1]}
\left[
\tilde Q_{a,d}(\tau)
\right].
\]

\begin{table*}[ht!]
\centering
\caption{Normalized hypervolume of recall--throughput curves over average recall $[0.9, 1]$. The final column gives mean $\pm$ standard deviation across datasets with rows sorted by decreasing mean. Colors denote \protect\ulc{clustering}{color_green}, \protect\ulc{tree}{color_purple}, \protect\ulc{hashing}{color_orange}, and \protect\ulc{graph}{color_blue} methods.}
\label{table:hypervolume}
\newcommand{\hvsummarystat}[2]{$#1$\,{\fontsize{7}{8}\selectfont\scalebox{0.72}{$\pm$}$#2$}}
\setlength{\tabcolsep}{4pt}
\renewcommand{\arraystretch}{0.95}
\begin{NiceTabular}{>{\small}l r r r r r r r r r r | >{\footnotesize}r }
\toprule
{\small algorithm} &
\rotatebox{90}{\small agnews-mxbai} &
\rotatebox{90}{\small arxiv-nomic} &
\rotatebox{90}{\small glove} &
\rotatebox{90}{\small gooaq-distilroberta} &
\rotatebox{90}{\small imagenet-clip} &
\rotatebox{90}{\small inaturalist-resnet} &
\rotatebox{90}{\small landmark-dino} &
\rotatebox{90}{\small landmark-nomic} &
\rotatebox{90}{\small msmarco-qwen} &
\rotatebox{90}{\small yahoo-minilm} &
\multicolumn{1}{c}{\rotatebox{90}{\small \shortstack{hypervolume\\(mean $\pm$ sd)}}} \\
\midrule
\rowcolor{light_blue} glass & 0.67 & 0.83 & 0.16 & 0.71 & 0.86 & 0.83 & 0.72 & 0.84 & 0.57 & 0.63 & \hvsummarystat{0.68}{0.20} \\
\rowcolor{light_blue} ngt-qg & 0.73 & 0.80 & 0.52 & 0.69 & 0.75 & 0.85 & 0.79 & 0.77 & 0.00 & 0.71 & \hvsummarystat{0.66}{0.24} \\
\rowcolor{light_blue} symphonyqg & 0.60 & 0.81 & 0.18 & 0.59 & 0.81 & 0.00 & 0.72 & 0.57 & 0.76 & 0.54 & \hvsummarystat{0.56}{0.26} \\
\rowcolor{light_blue} rabitq-hnsw & 0.35 & 0.42 & 0.09 & 0.30 & 0.42 & 0.67 & 0.36 & 0.49 & 0.38 & 0.27 & \hvsummarystat{0.38}{0.14} \\
\rowcolor{light_green} lorann & 0.25 & 0.33 & 0.06 & 0.22 & 0.61 & 0.50 & 0.47 & 0.63 & 0.07 & 0.25 & \hvsummarystat{0.34}{0.20} \\
\rowcolor{light_blue} vamana-leanvec(svs) & 0.34 & 0.35 & 0.06 & 0.32 & 0.30 & 0.48 & 0.32 & 0.37 & 0.37 & 0.25 & \hvsummarystat{0.32}{0.10} \\
\rowcolor{light_blue} hnswlib & 0.26 & 0.33 & 0.13 & 0.30 & 0.42 & 0.38 & 0.28 & 0.45 & 0.25 & 0.24 & \hvsummarystat{0.30}{0.09} \\
\rowcolor{light_blue} flatnav & 0.23 & 0.29 & 0.10 & 0.30 & 0.43 & 0.39 & 0.26 & 0.45 & 0.23 & 0.22 & \hvsummarystat{0.29}{0.10} \\
\rowcolor{light_blue} pynndescent & 0.24 & 0.31 & 0.14 & 0.35 & 0.45 & 0.09 & 0.30 & 0.41 & 0.20 & 0.26 & \hvsummarystat{0.27}{0.11} \\
\rowcolor{light_blue} vamana-lvq(svs) & 0.26 & 0.30 & 0.09 & 0.29 & 0.28 & 0.00 & 0.27 & 0.28 & 0.35 & 0.23 & \hvsummarystat{0.23}{0.10} \\
\rowcolor{light_green} scann & 0.15 & 0.21 & 0.26 & 0.11 & 0.35 & 0.33 & 0.32 & 0.29 & 0.04 & 0.22 & \hvsummarystat{0.23}{0.10} \\
\rowcolor{light_blue} nsg(faiss) & 0.20 & 0.23 & 0.00 & 0.22 & 0.30 & 0.47 & 0.20 & 0.31 & 0.20 & 0.15 & \hvsummarystat{0.23}{0.11} \\
\rowcolor{light_green} ivfpqfs(faiss) & 0.10 & 0.22 & 0.32 & 0.14 & 0.30 & 0.28 & 0.30 & 0.29 & 0.06 & 0.23 & \hvsummarystat{0.22}{0.09} \\
\rowcolor{light_green} rabitq-ivf & 0.15 & 0.18 & 0.13 & 0.10 & 0.25 & 0.32 & 0.24 & 0.30 & 0.07 & 0.23 & \hvsummarystat{0.20}{0.08} \\
\rowcolor{light_blue} vamana(diskann) & 0.14 & 0.15 & 0.13 & 0.14 & 0.21 & 0.20 & 0.15 & 0.16 & 0.16 & 0.14 & \hvsummarystat{0.16}{0.03} \\
\rowcolor{light_green} ivf(faiss) & 0.03 & 0.04 & 0.05 & 0.02 & 0.09 & 0.06 & 0.06 & 0.09 & 0.00 & 0.05 & \hvsummarystat{0.05}{0.03} \\
\rowcolor{light_purple} mrpt & 0.03 & 0.03 & 0.03 & 0.03 & 0.03 & 0.03 & 0.03 & 0.07 & 0.00 & 0.04 & \hvsummarystat{0.03}{0.02} \\
\rowcolor{light_purple} annoy & 0.02 & 0.03 & 0.02 & 0.01 & 0.06 & 0.04 & 0.03 & 0.05 & 0.01 & 0.02 & \hvsummarystat{0.03}{0.02} \\
\rowcolor{light_orange} falconnpp & 0.00 & 0.01 & 0.05 & 0.02 & 0.02 & 0.04 & 0.02 & 0.01 & 0.01 & 0.03 & \hvsummarystat{0.02}{0.01} \\
\rowcolor{light_orange} puffinn & 0.00 & 0.00 & 0.02 & 0.02 & 0.01 & 0.03 & 0.02 & 0.01 & 0.00 & 0.02 & \hvsummarystat{0.01}{0.01} \\
\bottomrule
\end{NiceTabular}
\end{table*}
 
\clearpage
\section{OOD performance gap}
\label{app:additional_ood}

In this section, we present the gap between the performance of in-distribution and out-of-distribution queries on the text-retrieval and text-to-image OOD datasets. For discussion, we refer to Section~\ref{sec:OOD_setting}.

\begin{figure}[!htpb]
    \vspace*{1cm}
    \centering
    \includegraphics[width=\linewidth]{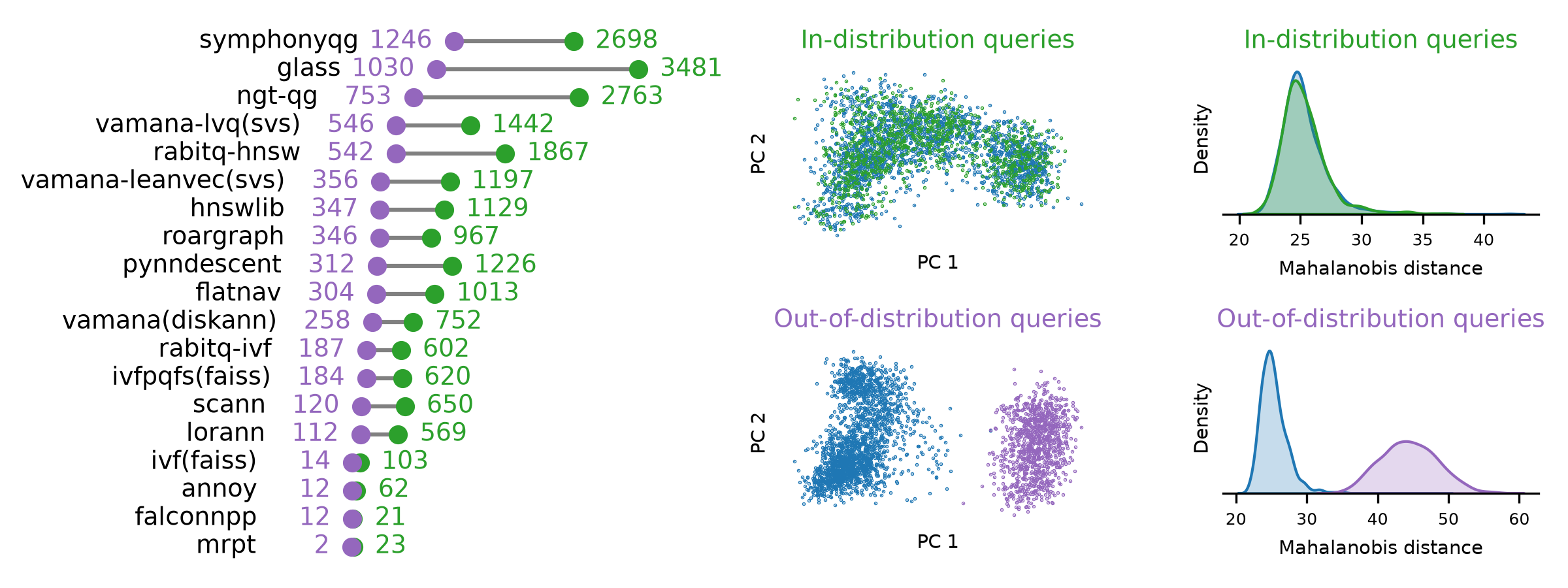}
    \caption{OOD performance gap on \texttt{hotpotqa-harrier}}
\end{figure}

\begin{figure}[!htpb]
    \vspace*{1cm}
    \centering
    \includegraphics[width=\linewidth]{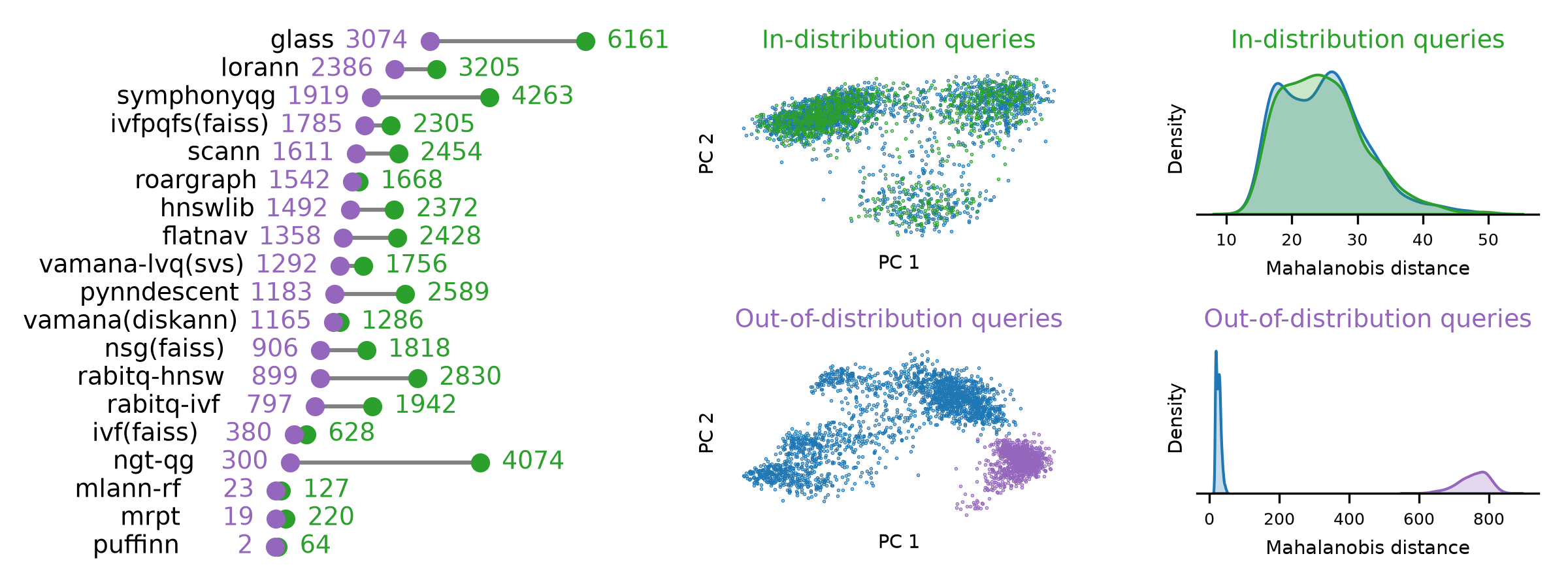}
    \caption{OOD performance gap on \texttt{imagenet-align}}
\end{figure}

\clearpage

\begin{figure}[!p]
    \vspace*{-5cm}
    \centering
    \includegraphics[width=\linewidth]{results/performance-gap-laion-clip-512-normalized.png}
    \caption{OOD performance gap on \texttt{laion-clip}}
    \vspace{2cm}
    \includegraphics[width=\linewidth]{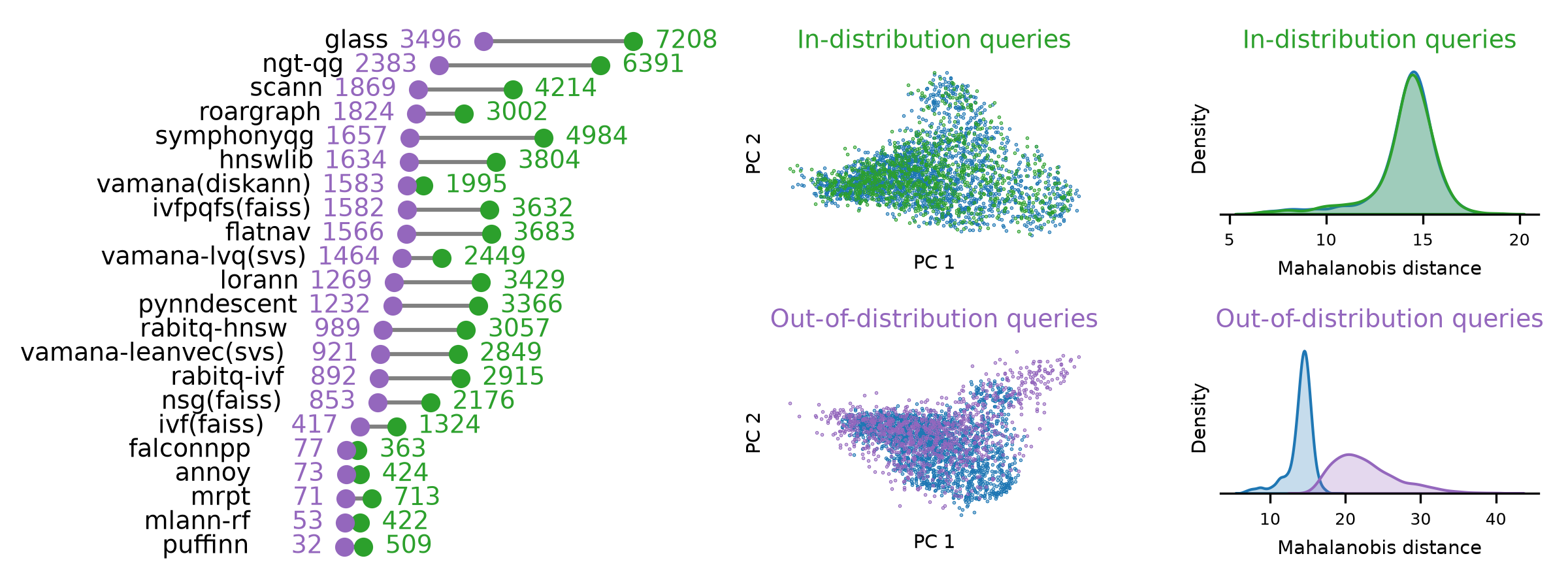}
    \caption{OOD performance gap on \texttt{yandex}}
\end{figure}

\end{document}